\documentclass{article} 
\usepackage{iclr2027_conference,times}

\usepackage{amsmath,amsfonts,bm}

\def\eqref#1{equation~\ref{#1}}

\def\1{\bm{1}}

\DeclareMathAlphabet{\mathsfit}{\encodingdefault}{\sfdefault}{m}{sl}
\SetMathAlphabet{\mathsfit}{bold}{\encodingdefault}{\sfdefault}{bx}{n}

\usepackage[utf8]{inputenc} 
\usepackage[T1]{fontenc}    
\usepackage{hyperref}       
\usepackage{url}            
\usepackage{booktabs}       
\usepackage{amsfonts}       
\usepackage{nicefrac}       
\usepackage{microtype}      
\usepackage{xcolor}         
\usepackage[table]{xcolor}
\usepackage[dvipsnames]{xcolor}
\usepackage{graphicx}
\usepackage{amsmath}
\usepackage{enumitem}
\usepackage{multirow}
\usepackage{subcaption}
\usepackage{wrapfig}
\usepackage{lipsum}
\usepackage{placeins}
\usepackage{comment}
\usepackage{bm}
\usepackage{colortbl}
\usepackage{array}  
\usepackage{amsthm}
\usepackage{tabularx}
\usepackage{makecell}
\usepackage{titletoc}

\definecolor{lightblue}{HTML}{ccd8e6}
\usepackage{amsmath,amssymb}
\usepackage{algorithm}
\usepackage{algpseudocode}

\title{
Robust Evidential Learning Through Latent Consistency
}

\author{
Charmaine Barker$^{1}$\thanks{Equal contribution},
Daniel Bethell$^{1}$\footnotemark[1],
Simos Gerasimou$^{1,2}$ \\
\\
$^{1}$Department of Computer Science, University of York, UK \\
$^{2}$Department of Elect. Eng., and Computer Science and Eng., Cyprus University of Technology, Cyprus \\
\texttt{\{charmaine.barker, daniel.bethell, simos.gerasimou\}@york.ac.uk}
}

\newcommand{\simos}[1]{\textcolor{black}{#1}}

\iclrfinalcopy 
\begin{document}

\maketitle

\begin{abstract}
Reliable uncertainty quantification is essential for deploying deep learning models in high-stakes settings, where out-of-distribution and adversarial inputs can induce confident but unreliable predictions. 
Evidential Deep Learning provides efficient uncertainty estimates in a single forward pass, but can still assign high evidential strength to inputs that are poorly supported by the learned representation, such as adversarial inputs. 
We introduce CLEAR, a lightweight, task-agnostic post-hoc method that improves evidential robustness without retraining or altering the base prediction.
Using held-out calibration data, CLEAR characterises the group-conditioned geometry of the model's latent space. At inference, it efficiently generates perturbation views directly in the latent space and measures their conflict relative to the calibrated geometry of the predicted group. High latent conflict indicates unsupported evidence, which CLEAR uses to selectively reduce evidential strength while retaining evidence for latent-consistent inputs.
On ImageNet$\rightarrow$CUB, CLEAR improves OOD and adversarial AUROC by $+8.29$ and $+5.01$ while running 17.4$\times$ faster than competing post-hoc methods while preserving predictive performance across classification, regression, and object detection benchmarks.

\end{abstract}

\section{Introduction}
\label{sec:Introduction}

Deep learning has enabled increasingly capable predictive systems across domains such as computer vision and natural language processing~\citep{dosovitskiy2020image,achiam2023gpt}. However, reliable deployment requires more than predictive accuracy. In safety-critical settings, models may encounter out-of-distribution (OOD) inputs that differ from the data observed during training, or adversarial inputs deliberately perturbed to induce erroneous predictions. In both cases, a model may remain highly confident despite operating beyond the conditions \simos{ or environment under which it has been trained.} 
Identifying such unreliable predictions is therefore fundamental to the safe deployment of deep learning systems.

Uncertainty Quantification (UQ) provides a means of expressing how strongly a model's prediction should be trusted~\citep{abdar2021review}. Approaches such as Bayesian neural networks~\citep{goan2020bayesian}, Monte Carlo Dropout~\citep{gal2016dropout}, and conformal methods~\citep{angelopoulos2023conformal} can provide reliable uncertainty estimates; however, they often require additional training or repeated inference. Evidential methods offer an efficient alternative by estimating uncertainty in a single deterministic forward pass. Evidential Deep Learning (EDL) models class probabilities using a Dirichlet distribution~\citep{sensoy2018evidential}, while Deep Evidential Regression (DER) uses a Normal-Inverse-Gamma distribution for continuous predictions~\citep{amini2020deep}. \simos{Despite their appeal}, evidential models can still assign strong evidence to OOD and adversarial inputs for which the prediction is poorly supported~\citep{kopetzki2021evaluating}. 

Post-hoc methods have emerged as a particularly effective approach for improving uncertainty robustness in evidential models, as they enhance uncertainty estimation without retraining the predictor and have demonstrated strong robustness to OOD and adversarial inputs~\citep{kopetzki2021evaluating,barker2026robust}. However, existing approaches may require repeated model evaluations or transformed inputs, reducing the computational advantages of evidential learning. They may also suppress evidence for valid in-distribution (ID) inputs, reducing retention despite correct predictions.

We introduce \textbf{C}onflict-aware \textbf{L}atent \textbf{E}vidence \textbf{A}djustment for adversarial \textbf{R}obustness (\textbf{CLEAR}), a task-agnostic post-hoc approach that evaluates whether the evidence assigned by a pretrained evidential model is supported by its expected latent representation. Rather than repeatedly evaluating transformed inputs through the full model, CLEAR first uses held-out ID data to calibrate the latent geometry associated with different regions of the output space. At inference, CLEAR then generates inexpensive perturbation views directly in latent space and measures their conflict relative to the calibrated geometry of the predicted output group. Predictions exhibiting latent behaviour consistent with the calibrated ID structure retain their evidential strength, whereas high latent conflict indicates unsupported evidence and results in increased uncertainty. CLEAR therefore improves uncertainty under OOD and adversarial inputs without retraining the predictor or replacing its base prediction.

\section{Related Work}
\label{sec:Related Work}

\textbf{Uncertainty Quantification.} UQ characterises uncertainty arising from incomplete model knowledge and variability in observations~\citep{he2023survey}. Input-dependent distributions capture aleatoric uncertainty~\citep{kendall2017uncertainties}, while Bayesian inference~\citep{goan2020bayesian,blei2017variational}, Monte Carlo Dropout~\citep{gal2016dropout}, deep ensembles~\citep{lakshminarayanan2017simple}, and distance-aware models~\citep{liu2020simple} address epistemic uncertainty. Ensembles with prediction intervals can account for both sources~\citep{pearce2018high}. Test-time augmentation measures predictive variability without explicitly separating them~\citep{ayhan2018test}. Conformal methods provide calibrated sets or intervals whose cardinality or width can serve as uncertainty indicators~\citep{angelopoulos2023conformal}. These approaches differ in their computational and calibration requirements, motivating efficient estimation of both uncertainty types.

\textbf{Evidential Deep Learning.} Evidential deep learning (EDL) applies Dempster-Shafer theory (DST)~\citep{dempster1968generalization} to model class probabilities with a Dirichlet distribution, enabling epistemic and aleatoric uncertainty estimation in a single forward pass. Its full formulation is given in Section~\ref{sec:Preliminaries}. EDL has been widely used for OOD detection by assigning greater uncertainty to unexpected inputs. However, its deterministic inference lacks the multiple predictive perspectives provided by methods such as Monte Carlo Dropout or deep ensembles, making overconfident errors difficult to recover from. This limitation is particularly evident under adversarial perturbations~\citep{kopetzki2021evaluating}.

Post-hoc EDL methods can improve the robustness of evidential models to adversarial and distribution-shifted inputs by refining uncertainty after the base model has been trained~\citep{barker2026robust, kopetzki2021evaluating}. However, many approaches rely on multiple forward passes or stochastic sampling, increasing computational cost and sacrificing one of EDL's key advantages: single-pass uncertainty estimation. Others are tailored to specific tasks, e.g., classification~\citep{barker2026robust}, limiting their generality. These limitations motivate a lightweight post-hoc approach that strengthens evidential uncertainty under challenging inputs while preserving the efficiency and task flexibility of the underlying task.

\section{Preliminaries}
\label{sec:Preliminaries}
We consider a supervised learning setting where $\mathcal{X}$ and $\mathcal{Y}$ denote the input and task-dependent output spaces, respectively. Given a dataset $\mathcal{D}=\{(x_i,y_i)\}_{i=1}^{N}$ with samples $(x_i,y_i)\in\mathcal{X}\times\mathcal{Y}$ drawn i.i.d. from an underlying distribution $p(x,y)$, the objective is to learn a predictive model with a latent representation $h(x)\in\mathbb{R}^d$ that captures the relationship between $x$ and $y$ and generalizes to unseen samples. CLEAR is a task-agnostic, post-hoc method that operates on pretrained evidential models. We begin by introducing the two evidential formulations considered throughout this work.

\textbf{Classification.}
Each target $y_i$ denotes one of $K$ classes, such that $y_i \in \{1,\ldots,K\}$. Evidential Deep Learning (EDL) replaces the softmax output of the model with a Dirichlet distribution over class probabilities. The model predicts non-negative evidence to define the Dirichlet parameters:
\begin{equation}
e(x) \in \mathbb{R}_{+}^{K}, \qquad \alpha(x)=e(x)+1
\end{equation}
An EDL model is trained by matching the Dirichlet mean to the one-hot target while discouraging unsupported evidence. For a one-hot target $y_i$, the standard objective combines a squared-error term, the Dirichlet predictive variance, and a KL penalty to the uniform Dirichlet prior:
\begin{equation}
\mathcal{L}_{\mathrm{EDL}}
=\sum_{k=1}^{K}(y_{ik}-\hat{p}_k(x_i))^2
+\sum_{k=1}^{K}
\frac{\hat{p}_k(x_i)(1-\hat{p}_k(x_i))}{S(x_i)+1} +\lambda_{\mathrm{KL}}\mathrm{KL}\!\left(\mathrm{Dir}(\alpha(x_i))\,\Vert\,\mathrm{Dir}(\mathbf{1})\right)
\end{equation}
Given an input $x_i$, the predictive categorical mean, total evidential strength, and scalar uncertainty are:
\begin{equation}
\hat{p}_k(x_i)=\frac{\alpha_k(x_i)}{S(x_i)},\qquad S(x_i)=\sum_{k=1}^{K}\alpha_k(x_i), \qquad u(x_i)=\frac{K}{S(x_i)}
\end{equation}
High $S(x_i)$ signifies concentrated evidence around the predicted class, while low $S(x_i)$ indicates uncertainty.

\textbf{Regression.}
Each target $y_i$ denotes a continuous value, i.e., $y_i \in \mathbb{R}$. Deep Evidential Regression (DER) models uncertainty by placing a Normal-Inverse-Gamma (NIG) distribution over the parameters of a Gaussian likelihood:
\begin{equation}
y \sim \mathcal{N}(\mu,\sigma^2), \qquad (\mu,\sigma^2)\sim \mathrm{NIG}(\gamma,\nu,\alpha,\beta)
\end{equation}
where $\gamma$ represents the predictive mean and $\nu> 0$, $\alpha > 1$, and $\beta >0$ determine the evidential strength and uncertainty. A DER model is trained using the NIG negative log-likelihood together with an evidence regulariser that penalises confident but inaccurate predictions:
\begin{equation}
\begin{aligned}
\mathcal{L}_{\mathrm{DER}}
={}&
\frac{1}{2}\log\left(\frac{\pi}{\nu}\right)
-\alpha\log\Omega
+\left(\alpha+\frac{1}{2}\right)
\log\left((y-\gamma)^2\nu+\Omega\right) \\
&+\log\Gamma(\alpha)
-\log\Gamma\left(\alpha+\frac{1}{2}\right)
+\lambda_{\mathrm{reg}}|y-\gamma|(2\nu+\alpha)
\end{aligned}
\end{equation}
where $\Omega = 2\beta(1+\nu)$. For an input, the predictive mean and combined aleatoric and epistemic variance are defined as:
\begin{equation}
\mathbb{E}[y]=\gamma, \qquad \mathrm{Var}(y)=\frac{\beta(1+\nu)}{\nu(\alpha-1)}
\end{equation}
The aleatoric and epistemic uncertainties are defined as:
\begin{equation}
u_{\mathrm{alea}}(x)=\frac{\beta(x)}{\alpha(x)-1},
\qquad
u_{\mathrm{epi}}(x)=\frac{\beta(x)}{\nu(x)(\alpha(x)-1)}
\end{equation}
Both EDL and DER encode prediction confidence through evidential strength: Dirichlet concentration in classification and NIG strength in regression. However, high evidence can still be assigned to inputs whose latent representation is inconsistent with the learned structure of the predicted class or output region. CLEAR addresses this by recalibrating evidential strength post-hoc using a latent consistency test, while leaving the base prediction unchanged.

\section{Conflict-Aware Latent Evidence Adjustment for Adversarial Robustness}
\label{sec:clear}

\begin{figure}
    \centering
    \includegraphics[width=0.95\linewidth]{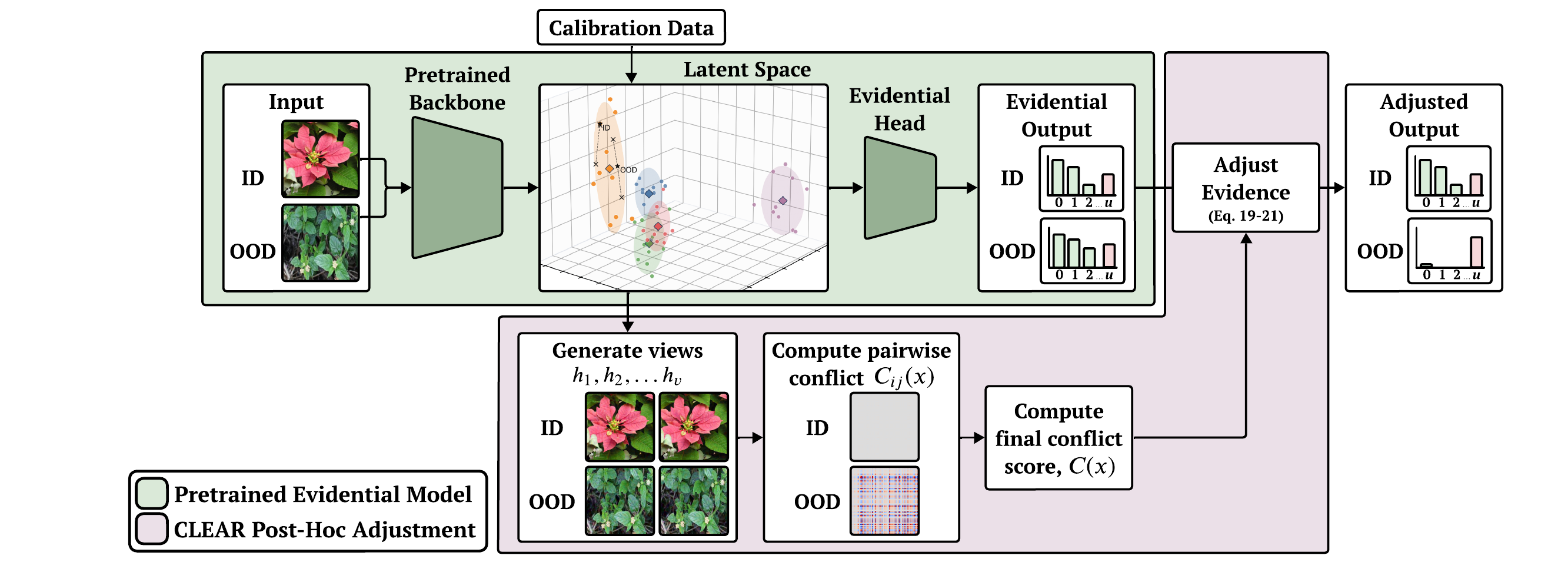}
    \caption{CLEAR architecture overview. A pretrained evidential model produces a latent representation and evidential output. CLEAR uses calibration-derived latent geometry to generate perturbed views, computes pairwise latent conflict, and aggregates this into the final conflict score $C(x)$. This score is used to down-weight unsupported evidential strength post-hoc while preserving the original prediction. The illustrative ID and OOD examples are drawn from Oxford Flowers and DeepWeeds.}
    \label{fig:architecture}
\end{figure}

Our \textbf{C}onflict-aware \textbf{L}atent \textbf{E}vidence \textbf{A}djustment for adversarial \textbf{R}obustness (\textbf{CLEAR}) approach, whose high-level workflow is shown in Figure~\ref{fig:architecture}, boosts uncertainty under OOD and adversarially attacked inputs by determining whether a model's evidential strength is supported by familiar ID latent structure. 
During calibration, CLEAR characterises the latent structure of ID data from a pretrained evidential model without modifying or retraining the predictor. At inference, the model produces a latent representation $h(x)$ and evidential output, from which CLEAR samples multiple latent perturbation views and forms all 
unordered pairs between them. Their compatibility with the calibrated latent geometry of the predicted group is then measured, and evidential strength is down-weighted when conflict is high. 
Consequently, inputs inconsistent 
with the learned ID structure receive greater uncertainty, while evidence for compatible ID inputs is preserved.

\subsection{Latent Calibration}
\label{sec:clear-calibration}
CLEAR first estimates the approximate latent geometry associated with each output group using a held-out calibration set $\mathcal{D}_{\mathrm{cal}}$. The evidential model is frozen, and CLEAR extracts the latent representation of each calibration example $h_i=h(x_i)$. Each example is assigned to an output group~$g_i$. For classification, the group corresponds directly to the class label $g_i=y_i$. For regression, where the target space is continuous, targets are partitioned into $B$ bins and $g_i=b(y_i)$ where $b(\cdot)$ maps a target value to its corresponding output bin. These bins are formed using calibration-set quantiles, giving approximately equal group cardinalities $N_g$.

CLEAR uses these groups to characterise the expected latent structure across different regions of the output space. For each group $g$, the group-conditioned latent mean and covariance are estimated as:
\begin{equation}
\mu_g=\frac{1}{N_g}\sum_{i:g_i=g}h_i,
\qquad
\Sigma_g=\mathrm{Cov}(h_i|g_i = g)
\end{equation}
Because the covariance is subsequently used to normalise latent representations, it is regularised to ensure stable inversion:
\begin{equation}
\widetilde{\Sigma}_g=\Sigma_g+\lambda_g I
\end{equation}
where $\lambda_g>0$ is a small stabilisation term. 
CLEAR then computes its Cholesky factor as $L_gL_g^\top=\widetilde{\Sigma}_g$. The factor $L_g$ provides an efficient transformation for expressing latent representations relative to the covariance structure of group $g$, which is used during inference to assess whether a prediction behaves consistently with that group. CLEAR also derives a characteristic latent scale for each group:
\begin{equation}
\rho_g=\sqrt{\frac{d}{\left\|L_g^{-1}\right\|_F^2}}
\end{equation}
where $d$ is the dimensionality of the latent representation and $\|\cdot\|_F$ denotes the Frobenius norm. This scale adapts the magnitude of the latent perturbations used at inference to each group's calibrated geometry. Together, $\mu_g$, $L_g$, and $\rho_g$ describe the latent geometry against which future predictions are evaluated. If an input is assigned strong evidence for group $g$, its local latent behaviour is expected to remain compatible with this calibrated structure.

\subsection{Latent Perturbation Views}
\label{sec:clear-views}
At inference time, the frozen evidential model produces its original evidential output and a latent representation $h=h(x)$ from the penultimate layer. As the final feature representation used by the evidential head, this layer captures information relevant to the model's prediction while retaining the feature structure needed to assess whether that prediction is supported by familiar ID data. CLEAR then determines the output group $\hat{g}$ associated with the base prediction.

CLEAR assesses whether this prediction is supported by examining the local behaviour around its latent representation. Using the characteristic scale of the predicted group, it samples $V$ perturbed latent views:
\begin{equation}
\label{eq:latent-views}
h_v=h+\tau\rho_{\hat{g}}\eta_v,
\qquad
\eta_v\sim\mathcal{N}(0,I),
\qquad
v=1,\ldots,V
\end{equation}
where $\tau>0$ controls the overall perturbation magnitude and $\rho_{\hat{g}}$ adapts this magnitude to the calibrated latent scale of the predicted group.

Rather than evaluating individual perturbations independently, CLEAR compares their behaviour with one another. It therefore forms all unordered pairs of latent views:
\begin{equation}
\mathcal{P}_V={(i,j):1\leq i<j\leq V}
\end{equation}
For a well-supported prediction, these neighbouring views should remain mutually compatible when interpreted relative to the calibrated latent geometry of the predicted group. A poorly supported prediction is expected to exhibit greater disagreement under the same perturbations. To measure this compatibility relative to the predicted group rather than in the original latent coordinate system, each perturbation view is centred and whitened~\citep{ermolov2021whitening} using its calibrated statistics:
\begin{equation}
q_v=L_{\hat{g}}^{-1}\left(h_v-\mu_{\hat{g}}\right),
\qquad
v=1,\ldots,V
\end{equation}
The whitened views are then stacked $Q=[q_1,\ldots,q_V]^\top\in\mathbb{R}^{V\times d}$. CLEAR then measures the off-diagonal interactions between distinct latent dimensions across each unordered pair of views. For $(i,j)\in\mathcal{P}_V$, the pairwise conflict is:
\begin{equation}
C_{ij}(x)=
\frac{1}{d(d-1)}
\sum_{\substack{a,b\in\{1,\ldots,d\}\\a\neq b}}
q_{i,a}^{2}q_{j,b}^{2}.
\end{equation}
Averaging across all view pairs gives the raw conflict score:
\begin{equation}
C_{\mathrm{raw}}(x)=\frac{1}{|\mathcal{P}_V|}\sum_{(i,j)\in\mathcal{P}_V}C_{ij}(x)
\end{equation}
Large values indicate stronger cross-dimensional interactions after whitening with respect to the predicted group's latent geometry. CLEAR interprets this as latent conflict: the evidential head associates the input with group $\hat{g}$, but its local latent behaviour is inconsistent with the structure calibrated for that group.

A single global reference conflict level $r$ is estimated using the held-out calibration set. For each calibration example, CLEAR repeatedly samples the latent perturbation views and computes the corresponding raw conflict score. The resulting scores are averaged across the repeated draws to obtain a stable conflict estimate for each example:
\begin{equation}
\overline{C}_{\mathrm{raw}}(x_i)=\frac{1}{T}\sum_{t=1}^{T}C_{\mathrm{raw}}^{(t)}(x_i)
\end{equation}
where $T$ denotes the number of repeated perturbation draws. Repeating the stochastic perturbations reduces variation in the reference estimate for each calibration example. The overall reference conflict is then defined as the median across the calibration set:
\begin{equation}
r=\operatorname{median}_{x_i\in\mathcal{D}_{\mathrm{cal}}}\overline{C}_{\mathrm{raw}}(x_i)
\end{equation}
The raw conflict score is converted into a bounded value:
\begin{equation}
C(x)=\max\left(0,\frac{C_{\mathrm{raw}}(x)-r}{C_{\mathrm{raw}}(x)+r}\right)
\end{equation}
where $r>0$ is the median reference conflict estimated from the calibration set. This mapping leaves low-conflict inputs unchanged and smoothly increases towards one as the raw conflict grows. The corresponding evidence scaling factor is:
\begin{equation}
s(x)=\exp(-\delta C(x))
\end{equation}
where $\delta>0$ controls the strength of evidence down-weighting. Thus, $s(x)\approx 1$ for latent-consistent inputs, while high-conflict inputs receive reduced evidential strength.

CLEAR uses the same latent conflict mechanism for classification and regression; only the final adjustment to the evidential output differs. In both cases, the objective is to reduce unsupported evidential strength without replacing the prediction produced by the frozen base model.

For classification, the original EDL model outputs evidence $e(x)$ and Dirichlet parameters $\alpha(x)=e(x)+\mathbf{1}$. CLEAR rescales the evidence directly $e'(x)=s(x)e(x)$ and forms the adjusted Dirichlet parameters $\alpha'(x)=e'(x)+\mathbf{1}$. The resulting predictive probabilities are then:
\begin{equation}
\hat{p}'_k(x)=\frac{\alpha'_k(x)}{\sum_j \alpha'_j(x)}.
\end{equation}
When the latent conflict is low, $s(x)$ is close to one, and the original evidential prediction is preserved. When conflict is high, the Dirichlet concentration is reduced, increasing uncertainty without requiring any change to the classifier.

For regression, the base DER model outputs NIG parameters $(\gamma(x),\nu(x),\alpha(x),\beta(x))$. CLEAR preserves the predictive mean $\gamma(x)$ and scale parameter $\beta(x)$, and down-weights the evidence-related parameters:
\begin{equation}
\nu'(x)=s(x)\nu(x),
\qquad
\alpha'(x)=1+s(x)(\alpha(x)-1),
\end{equation}
The transformation keeps the NIG distribution valid because $\nu'>0$, $\alpha'>1$, and $\beta>0$. It also ensures that CLEAR does not move the point prediction. Instead, it reduces evidential strength when the predicted output region is not supported by the latent representation.

\begin{figure}[h]
    \centering
    \includegraphics[width=\textwidth]{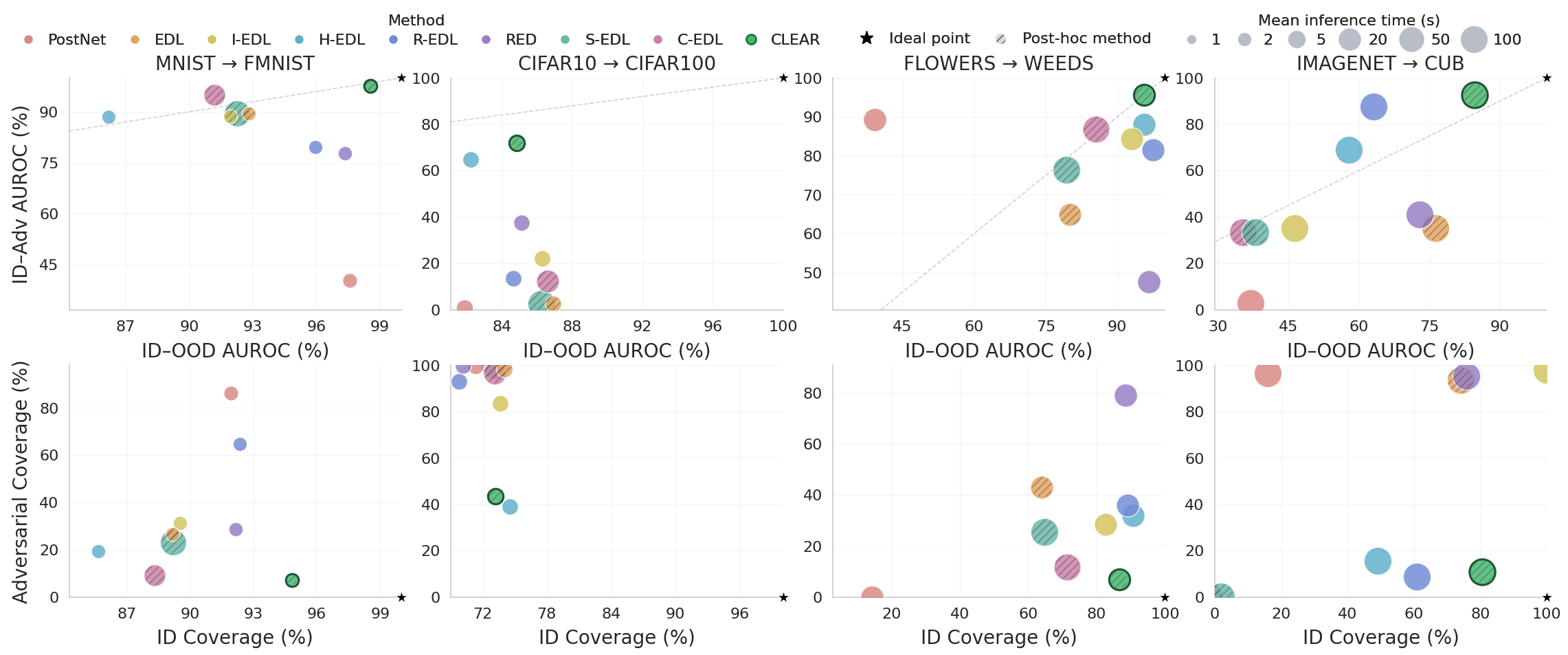}
    \caption{Comparison of nine uncertainty-estimation methods across MNIST→FashionMNIST, CIFAR10→CIFAR100, Oxford Flowers→Deep Weeds, and ImageNet→CUB. Rows show ID–OOD vs ID–Adv. AUROC (top) and ID vs Adv. coverage (bottom). Marker size denotes mean inference time, and the black star marks the ideal operating point.}
    \label{fig:core-results}
\end{figure}

\section{Results}
\label{sec:Results}

We evaluate CLEAR in a comprehensive series of experiments comparing it against state-of-the-art EDL-based and other competitive UQ approaches over 10 independent runs. 
Our evaluation focuses on both performance and uncertainty estimates produced per approach for OOD and adversarially attacked data.

\textbf{Comparative Approaches.} We compare CLEAR against Posterior Networks~\citep{charpentier2020posterior}, Evidential Deep Learning (EDL)~\citep{sensoy2018evidential}, Fisher Information-based EDL ($\mathcal{I}$-EDL)~\citep{deng2023uncertainty}, Hyper-Opinion EDL (H-EDL)~\citep{qu2024hyper}, Relaxed EDL (R-EDL)~\citep{chen2024r}, Regularized Evidential Model (RED)~\citep{pandey2023learn}, Smoothed EDL (S-EDL)~\citep{kopetzki2021evaluating}, and Conflict-aware Evidential Deep Learning (C-EDL)~\citep{barker2026robust} to represent a range of approaches in EDL and UQ that allow for a fair comparison.

\textbf{Datasets.} 
Adopting the procedure on EDL-based evaluation from recent research~\citep{deng2023uncertainty, chen2024r}, we evaluate all approaches on the MNIST~\citep{lecun1998gradient}, FashionMNIST~\citep{xiao2017fashion}, KMNIST~\citep{clanuwat2018deep}, EMNIST~\citep{cohen2017emnist}, CIFAR10~\citep{krizhevsky2009learning}, CIFAR100~\citep{krizhevsky2009learning}, SVHN~\citep{netzer2011reading}, Oxford Flowers~\citep{Nilsback08}, Deep Weeds~\citep{olsen2019deepweeds}, ImageNet~\citep{deng2009imagenet}, and CUB~\citep{welinder2010caltech} datasets which were selected to cover a diverse set of domains and challenges. In the following experiments, near-OOD datasets are those that share some degree of class overlap with the ID dataset~\citep{yang2022openood}.

\subsection{Core Results}
\label{sec:results-core}

\begin{table*}[tb]
\caption{Mean accuracy, OOD detection, and adversarial attack detection performance with standard deviation (sd) of the comparative approaches with a variety of ID and OOD datasets in order of dataset difficulty. The adversarial attack is an $L_2$-PGD attack. For compactness, we denote adversarial AUROC as AAUROC. \colorbox{lightblue}{Highlighted} cells denote the best performance for each metric. * indicates datasets classed as Near-OOD. Isolated best values arising from degenerate behaviour, such as near-universal acceptance/rejection or severely degraded ID accuracy, are not highlighted.}
\label{tab:main-results}
\centering
\resizebox{0.95\textwidth}{!}{%
\begin{tabular}{r|cccccccc|c}
\toprule
  &
  \multicolumn{1}{c}{\textbf{PostNet}} &
  \multicolumn{1}{c}{\textbf{EDL}} &
  \multicolumn{1}{c}{\textbf{I-EDL}}  &
  \multicolumn{1}{c}{\textbf{H-EDL}} &
  \multicolumn{1}{c}{\textbf{R-EDL}} &
    \textbf{RED}& \textbf{S-EDL}&\multicolumn{1}{c}{\textbf{C-EDL}} &
  \multicolumn{1}{|c}{\textbf{CLEAR}} \\
  \midrule
 Type& \multicolumn{6}{c|}{In-Training}&  \multicolumn{3}{c}{Post-Hoc}\\
    \midrule
      \multicolumn{10}{c}{\textbf{MNIST $\rightarrow$ FashionMNIST}}\\
    \midrule
    ID Acc ($\%$) $\uparrow$ &  $98.96 \pm 0.11$&  $99.26 \pm 0.07$&  $99.20 \pm 0.07$&  $98.33 \pm 0.38$&  $99.18 \pm 0.08$&   $99.01 \pm 0.12$&  $99.26 \pm 0.07$&\cellcolor{lightblue}\bm{$99.28 \pm 0.04$}&  $99.26 \pm 0.07$\\
    ID Cov ($\%$) $\uparrow$ &  $91.93 \pm 1.89$&  $89.16 \pm 2.73$&  $89.49 \pm 3.06$&  $85.63 \pm 5.84$&  $92.33 \pm 2.10$ &   $92.15 \pm 1.48$&  $89.19 \pm 2.67$&$88.29 \pm 3.54$&  \cellcolor{lightblue}\bm{$94.83 \pm 0.85$}\\
    OOD Cov ($\%$) $\downarrow$ &  $5.14 \pm 2.63$&  $12.57 \pm 8.80$&  $13.20 \pm 10.46$&  $21.84 \pm 16.94$&  $7.83 \pm 5.99$&   $3.64 \pm 1.03$&  $12.68 \pm 8.93$&$16.60 \pm 9.68$&  \cellcolor{lightblue}\bm{$1.84 \pm 0.91$}\\
    Adv Cov ($\%$) $\downarrow$ &  $86.16 \pm 10.82$&  $26.73 \pm 13.70$&  $31.36 \pm 16.33$&  $19.44 \pm 14.92$ &  $64.58 \pm 12.73$ &   $28.79 \pm 9.32$&  $23.05 \pm 13.12$&$9.20 \pm 7.30$&  \cellcolor{lightblue}\bm{$7.21 \pm 3.90$}\\
    AUROC ($\%$) $\uparrow$ &  $97.55 \pm 1.12$&  $92.54 \pm 5.70$&  $91.94 \pm 7.22$&  $86.18 \pm 12.94$&  $95.95 \pm 3.22$&   $97.34 \pm 0.85$&  $92.52 \pm 5.73$&$91.20 \pm 6.14$&  \cellcolor{lightblue}\bm{$98.54 \pm 0.36$}\\
    AAUROC ($\%$) $\uparrow$ &  $40.19 \pm 12.45$&  $88.78 \pm 5.29$&  $88.75 \pm 4.11$&  $88.59 \pm 10.79$&  $79.67 \pm 3.17$&   $77.72 \pm 8.60$&  $90.23 \pm 4.71$&$94.91 \pm 2.13$&  \cellcolor{lightblue}\bm{$97.66 \pm 0.53$}\\
    \midrule
      \multicolumn{10}{c}{\textbf{MNIST $\rightarrow$ KMNIST}}\\
    \midrule
    ID Acc ($\%$) $\uparrow$ & $98.96 \pm 0.11$& $99.26 \pm 0.07$& $99.20 \pm 0.07$& $98.33 \pm 0.38$ & $99.18 \pm 0.08$&   $99.01 \pm 0.12$& $99.26 \pm 0.07$&\cellcolor{lightblue}\bm{$99.28 \pm 0.04$}& $99.26 \pm 0.07$\\
    ID Cov ($\%$) $\uparrow$ & $93.38 \pm 0.89$& $93.93 \pm 1.16$& $93.70 \pm 0.95$& $91.40 \pm 1.94$& $92.63 \pm 0.83$&   $94.50 \pm 1.04$& $94.16 \pm 0.94$&$93.55 \pm 0.83$& \cellcolor{lightblue}\bm{$95.29 \pm 0.39$}\\
    OOD Cov ($\%$) $\downarrow$ & $3.97 \pm 0.68$& $3.58 \pm 1.11$& $3.06 \pm 0.90$& $8.00 \pm 2.55$& $3.15 \pm 0.87$&   $7.26 \pm 1.28$& $3.76 \pm 1.06$&$4.12 \pm 1.10$& \cellcolor{lightblue}\bm{$2.07 \pm 0.66$}\\
    Adv Cov ($\%$) $\downarrow$ & $83.34 \pm 6.22$& $30.08 \pm 6.78$& $30.82 \pm 4.52$& \cellcolor{lightblue}\bm{$4.69 \pm 2.12$}& $56.63 \pm 7.00$&   $51.92 \pm 15.06$& $26.41 \pm 5.86$&$6.71 \pm 2.47$ & $13.76 \pm 3.47$\\
    AUROC ($\%$) $\uparrow$ & $98.41 \pm 0.26$& $98.73 \pm 0.25$& $98.79 \pm 0.23$ & $97.18 \pm 0.98$& $98.40 \pm 0.20$&   $97.16 \pm 0.54$& $98.73 \pm 0.25$&$98.58 \pm 0.33$& \cellcolor{lightblue}\bm{$99.16 \pm 0.10$}\\
    AAUROC ($\%$) $\uparrow$ & $54.48 \pm 7.98$& $93.63 \pm 1.57$& $93.37 \pm 1.42$& \cellcolor{lightblue}\bm{$98.04 \pm 0.73$}& $87.54 \pm 2.05$&   $76.14 \pm 5.72$& $94.53 \pm 1.33$& $97.51 \pm 0.59$& $97.36 \pm 0.39$\\
    \midrule
      \multicolumn{10}{c}{\textbf{MNIST $\rightarrow$ EMNIST*}}\\
    \midrule
    ID Acc ($\%$) $\uparrow$ & $98.96 \pm 0.11$&  \cellcolor{lightblue}\bm{$99.26 \pm 0.07$}& $99.20 \pm 0.07$& $98.33 \pm 0.38$& $99.18 \pm 0.08$&   $99.01 \pm 0.12$& \cellcolor{lightblue}\bm{$99.26 \pm 0.07$}&$99.25 \pm 0.06$& \cellcolor{lightblue}\bm{$99.26 \pm 0.07$}\\
    ID Cov ($\%$) $\uparrow$ & $91.32 \pm 0.47$& $90.95 \pm 0.99$& $91.26 \pm 0.95$& $87.01 \pm 2.88$& $91.04 \pm 0.94$&   $90.51 \pm 1.59$& $91.21 \pm 0.83$&$91.18 \pm 0.75$& \cellcolor{lightblue}\bm{$93.01 \pm 1.07$}\\
    OOD Cov ($\%$) $\downarrow$ & $11.44 \pm 0.72$& $8.52 \pm 0.93$& $9.68 \pm 1.18$& $13.22 \pm 3.26$ & $9.52 \pm 0.92$&   $14.81 \pm 1.55$&  $8.70 \pm 0.85$&$10.09 \pm 0.99$&  \cellcolor{lightblue}\bm{$7.60 \pm 0.87$}\\
    Adv Cov ($\%$) $\downarrow$ & $67.98 \pm 7.06$& $9.58 \pm 4.35$& $12.17 \pm 5.25$& $7.23 \pm 3.14$& $35.93 \pm 4.13$&   $45.63 \pm 11.36$& $7.86 \pm 3.70$&\cellcolor{lightblue}\bm{$3.81 \pm 0.86$}& $3.96 \pm 1.50$\\
    AUROC ($\%$) $\uparrow$ & $95.25 \pm 0.30$& \cellcolor{lightblue}\bm{$96.47 \pm 0.26$}& $96.12 \pm 0.51$& $94.25 \pm 1.32$& $96.22 \pm 0.26$&   $91.95 \pm 1.10$& \cellcolor{lightblue}\bm{$96.47 \pm 0.26$}&$95.98 \pm 0.30$& $97.37 \pm 0.15$\\
    AAUROC ($\%$) $\uparrow$ & $65.18 \pm 7.24$& $95.53 \pm 0.81$& $95.30 \pm 0.97$& $96.37 \pm 1.18$& $90.61 \pm 1.87$&   $71.20 \pm 6.79$& $96.04 \pm 0.67$&$97.55 \pm 0.30$& \cellcolor{lightblue}\bm{$97.63 \pm 0.22$}\\
    \midrule
      \multicolumn{10}{c}{\textbf{CIFAR10 $\rightarrow$ SVHN}}\\
    \midrule
    ID Acc ($\%$) $\uparrow$ & $89.09 \pm 0.26$&  \cellcolor{lightblue}\bm{$89.76 \pm 0.19$}& $88.71 \pm 0.24$& $85.08 \pm 0.37$& $89.28 \pm 0.23$&   $88.88 \pm 0.30$& $89.66 \pm 0.11$&$90.11 \pm 0.19$& \cellcolor{lightblue}\bm{$89.76 \pm 0.19$}\\
    ID Cov ($\%$) $\uparrow$ & $79.25 \pm 1.85$& $82.45 \pm 1.09$& $84.77 \pm 1.13$& \cellcolor{lightblue}\bm{$86.72 \pm 0.97$}& $81.66 \pm 1.48$&   $79.82 \pm 1.16$& $75.60 \pm 1.55$ &$82.80 \pm 1.51$& $76.55 \pm 1.69$\\
    OOD Cov ($\%$) $\downarrow$ & $12.99 \pm 2.24$& $9.23 \pm 1.30$& \cellcolor{lightblue}\bm{$6.56 \pm 1.06$}& $9.11 \pm 1.93$& $11.81 \pm 1.91$&    $8.18 \pm 1.47$& $10.21 \pm 1.51$&$10.62 \pm 1.71$ & $13.80 \pm 1.96$\\
    Adv Cov ($\%$) $\downarrow$ & $99.92 \pm 0.04$& $99.39 \pm 0.11$& $99.25 \pm 0.45$& $50.04 \pm 12.59$& $95.84 \pm 0.71$&    $99.99 \pm 0.01$ & $99.33 \pm 0.12$&$99.28 \pm 0.11$& \cellcolor{lightblue}\bm{$36.38 \pm 8.39$}\\
    AUROC ($\%$) $\uparrow$ & $90.55 \pm 1.02$& $93.18 \pm 0.38$& $94.44 \pm 0.31$& \cellcolor{lightblue}\bm{$95.21 \pm 1.05$}& $92.39 \pm 0.59$&   $92.44 \pm 0.84$& $88.94 \pm 0.63$&$93.10 \pm 0.44$& $89.59 \pm 0.65$\\
    AAUROC ($\%$) $\uparrow$ &  $0.14 \pm 0.06$& $1.15 \pm 0.16$& $23.65 \pm 1.90$& $61.62 \pm 11.10$& $15.14 \pm 1.60$&   $33.84 \pm 1.10$& $3.72 \pm 0.50$&$10.97 \pm 0.68$& \cellcolor{lightblue}\bm{$77.45 \pm 0.42$}\\
    \midrule
      \multicolumn{10}{c}{\textbf{CIFAR10 $\rightarrow$ CIFAR100*}}\\
    \midrule
    ID Acc ($\%$) $\uparrow$ & $89.09 \pm 0.26$& $89.76 \pm 0.19$& $88.71 \pm 0.24$& $85.08 \pm 0.37$& $89.28 \pm 0.23$&   $88.88 \pm 0.30$& $89.61 \pm 0.16$&\cellcolor{lightblue}\bm{$90.12 \pm 0.31$}& $89.76 \pm 0.19$\\
    ID Cov ($\%$) $\uparrow$ & $70.78 \pm 3.35$ & $73.39 \pm 1.42$& $73.66 \pm 1.75$& \cellcolor{lightblue}\bm{$74.58 \pm 2.80$}& $69.85 \pm 3.47$&   $70.80 \pm 1.64$& $73.45 \pm 2.65$&$73.55 \pm 1.34$& $73.22 \pm 1.71$\\
    OOD Cov ($\%$) $\downarrow$ & $21.38 \pm 3.01$& \cellcolor{lightblue}\bm{$15.58 \pm 1.42$}& $16.50 \pm 1.49$& $24.57 \pm 3.00$& $16.89 \pm 3.79$&   $15.66 \pm 1.69$& $15.87 \pm 2.66$&$15.68 \pm 1.56$& $20.08 \pm 1.66$\\
    Adv Cov ($\%$) $\downarrow$ & $99.47 \pm 0.14$ & $98.60 \pm 0.15$ & $83.56 \pm 3.28$ & \cellcolor{lightblue}\bm{$39.02 \pm 8.94$}& $92.97 \pm 0.60$ & $99.95 \pm 0.03$ & $98.54 \pm 0.15$ & $97.96 \pm 0.13$ & $43.43 \pm 5.59$ \\
    AUROC ($\%$) $\uparrow$ & $81.84 \pm 0.52$& $86.55 \pm 0.23$& $86.29 \pm 0.28$& $82.20 \pm 0.53$& $84.63 \pm 0.20$&   $85.10 \pm 0.24$& $86.58 \pm 0.23$&\cellcolor{lightblue}\bm{$86.59 \pm 0.24$}& $84.83 \pm 0.20$\\
    AAUROC ($\%$) $\uparrow$ & $0.64 \pm 0.12$ & $2.01 \pm 0.16$ & $22.04 \pm 1.24$ & $64.68 \pm 7.91$ & $13.54 \pm 0.84$ & $37.52 \pm 0.70$ & $2.83 \pm 0.19$ & $12.22 \pm 0.30$ & \cellcolor{lightblue}\bm{$71.93 \pm 0.45$}\\
    \midrule
     \multicolumn{10}{c}{\textbf{Oxford Flowers (low-shot) $\rightarrow$ Deep Weeds}}\\
    \midrule
    ID Acc ($\%$) $\uparrow$ & $5.92 \pm 1.49$ & $80.54 \pm 0.33$& $81.09 \pm 0.86$& $74.31 \pm 1.22$& \cellcolor{lightblue}\bm{$85.18 \pm 0.85$}&   $83.27 \pm 0.33$& $79.41 \pm 0.31$&$80.63 \pm 0.31$& $80.54 \pm 0.32$\\
    ID Cov ($\%$) $\uparrow$ & $14.30 \pm 9.90$ & $64.00 \pm 5.55$& $82.62 \pm 4.11$& \cellcolor{lightblue}\bm{$90.58 \pm 1.89$}& $89.11 \pm 2.28$&   $88.43 \pm 2.84$& $64.63 \pm 4.24$&$71.41 \pm 4.69$& $86.59 \pm 1.52$\\
    OOD Cov ($\%$) $\downarrow$ & $8.47 \pm 7.22$ & $18.78 \pm 4.68$& $8.79 \pm 3.91$& $13.14 \pm 2.04$& $6.24 \pm 1.35$&   $7.41 \pm 2.65$& $20.57 \pm 4.75$&$15.97 \pm 3.41$& \cellcolor{lightblue}\bm{$5.09 \pm 2.12$}\\
    Adv Cov ($\%$) $\downarrow$ & $0.00 \pm 0.01$ & $43.15 \pm 7.80$ & $28.53 \pm 6.32$& $32.09 \pm 5.39$& $36.13 \pm 3.48$&   $79.03 \pm 3.29$& $25.55 \pm 6.52$&$11.80 \pm 3.16$& \cellcolor{lightblue}\bm{$6.77 \pm 2.88$}\\
    AUROC ($\%$) $\uparrow$ & $39.16 \pm 7.07$ & $80.11 \pm 4.12$& $93.07 \pm 2.12$& $95.60 \pm 0.93$& \cellcolor{lightblue}\bm{$97.48 \pm 0.79$}&   $96.66 \pm 0.89$& $79.38 \pm 4.21$&$85.47 \pm 2.95$& $95.57 \pm 0.62$\\
    AAUROC ($\%$) $\uparrow$ & $89.35 \pm 2.71$ & $64.91 \pm 2.65$& $84.43 \pm 2.34$& $87.99 \pm 2.43$& $81.49 \pm 2.85$&   $47.68 \pm 1.76$& $76.34 \pm 2.40$&$86.90 \pm 1.43$& \cellcolor{lightblue}\bm{$95.60 \pm 0.84$}\\
    \midrule
     \multicolumn{10}{c}{\textbf{ImageNet $\rightarrow$ CUB}}\\
    \midrule
    ID Acc ($\%$) $\uparrow$ & $37.69 \pm 0.25$ & \cellcolor{lightblue}\bm{$79.99 \pm 0.03$} & $79.53 \pm 0.01$ & $78.71 \pm 0.00$ & $77.14 \pm 0.03$ & $79.92 \pm 0.03$ & $75.60 \pm 0.02$ & $76.09 \pm 0.05$ & $75.86 \pm 0.04$ \\
    ID Cov ($\%$) $\uparrow$ & $15.97 \pm 1.31$ & $74.02 \pm 1.98$ & \cellcolor{lightblue}\bm{$99.97 \pm 0.04$} & $49.05 \pm 0.00$ & $60.93 \pm 0.12$ & $75.68 \pm 1.48$ & $0.03 \pm 0.02$ & $0.03 \pm 0.02$ & $80.54 \pm 0.39$ \\
    OOD Cov ($\%$) $\downarrow$ & $11.55 \pm 2.13$ & $33.68 \pm 2.42$ & $99.93 \pm 0.03$ & $31.69 \pm 0.00$ & $41.51 \pm 0.11$ & $36.49 \pm 2.01$ & $0.17 \pm 0.11$ & $0.12 \pm 0.06$ & \cellcolor{lightblue}\bm{$23.67 \pm 0.55$} \\
    Adv Cov ($\%$) $\downarrow$ & $96.61 \pm 0.37$ & $93.50 \pm 0.44$ & $98.02 \pm 0.88$ & $15.48 \pm 0.22$ & \cellcolor{lightblue}\bm{$8.84 \pm 0.19$} & $95.42 \pm 1.82$ & $0.34 \pm 0.18$ & $0.14 \pm 0.07$ & $10.84 \pm 0.36$ \\
    AUROC ($\%$) $\uparrow$ & $36.97 \pm 1.41$ & $76.34 \pm 0.45$ & $46.26 \pm 0.05$ & $57.87 \pm 0.00$ & $63.05 \pm 0.06$ & $72.87 \pm 0.29$ & $36.73 \pm 0.21$ & $36.52 \pm 0.21$ & \cellcolor{lightblue}\bm{$84.63 \pm 0.02$} \\
    AAUROC ($\%$) $\uparrow$ & $2.74 \pm 0.30$ & $35.28 \pm 0.26$ & $35.08 \pm 0.10$ & $68.98 \pm 0.05$ & $87.60 \pm 0.08$ & $41.04 \pm 0.13$ & $32.89 \pm 0.21$ & $33.82 \pm 0.26$ & \cellcolor{lightblue}\bm{$92.61 \pm 0.03$} \\
    \bottomrule
\end{tabular}
}
\end{table*}

Table~\ref{tab:main-results} shows that CLEAR provides consistently strong uncertainty performance across the core classification benchmarks while preserving the predictive performance of the underlying evidential model. This behaviour is also reflected in Figure~\ref{fig:core-results}, where CLEAR generally occupies the desirable operating region across both OOD and adversarial objectives. On MNIST$\rightarrow$FashionMNIST, CLEAR achieves the highest ID coverage ($94.83\%$), lowest OOD and adversarial coverage ($1.84\%$ and $7.21\%$), and highest AUROC and AAUROC ($98.54\%$ and $97.66\%$). This performance largely persists across the remaining MNIST settings: on MNIST$\rightarrow$KMNIST, CLEAR achieves the highest ID coverage, lowest OOD coverage, and highest AUROC, while retaining an AAUROC of $97.36\%$; on the near-OOD MNIST$\rightarrow$EMNIST setting, it again achieves the strongest ID/OOD coverage together with the highest AUROC and AAUROC.

The advantage of CLEAR becomes particularly apparent on the more challenging CIFAR10 experiments. On CIFAR10$\rightarrow$SVHN, CLEAR achieves the lowest adversarial coverage ($36.38\%$) and highest AAUROC ($77.45\%$), while on CIFAR10$\rightarrow$CIFAR100 it again achieves the highest AAUROC ($71.93\%$) while remaining competitive in ID retention and OOD detection. Figure~\ref{fig:core-results} shows that CLEAR therefore maintains strong performance across both uncertainty objectives rather than favouring one form of distribution shift. Cases where CLEAR is weaker on an individual metric, including adversarial coverage on MNIST$\rightarrow$KMNIST and OOD separation on the CIFAR10 benchmarks, are examined in more detail in Appendix~\ref{sec:appendix-Additional Discussion}.

This pattern continues on the larger and more difficult benchmarks. On Oxford Flowers$\rightarrow$DeepWeeds, CLEAR achieves the lowest non-degenerate OOD and adversarial coverage ($5.09\%$ and $6.77\%$) and the highest AAUROC ($95.60\%$), while retaining an AUROC of $95.57\%$. On ImageNet$\rightarrow$CUB, a substantially harder setting due to ImageNet's large-scale 1000-class output space and highly diverse visual representation, CLEAR achieves the highest AUROC and AAUROC, at $84.63\%$ and $92.61\%$, while retaining $80.54\%$ ID coverage. The degenerate operating points observed for several comparative methods on these harder benchmarks are also discussed in Appendix~\ref{sec:appendix-Additional Discussion}. Together, these results show that CLEAR remains effective as dataset scale and difficulty increase.

\begin{wraptable}[30]{l}{0.46\textwidth}
\caption{Mean accuracy, OOD and Adv. detection performance with standard deviation on ID and OOD regression datasets. Attacks use $L_\infty$-PGD; AAUROC denotes adversarial AUROC. \colorbox{lightblue}{Highlighted} cells indicate the best result per metric, and * marks Near-OOD datasets.}
\label{tab:regression-results}
\centering
\resizebox{\linewidth}{!}{%
\begin{tabular}{r|cc|c}
\toprule
  &
  \multicolumn{1}{c}{\textbf{DER}} &
  \multicolumn{1}{c}{\textbf{UR-ERN}} &
  \multicolumn{1}{|c}{\textbf{CLEAR}} \\
    \midrule
      \multicolumn{4}{c}{\textbf{California Housing $\rightarrow$ California Housing (Price Extremes)*}}\\
    \midrule
    ID MAE $\downarrow$& \cellcolor{lightblue}\bm{$\$25.26k \pm 0.61$}& $\$25.33k \pm 0.56$&  \cellcolor{lightblue}\bm{$\$25.26k \pm 0.61$}\\
    ID RMSE $\downarrow$& \cellcolor{lightblue}\bm{$\$32.84k \pm 0.60$}& $\$32.95k \pm 0.66$&\cellcolor{lightblue}\bm{$\$32.84k \pm 0.60$}\\
    ID Cov ($\%$) $\uparrow$ & \cellcolor{lightblue}\bm{$93.01 \pm 5.53$}& $85.05 \pm 28.54$&  $57.18 \pm 2.98$\\
    OOD Cov ($\%$) $\downarrow$ & $91.60 \pm 5.57$& $85.05 \pm 28.54$&  \cellcolor{lightblue}\bm{$26.34 \pm 4.11$}\\
    Adv Cov ($\%$) $\downarrow$ & $99.90 \pm 0.19$& $89.89 \pm 29.96$&  \cellcolor{lightblue}\bm{$1.05 \pm 1.26$}\\
    AUROC ($\%$) $\uparrow$ & $41.62 \pm 3.15$ & $40.36 \pm 3.32$&  \cellcolor{lightblue}\bm{$71.01 \pm 1.99$}\\
    AAUROC ($\%$) $\uparrow$ & $2.48 \pm 1.77$& $2.32 \pm 1.63$&  \cellcolor{lightblue}\bm{$89.75 \pm 2.42$}\\
    \midrule
      \multicolumn{4}{c}{\textbf{Wine Quality (White Wine) $\rightarrow$ Wine Quality (Red Wine)*}}\\
    \midrule
    ID MAE $\downarrow$& \cellcolor{lightblue}\bm{$0.56 \pm 0.01$}& \cellcolor{lightblue}\bm{$0.56 \pm 0.01$}&  \cellcolor{lightblue}\bm{$0.56 \pm 0.01$}\\
    ID RMSE $\downarrow$& \cellcolor{lightblue}\bm{$0.72 \pm 0.01$}& \cellcolor{lightblue}\bm{$0.72 \pm 0.01$}&\cellcolor{lightblue}\bm{$0.72 \pm 0.01$}\\
    ID Cov ($\%$) $\uparrow$ & $82.30 \pm 10.69$& \cellcolor{lightblue}\bm{$85.37 \pm 10.76$}&  $77.54 \pm 4.95$\\
    OOD Cov ($\%$) $\downarrow$ & $68.17 \pm 17.13$& $70.67 \pm 19.92$ &  \cellcolor{lightblue}\bm{$14.75 \pm 3.76$}\\
    Adv Cov ($\%$) $\downarrow$ & $68.17 \pm 17.13$& $80.79 \pm 15.13$&  \cellcolor{lightblue}\bm{$12.25 \pm 3.64$}\\
    AUROC ($\%$) $\uparrow$ & $54.55 \pm 10.86$& $54.76 \pm 10.80$&  \cellcolor{lightblue}\bm{$89.10 \pm 1.45$}\\
    AAUROC ($\%$) $\uparrow$ & $41.40 \pm 10.13$& $42.01 \pm 10.57$&  \cellcolor{lightblue}\bm{$89.99 \pm 1.31$}\\
    \midrule
      \multicolumn{4}{c}{\textbf{Airfoil Self-Noise $\rightarrow$ Airfoil Self-Noise (Covariant Extremes- Frequency)*}}\\
    \midrule
    ID MAE $\downarrow$& $4.12 \pm 0.80$ & \cellcolor{lightblue}\bm{$3.92 \pm 0.98$}& $4.12 \pm 0.80$ \\
    ID RMSE $\downarrow$& $4.88 \pm 0.78$ & \cellcolor{lightblue}\bm{$4.74 \pm 0.94$}& $4.88 \pm 0.78$ \\
    ID Cov ($\%$) $\uparrow$ & $0.00 \pm 0.00$ & $0.00 \pm 0.00$ & \cellcolor{lightblue}\bm{$72.87 \pm 5.79$}\\
    OOD Cov ($\%$) $\downarrow$ & $0.00 \pm 0.00$ & $0.00 \pm 0.00$ & \cellcolor{lightblue}\bm{$20.46 \pm 5.84$}\\
    Adv Cov ($\%$) $\downarrow$ & $0.00 \pm 0.00$ & $0.00 \pm 0.00$ & \cellcolor{lightblue}\bm{$0.62 \pm 1.85$}\\
    AUROC ($\%$) $\uparrow$ & $14.91 \pm 3.35$ & $15.15 \pm 3.46$ & \cellcolor{lightblue}\bm{$82.81 \pm 3.78$}\\
    AAUROC ($\%$) $\uparrow$ & $0.54 \pm 0.38$ & $0.56 \pm 0.41$ & \cellcolor{lightblue}\bm{$94.79 \pm 2.21$}\\
    \midrule
      \multicolumn{4}{c}{\textbf{UTKFace $\rightarrow$ Oxford Flowers}}\\
    \midrule
    ID MAE $\downarrow$& $7.91 \pm 0.14$& \cellcolor{lightblue}\bm{$7.89 \pm 0.14$}&  $7.99 \pm 0.18$\\
    ID RMSE $\downarrow$& $10.41 \pm 0.23$& \cellcolor{lightblue}\bm{$10.38 \pm 0.19$}&$10.48 \pm 0.25$\\
    ID Cov ($\%$) $\uparrow$ & \cellcolor{lightblue}\bm{$81.82 \pm 12.89$}& $81.08 \pm 10.14$&  $76.81 \pm 3.05$\\
    OOD Cov ($\%$) $\downarrow$ & $42.42 \pm 12.72$& $41.55 \pm 9.49$&  \cellcolor{lightblue}\bm{$14.98 \pm 3.23$}\\
    Adv Cov ($\%$) $\downarrow$ & $49.82 \pm 12.68$& $49.31 \pm 10.53$&  \cellcolor{lightblue}\bm{$18.09 \pm 3.33$}\\
    AUROC ($\%$) $\uparrow$ & $73.99 \pm 9.06$& $74.57 \pm 8.15$&  \cellcolor{lightblue}\bm{$87.53 \pm 3.15$}\\
    AAUROC ($\%$) $\uparrow$ & $64.38 \pm 11.62$& $64.86 \pm 10.35$&  \cellcolor{lightblue}\bm{$85.38 \pm 3.53$}\\
    \midrule
      \multicolumn{4}{c}{\textbf{UTKFace $\rightarrow$ UTKFace (Age Extremes)*}}\\
    \midrule
    ID MAE $\downarrow$& \cellcolor{lightblue}\bm{$7.98 \pm 0.13$}& $8.00 \pm 0.18$&  \cellcolor{lightblue}\bm{$7.98 \pm 0.13$}\\
    ID RMSE $\downarrow$& \cellcolor{lightblue}\bm{$10.47 \pm 0.17$}& $10.50 \pm 0.25$& \cellcolor{lightblue}\bm{$10.47 \pm 0.17$}\\
    ID Cov ($\%$) $\uparrow$ & $56.81 \pm 10.16$& \cellcolor{lightblue}\bm{$57.66 \pm 7.94$}&  $53.15 \pm 4.52$\\
    OOD Cov ($\%$) $\downarrow$ & $39.74 \pm 8.91$& $41.01 \pm 8.47$&  \cellcolor{lightblue}\bm{$37.31 \pm 3.20$}\\
    Adv Cov ($\%$) $\downarrow$ & $61.32 \pm 5.88$& $62.59 \pm 5.28$&  \cellcolor{lightblue}\bm{$44.01 \pm 2.91$}\\
    AUROC ($\%$) $\uparrow$ & \cellcolor{lightblue}\bm{$61.63 \pm 1.95$}& $61.31 \pm 1.91$&  $60.77 \pm 1.88$\\
    AAUROC ($\%$) $\uparrow$ & $45.64 \pm 2.88$& $45.01 \pm 2.99$&  \cellcolor{lightblue}\bm{$53.32 \pm 2.41$}\\
    \bottomrule
\end{tabular}
}
\end{wraptable}

Overall, CLEAR either achieves the strongest result or remains competitive across ID retention, OOD detection, and adversarial detection, without consistently sacrificing one objective for another.

\subsection{Regression Results}
\label{sec:results-regression}
Across the regression benchmarks, CLEAR consistently improves separation between ID predictions and both OOD and adversarial inputs while preserving comparable predictive error. On California Housing, CLEAR reduces OOD coverage from $91.60\%$ with DER to $26.34\%$ and adversarial coverage from $99.90\%$ to $1.05\%$, while increasing AUROC and AAUROC to $71.01\%$ and $89.75\%$, respectively. This is accompanied by lower ID coverage ($57.18\%$), a trade-off also observed more mildly on several other regression benchmarks and discussed further in Appendix~\ref{sec:appendix-Additional Discussion}. Importantly, the underlying regression prediction is unchanged, so the stronger rejection does not correspond to degraded predictive accuracy. Similar improvements are observed on Wine Quality, where CLEAR achieves an AUROC of $89.10\%$ and AAUROC of $89.99\%$ without changing MAE or RMSE.

The Airfoil experiment provides an especially clear example under covariate shift, where OOD samples are defined by extreme values of the input frequency variable rather than by extremes of the regression target. This makes the shift difficult for uncertainty estimates that do not adequately capture changes in the input representation: DER and UR-ERN retain $0\%$ of both ID and OOD samples under the selected threshold, indicating little useful separation between the two distributions. In contrast, CLEAR retains $72.87\%$ ID coverage while reducing OOD and adversarial coverage to $20.46\%$ and $0.62\%$, respectively, with an AAUROC of $94.79\%$. This suggests that CLEAR's latent-space conflict is sensitive to changes in the covariate distribution even when the prediction target itself is not the variable used to define the shift.

This behaviour remains consistent on the UTKFace experiments. For UTKFace$\rightarrow$Oxford Flowers, CLEAR reduces OOD and adversarial coverage to $14.98\%$ and $18.09\%$, while increasing AUROC and AAUROC to $87.53\%$ and $85.38\%$. The subtler UTKFace age-extremes shift is more competitive, because the OOD samples remain within the same face domain and differ only in their target distribution, resulting in less distinct separation between their latent representations and those of ID samples. Consequently, CLEAR achieves an AUROC of $60.77\%$ remaining close to the best value of $61.63\%$, while still improving adversarial coverage to $44.01\%$ and AAUROC to $53.32\%$. Overall, these results show that CLEAR generalises its uncertainty improvements beyond classification, maintaining strong predictive performance while improving robustness to both distribution shift and adversarial perturbation.

\subsection{Object Detection Results}
\label{sec:objdet}

\begin{wrapfigure}[23]{r}{0.4\textwidth}
    \centering
    \includegraphics[width=\linewidth]{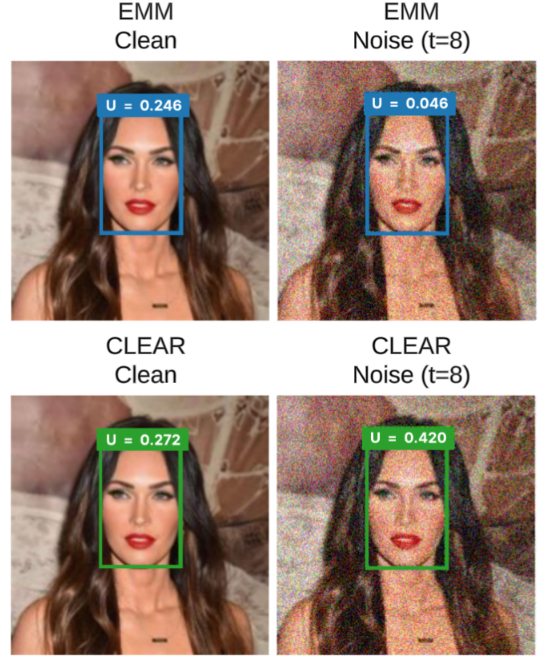}
    \caption{Qualitative object detection comparison of EMM and CLEAR uncertainty under a covariate Gaussian noise on CelebA.}
    \label{fig:clear_objdet_example}
\end{wrapfigure}

CLEAR also generalises effectively to object detection, where uncertainty must reflect not only whether a detection is reliable but also the quality of its localisation. As shown in Table~\ref{tab:clear_clean_uq}, Appendix~\ref{sec:appendix-Full Object Detection Results}, CLEAR improves every evaluated uncertainty metric over EMM on clean CelebA detections, increasing AUROC from $50.3\%$ to $58.5\%$ and AUPRC from $12.5\%$ to $21.1\%$, while reducing FPR95 from $90.7\%$ to $82.4\%$ and ECE from $84.4\%$ to $75.0\%$. Importantly, the correlation between uncertainty and localisation error increases from approximately zero ($r=0.003$) for EMM to $r=0.168$ for CLEAR, indicating that poorly localised detections are more consistently assigned greater uncertainty. This is important for object detection, where a confident detection does not necessarily imply that its predicted bounding box is accurately localised.

This localisation-aware behaviour persists under covariate corruption. Table~\ref{tab:clear_corruption_pearson}, Appendix~\ref{sec:appendix-Full Object Detection Results}, shows that across brightness, blur, noise, compression, and resize/crop shifts, CLEAR produces a stronger positive association between predictive uncertainty and localisation error than EMM in every case, increasing the overall Pearson correlation from $0.010$ to $0.049$. Although the absolute correlations remain modest, the consistent improvement across all five shifts indicates that CLEAR better reflects degradation in bounding-box quality as the input distribution changes. This trend is further illustrated in Figure~\ref{fig:clear_corruption_ranking_pearson}, where CLEAR maintains a stronger uncertainty-localisation relationship across the evaluated corruptions. Together, these results demonstrate that CLEAR's latent-conflict mechanism transfers beyond classification and regression to structured localisation outputs while producing uncertainty that is more informative of detection quality. Full training details for this setup are described in Appendix~\ref{sec:appendix-Models}.

\subsection{Further Experiments and Ablation}
\label{sec:further-exp}
An extensive set of additional experiments is provided in the Appendix, including corrupted benchmarks (Appendix~\ref{sec:appendix-Corrupted Benchmarks}) and additional uncertainty metrics (Appendix~\ref{sec:appendix-Extra Metrics Analysis}). We then examine CLEAR through hyperparameter ablations (Appendix~\ref{sec:appendix-Ablation Analysis}), uncertainty-threshold analysis (Appendix~\ref{sec:appendix-Threshold Analysis}), adversarial attack analysis (Appendix~\ref{sec:appendix-Adversarial Attack Analysis}), and extended object-detection results (Appendix~\ref{sec:appendix-Full Object Detection Results}). Finally, we investigate the latent mechanism through held-out validation of the calibrated geometry (Appendix~\ref{sec:appendix-Held-Out Analysis of the Latent Whitening Geometry}), sensitivity to latent representation depth (Appendix~\ref{sec:appendix-Sensitivity to Latent Representation Depth}), and qualitative visualisation of the generated latent perturbation views (Appendix~\ref{sec:appendix-Qualitative Analysis of Latent Perturbation Views}). Across these analyses, CLEAR maintains strong uncertainty performance under corruption and increasing attack strength, remains broadly stable across its hyperparameters and thresholding method, and exhibits consistent behaviour across different latent representations.

\section{Conclusions}
\label{sec:Conclusions}
We introduced CLEAR, a lightweight post-hoc method for improving evidential uncertainty without retraining or altering the underlying prediction. Across a broad set of datasets spanning classification, regression, and object detection, CLEAR consistently improves uncertainty quality while preserving predictive performance. A thorough evaluation across diverse OOD settings, covariate shifts, adversarial attacks, and uncertainty thresholds demonstrates that CLEAR remains effective under substantially different forms of distributional and adversarial challenge. Importantly, CLEAR improves complementary uncertainty objectives jointly, avoiding the pronounced trade-offs exhibited by several comparative methods, while introducing only modest additional inference overhead. These results demonstrate that latent conflict provides a practical and broadly applicable signal for identifying unsupported evidential confidence. Future work will investigate evaluating CLEAR on denser structured prediction tasks, such as semantic segmentation, where uncertainty must be estimated over spatially distributed outputs.


\bibliography{iclr2027_conference}
\bibliographystyle{iclr2027_conference}


\setlength{\parindent}{0pt}
\setlength{\parskip}{0.5em}
\newpage
\onecolumn
\appendix

\setcounter{secnumdepth}{2}
\setcounter{tocdepth}{2}

\startcontents[appendices]

\section*{Appendix Contents}
\printcontents[appendices]{}{1}{}


\section{Implementation Details}
\label{sec:appendix-Implementation Details}

Algorithms~\ref{alg:clear_calibration} and~\ref{alg:clear_inference} summarise the calibration and inference stages of CLEAR. The pretrained evidential model remains frozen and in evaluation mode throughout both stages. Calibration is performed once, after which the group statistics and global reference conflict are reused for subsequent inputs.

\begin{algorithm}[tb]
\caption{CLEAR calibration}
\label{alg:clear_calibration}
\small
\begin{algorithmic}[1]
\Require Frozen evidential model; calibration set $\mathcal{D}_{\mathrm{cal}}$
\Require Latent views $V$, perturbation magnitude $\tau$,
         reference draws $T$, stabilisation terms $\lambda_g$,
         regression bins $B$
\Ensure Group statistics $\{\mu_g,L_g,\rho_g\}_g$, reference $r$,
        and bin map $b(\cdot)$ for regression

\If{regression}
    \State Construct $b(\cdot)$ using $B$ target-quantile bins
           from $\mathcal{D}_{\mathrm{cal}}$.
\EndIf
\ForAll{$(x_i,y_i)\in\mathcal{D}_{\mathrm{cal}}$}
    \State Obtain $h_i=h(x_i)$ and the original evidential output.
    \State $g_i\gets y_i$ for classification, or $g_i\gets b(y_i)$
           for regression.
\EndFor
\ForAll{output groups $g$}
    \State $N_g\gets|\{i:g_i=g\}|$.
    \State $\mu_g\gets\frac{1}{N_g}\sum_{i:g_i=g}h_i$;
           $\Sigma_g\gets\operatorname{Cov}(h_i\mid g_i=g)$.
    \State $\widetilde{\Sigma}_g\gets\Sigma_g+\lambda_g I$.
    \State Compute $L_g$ such that
           $L_gL_g^\top=\widetilde{\Sigma}_g$.
    \State $\rho_g\gets
           \sqrt{\frac{d}{\left\|L_g^{-1}\right\|_F^2}}$.
\EndFor
\ForAll{$(x_i,y_i)\in\mathcal{D}_{\mathrm{cal}}$}
    \State $\hat g\gets\arg\max_k\hat p_k(x_i)$ for classification,
           or $\hat g\gets b(\gamma(x_i))$ for regression.
    \For{$t=1,\ldots,T$}
        \State Compute $C_{\mathrm{raw}}^{(t)}(x_i)$ using
               lines~\ref{algline:clear_views}--\ref{algline:clear_raw}
               of Algorithm~\ref{alg:clear_inference}
               for $x_i$, with $h=h_i$ and fresh perturbations.
    \EndFor
    \State $\overline{C}_{\mathrm{raw}}(x_i)
           \gets\frac{1}{T}\sum_{t=1}^{T}
           C_{\mathrm{raw}}^{(t)}(x_i)$.
\EndFor
\State $r\gets\operatorname{median}_{x_i\in\mathcal{D}_{\mathrm{cal}}}
       \overline{C}_{\mathrm{raw}}(x_i)$.
\State \Return $\{\mu_g,L_g,\rho_g\}_g$, $r$,
       and $b(\cdot)$ for regression.
\end{algorithmic}
\end{algorithm}

During calibration, class labels or target-quantile bins determine the groups used to estimate $\mu_g$, $L_g$, and $\rho_g$. Reference conflict is then evaluated relative to the group $\hat g$ selected by each example's original prediction, matching the inference procedure. Each calibration example receives $T$ independent perturbation draws, each containing $V$ latent views. The raw conflict scores are averaged per example before taking the median across $\mathcal{D}_{\mathrm{cal}}$ to obtain $r$.

\begin{algorithm}[tb]
\caption{CLEAR inference}
\label{alg:clear_inference}
\small
\begin{algorithmic}[1]
\Require Frozen evidential model; input $x$;
         calibrated $\{\mu_g,L_g,\rho_g\}_g$ and $r$
\Require Latent views $V$, perturbation magnitude $\tau$,
         adjustment strength $\delta$;
         bin map $b(\cdot)$ for regression
\Ensure Adjusted evidential output

\State Obtain $h=h(x)$ and the original evidential output
       in one forward pass.
\State $\hat g\gets\arg\max_k\hat p_k(x)$ for classification,
       or $\hat g\gets b(\gamma(x))$ for regression.
\For{$v=1,\ldots,V$}\label{algline:clear_views}
    \State Sample $\eta_v\sim\mathcal{N}(0,I)$ independently.
    \State $h_v\gets h+\tau\rho_{\hat g}\eta_v$.
    \State $q_v\gets L_{\hat g}^{-1}(h_v-\mu_{\hat g})$.
\EndFor
\State $Q\gets[q_1,\ldots,q_V]^\top\in\mathbb{R}^{V\times d}$.
\State $\mathcal{P}_V\gets\{(i,j):1\leq i<j\leq V\}$.
\ForAll{$(i,j)\in\mathcal{P}_V$}
    \State $C_{ij}(x)\gets
           \frac{1}{d(d-1)}
           \sum_{a\neq b}q_{i,a}^{2}q_{j,b}^{2}$.
\EndFor
\State $C_{\mathrm{raw}}(x)\gets
       \frac{1}{|\mathcal{P}_V|}
       \sum_{(i,j)\in\mathcal{P}_V}C_{ij}(x)$.
       \label{algline:clear_raw}
\State $C(x)\gets\max\left(
       0,\frac{C_{\mathrm{raw}}(x)-r}
              {C_{\mathrm{raw}}(x)+r}\right)$.
\State $s(x)\gets\exp(-\delta C(x))$.
\If{classification}
    \State $e'(x)\gets s(x)e(x)$.
    \State $\alpha'(x)\gets e'(x)+1$.
    \State \Return $\alpha'(x)$.
\Else
    \State $\nu'(x)\gets s(x)\nu(x)$.
    \State $\alpha'(x)\gets1+s(x)(\alpha(x)-1)$.
    \State \Return
           $(\gamma(x),\nu'(x),\alpha'(x),\beta(x))$.
\EndIf
\end{algorithmic}
\end{algorithm}

At inference, a single forward pass produces the latent representation and original evidential output. Each input then receives one set of $V$ latent views, with no additional evaluations of the backbone or evidential head. Whitening is implemented using triangular solves with the stored Cholesky factor $L_{\hat g}$. Pairwise conflict is computed using vector reductions, avoiding explicit construction of cross-dimensional interaction matrices. The resulting $C_{\mathrm{raw}}(x)$ is mapped to $C(x)$ and used to obtain the evidence scaling factor $s(x)$. The final adjustment preserves the predicted class in classification and the predictive mean $\gamma(x)$ in regression.


\section{Theoretical Analysis}
\label{sec:appendix-Theoretical Analysis}
This appendix provides the theoretical analysis to complement CLEAR presented in Section~\ref{sec:clear}.

\subsection{Proposition 1}

\textbf{Proposition 1.} 
CLEAR preserves the base prediction from the underlying model while reducing evidential strength in proportion to the latent conflict. In classification:
\begin{equation}
\text{arg}\max_k \hat{p}'_k(x) = \text{arg}\max_k \hat{p}_k(x), \qquad u'(x) \geq u(x)
\end{equation}
while in regression:
\begin{equation}
\mathbb{E}'[y] = \mathbb{E}[y] = \gamma(x) \\
u'_{\text{alea}}(x) = \exp(\delta C(x)) u_{\text{alea}}(x), \qquad u'_{\text{epi}}(x) = \exp(2\delta C(x)) u_{\text{epi}}(x)
\end{equation}

\begin{proof}
We first consider the classification setting, where CLEAR rescales the original evidence:
\begin{equation}
e'(x) = s(x)e(x)
\end{equation}
such that the Dirichlet parameters are also adjusted:
\begin{equation}
\alpha'_k(x) = 1 + s(x)e_ k(x)
\end{equation}
For any two classes $j$ and $k$:
\begin{equation}
\alpha'_j(x) - \alpha'_k(x) = s(x)(e_j(x) - e_k(x))
\end{equation}
and since $s(x) > 0$, multiplication by $s(x)$preserves the ordering of the original evidence terms. Moreover, because:
\begin{equation}
\alpha_j(x) - \alpha_k(x) = e_j(x) - e_k(x)
\end{equation}
the ordering of the original and adjusted Dirichlet parameters is identical. The predictive probabilities in both cases are obtained by normalising these parameters by a positive quantity. Therefore:
\begin{equation}
\text{arg}\max_k \hat{p}'_k(x) = \text{arg}\max_k \hat{p}_k(x)
\end{equation}
Thus, the CLEAR adjustment cannot change the predicted class. Given the original total evidential strength is:
\begin{equation}
S(x) = \sum^K_{k=1}\alpha_k(x) = K + \sum^K_{k=1}e_k(x)
\end{equation}
and the adjusted total strength is:
\begin{equation}
\begin{aligned}
S'(x) &= \sum^K_{k=1}\alpha'_k(x) \\
 &= K + s(x) \sum^K_{k=1}e_k(x) \\
 &= K + s(x)(S_(x) - K)
\end{aligned}
\end{equation}
Because $0 < s(x) \leq 1$ and $S(x) - K \geq 0$:
\begin{equation}
S'(x) \leq S(x)
\end{equation}
Using the scalar uncertainty definition from Equation 3:
\begin{equation}
u'(x) = \frac{K}{S'(x)} = \frac{K}{K + s(x)(S(x) - K)}
\end{equation}
It follows directly that:
\begin{equation}
u'(x) \geq \frac{K}{S(x)} = u(x)
\end{equation}
Hence, CLEAR preserves the predicted class while reducing the total evidential strength, therefore increasing, or leaving unchanged, the corresponding scalar uncertainty. For completeness, this exact adjustment can also be expressed as a contraction of the original predictive distribution towards the uniform distribution. Let:
\begin{equation}
\lambda(x) = \frac{s(x)S(x)}{K+s(x)(S(x)-K)}
\end{equation}
Since, $0 < s(x) \leq 1$, it follows that $0 < \lambda(x) \leq 1$, and the adjusted predictive probability satisfies:
\begin{equation}
\hat{p}'_k(x) = \lambda(x)\hat{p}_k(x) + (1-\lambda(x))\frac{1}{K}
\end{equation}
Thus, increasing conflict contracts the predictive probabilities towards the uniform distribution without changing their ordering.

We now consider the regression setting. From Equation 21, CLEAR leaves $\gamma(x)$ and $\beta(x)$ unchanged and applies:
\begin{equation}
\nu'(x) = s(x)\nu(x), \qquad \alpha'(x) = 1 + s(x)(\alpha(x) -1)
\end{equation}
Since the predictive mean in Equation 6 is:
\begin{equation}
\mathbb{E}[y] = \gamma(x)
\end{equation}
and CLEAR does not modify $\gamma(x)$:
\begin{equation}
\mathbb{E}'[y] = \mathbb{E}[y] = \gamma(x)
\end{equation}
For the aleatoric uncertainty defined in Equation 7:
\begin{equation}
\begin{aligned}
u'_{alea}(x) &= \frac{\beta(x)}{\alpha(x)-1} \\
&= \frac{\beta(x)}{s(x)(\alpha(x)-1)} \\
&= \frac{1}{s(x)} u_{alea}(x)
\end{aligned}
\end{equation}
Since $s(x) = \exp(-\delta C(x))$, then:
\begin{equation}
u'_{alea}(x) = \exp(\delta C(x))u_{alea}(x)
\end{equation}
Similarly, the adjusted epistemic uncertainty is:
\begin{equation}
\begin{aligned}
u'_{epi}(x) &= \frac{\beta(x)}{\nu'(x)(\alpha(x)-1)} \\
&= \frac{\beta(x)}{s(x)\nu(x)s(x)(\alpha(x)-1)} \\
&= \frac{1}{s(x)^2}u_{epi}(x)
\end{aligned}
\end{equation}
and therefore:
\begin{equation}
u'_{epi}(x) = \exp(2 \delta C(x))u_{epi}(x)
\end{equation}
Because $C(x) \geq 0$ and $\delta > 0$, both multiplicative factors are at least one. Consequently, CLEAR leaves the regression mean unchanged while monotonically increasing aleatoric and epistemic uncertainty as latent conflict increases.
\end{proof}


\section{Additional Experiments}
\label{sec:appendix-Additional Experiments}

\subsection{Corrupted Benchmarks}
\label{sec:appendix-Corrupted Benchmarks}
Table~\ref{tab:corrupted-results} evaluates performance under corrupted versions of the ID datasets. On MNIST$\rightarrow$MNIST-C, CLEAR achieves the strongest OOD and adversarial detection, with OOD and adversarial coverage of $19.26\%$ and $3.34\%$, and AUROC and ID-Adv. AUROC of $89.85\%$ and $96.51\%$, respectively. On CIFAR-10-C, CLEAR remains competitive for corruption detection and provides the strongest adversarial separation at both severity levels, achieving ID-Adv. AUROC of $69.90\%$ at severity 1 and $66.32\%$ at severity 5. Although I-EDL attains the highest ID-OOD AUROC at severity 5, CLEAR consistently gives the lowest adversarial coverage across both severities. The increase in ID-OOD AUROC from severity 1 to severity 5 for most methods also suggests that stronger corruptions become easier to distinguish as they move further from the clean ID distribution, while CLEAR continues to provide further robust separation under adversarial shift.

\begin{table*}[tb]
\caption{Mean accuracy, OOD detection, and adversarial attack detection performance with standard deviation of the comparative approaches with a variety of ID and OOD datasets in order of dataset difficulty. The adversarial attack is an $L_2$-PGD attack. For compactness, we denote adversarial AUROC as AAUROC. \colorbox{lightblue}{Highlighted} cells denote the best performance for each metric. * indicates datasets classed as Near-OOD.}
\label{tab:corrupted-results}
\centering
\resizebox{0.95\textwidth}{!}{%
\begin{tabular}{r|cccccccc|c}
\toprule
  &
  \multicolumn{1}{c}{\textbf{PostNet}} &
  \multicolumn{1}{c}{\textbf{EDL}} &
  \multicolumn{1}{c}{\textbf{I-EDL}}  &
  \multicolumn{1}{c}{\textbf{H-EDL}} &
  \multicolumn{1}{c}{\textbf{R-EDL}} &
    \textbf{RED}& \textbf{S-EDL}&\multicolumn{1}{c}{\textbf{C-EDL}} &
  \multicolumn{1}{|c}{\textbf{CLEAR}} \\
  \midrule
 Type& \multicolumn{6}{c|}{In-Training}&  \multicolumn{3}{c}{Post-Hoc}\\
    \midrule
      \multicolumn{10}{c}{\textbf{MNIST $\rightarrow$ MNIST-C*}}\\
    \midrule
    ID Acc ($\%$) $\uparrow$ &  $98.96 \pm 0.11$&  $99.26 \pm 0.07$&   $99.20 \pm 0.07$&  $98.33 \pm 0.38$&  $99.18 \pm 0.08$&   $99.01 \pm 0.12$&  $99.26 \pm 0.07$&\cellcolor{lightblue}\bm{$99.28 \pm 0.04$}&  $99.26 \pm 0.07$\\
    ID Cov ($\%$) $\uparrow$ &  $84.57 \pm 2.09$&  $83.08 \pm 1.15$&  $81.24 \pm 2.21$&  $77.86 \pm 3.37$&  $82.33 \pm 2.21$&   \cellcolor{lightblue}\bm{$86.52 \pm 2.64$}&  $81.77 \pm 2.61$&$81.38 \pm 1.93$&  $84.14 \pm 1.46$\\
    OOD Cov ($\%$) $\downarrow$ &  $29.15 \pm 2.49$&  $27.46 \pm 4.19$&  $26.24 \pm 6.49$&  $35.50 \pm 5.17$&  $25.75 \pm 3.15$&   $29.34 \pm 2.66$&  $26.27 \pm 4.47$&$30.35 \pm 3.92$&  \cellcolor{lightblue}\bm{$19.26 \pm 1.97$}\\
    Adv Cov ($\%$) $\downarrow$ &  $49.64 \pm 6.84$&  $12.85 \pm 3.91$&  $13.44 \pm 5.42$&  $19.44 \pm 6.08$&  $23.38 \pm 4.62$&   $47.47 \pm 6.96$&  $10.23 \pm 4.10$&$5.88 \pm 3.45$&  \cellcolor{lightblue}\bm{$3.34 \pm 0.92$}\\
    AUROC ($\%$) $\uparrow$ &  $84.11 \pm 1.12$&  $83.89 \pm 3.67$&  $83.16 \pm 4.51$&  $77.01 \pm 5.08$&  $84.84 \pm 1.43$&   $81.52 \pm 0.93$&  $83.89 \pm 3.68$&$82.02 \pm 3.52$&  \cellcolor{lightblue}\bm{$89.85 \pm 0.34$}\\
    AAUROC ($\%$) $\uparrow$ &  $61.89 \pm 5.32$&  $90.49 \pm 2.59$&  $90.02 \pm 2.63$&  $85.90 \pm 5.12$&  $85.47 \pm 1.85$&   $65.77 \pm 4.61$&  $91.60 \pm 2.50$&$93.99 \pm 1.94$&  \cellcolor{lightblue}\bm{$96.51 \pm 0.37$}\\
    \midrule
      \multicolumn{10}{c}{\textbf{CIFAR-10 $\rightarrow$ CIFAR-10-C (Severity-1)*}}\\
    \midrule
    ID Acc ($\%$) $\uparrow$ & $88.89 \pm 0.22$ & $89.76 \pm 0.19$ & $88.56 \pm 0.13$ & $85.05 \pm 0.21$ & $89.10 \pm 0.20$ & $88.61 \pm 0.27$ & $89.64 \pm 0.12$ & \cellcolor{lightblue}\bm{$90.11 \pm 0.15$}& $89.76 \pm 0.19$ \\
    ID Cov ($\%$) $\uparrow$ & $59.54 \pm 5.22$ & $55.26 \pm 5.10$ & $55.42 \pm 5.01$ & \cellcolor{lightblue}\bm{$63.84 \pm 6.26$}& $55.64 \pm 4.14$ & $55.83 \pm 4.79$ & $58.59 \pm 3.40$ & $55.64 \pm 7.81$ & $55.69 \pm 3.47$ \\
    OOD Cov ($\%$) $\downarrow$ & $50.35 \pm 5.03$ & \cellcolor{lightblue}\bm{$43.85 \pm 4.94$}& $43.97 \pm 5.10$ & $54.72 \pm 6.51$ & $44.63 \pm 4.05$ & $44.88 \pm 4.79$ & $46.83 \pm 3.54$ & $44.44 \pm 7.76$ & $44.22 \pm 3.53$ \\
    Adv Cov ($\%$) $\downarrow$ & $90.95 \pm 1.12$ & $97.53 \pm 0.13$ & $82.10 \pm 1.29$ & $30.99 \pm 8.19$ & $89.92 \pm 1.89$ & $93.20 \pm 17.66$ & $97.39 \pm 0.15$ & $94.28 \pm 1.80$ & \cellcolor{lightblue}\bm{$12.24 \pm 3.45$}\\
    AUROC ($\%$) $\uparrow$ & $56.23 \pm 0.19$ & $57.84 \pm 0.13$ & \cellcolor{lightblue}\bm{$57.97 \pm 0.16$}& $56.29 \pm 0.39$ & $57.40 \pm 0.14$ & $57.77 \pm 0.16$ & $57.93 \pm 0.09$ & $57.78 \pm 0.14$ & $57.94 \pm 0.12$ \\
    AAUROC ($\%$) $\uparrow$ & $9.19 \pm 1.07$ & $2.72 \pm 0.13$ & $14.66 \pm 0.71$ & $67.91 \pm 5.90$ & $11.52 \pm 0.51$ & $37.20 \pm 0.48$ & $3.28 \pm 0.15$ & $13.23 \pm 0.31$ & \cellcolor{lightblue}\bm{$69.90 \pm 0.42$}\\
    \midrule
      \multicolumn{10}{c}{\textbf{CIFAR-10 $\rightarrow$ CIFAR-10-C (Severity-5)*}}\\
    \midrule
    ID Acc ($\%$) $\uparrow$ & $88.82 \pm 0.11$ & $89.74 \pm 0.18$ & $88.59 \pm 0.11$ & $85.07 \pm 0.23$ & $89.08 \pm 0.19$ & $88.59 \pm 0.26$ & $89.64 \pm 0.14$ & \cellcolor{lightblue}\bm{$90.14 \pm 0.09$}& $89.77 \pm 0.17$ \\
    ID Cov ($\%$) $\uparrow$ & $70.85 \pm 1.32$ & $71.72 \pm 1.84$ & \cellcolor{lightblue}\bm{$75.55 \pm 2.03$}& $75.48 \pm 1.75$ & $67.72 \pm 3.89$ & $70.68 \pm 3.30$ & $71.96 \pm 2.91$ & $73.02 \pm 2.06$ & $65.61 \pm 1.34$ \\
    OOD Cov ($\%$) $\downarrow$ & $33.13 \pm 1.14$ & $29.27 \pm 1.89$ & $31.77 \pm 1.82$ & $35.52 \pm 2.03$ & $27.65 \pm 3.84$ & $29.73 \pm 3.49$ & $29.61 \pm 3.00$ & $30.98 \pm 2.27$ & \cellcolor{lightblue}\bm{$26.84 \pm 1.58$}\\
    Adv Cov ($\%$) $\downarrow$ & $97.66 \pm 0.10$ & $98.53 \pm 0.21$ & $86.48 \pm 3.30$ & $40.30 \pm 7.44$ & $92.04 \pm 0.78$ & $99.75 \pm 0.09$ & $98.50 \pm 0.22$ & $97.61 \pm 0.31$ & \cellcolor{lightblue}\bm{$31.63 \pm 3.58$}\\
    AUROC ($\%$) $\uparrow$ & $74.70 \pm 0.38$ & $77.85 \pm 0.09$ & \cellcolor{lightblue}\bm{$79.05 \pm 0.25$}& $76.04 \pm 0.88$ & $76.97 \pm 0.30$ & $77.18 \pm 0.26$ & $77.86 \pm 0.08$ & $77.59 \pm 0.07$ & $75.94 \pm 0.17$ \\
    AAUROC ($\%$) $\uparrow$ & $2.52 \pm 0.09$ & $2.14 \pm 0.23$ & $21.18 \pm 0.75$ & $63.48 \pm 6.86$ & $14.82 \pm 1.23$ & $38.00 \pm 0.58$ & $3.04 \pm 0.29$ & $14.49 \pm 0.43$ & \cellcolor{lightblue}\bm{$66.32 \pm 0.53$} \\
    \bottomrule
\end{tabular}
}
\end{table*}

\subsection{Extra Metrics Analysis}
\label{sec:appendix-Extra Metrics Analysis}
Table~\ref{tab:extra-metrics-results} provides additional insight into how CLEAR modifies the uncertainty produced by the underlying evidential model. Across the MNIST and CIFAR settings, CLEAR produces consistently larger confidence reductions for OOD and adversarial inputs, supporting the intended behaviour of the method: latent conflict primarily acts on predictions whose representations are inconsistent with the calibrated ID geometry, rather than uniformly suppressing evidence. This effect is strongest on settings where ID and shifted inputs are well separated in latent space, while the smaller confidence drops observed on the more challenging Oxford Flowers and ImageNet benchmarks suggest that the distinction becomes less pronounced when the underlying representation itself provides weaker separation. NLL and ECE show a different behaviour because these metrics evaluate calibration of the ID predictive distribution rather than separation between ID and unreliable inputs. CLEAR does not explicitly optimise either quantity and, by reducing evidential strength for latent-conflicting samples while preserving the base prediction, can produce more conservative confidence estimates without necessarily improving conventional calibration metrics. The results therefore highlight the distinction between predictive calibration and uncertainty separation: CLEAR is designed to expose unsupported evidence under distribution shift and attack, with its strongest gains appearing in the confidence separation between reliable and unreliable inputs. The additional latent conflict computation also introduces only a modest inference overhead relative to the underlying single-pass models; this computational behaviour is examined further in Section~\ref{sec:Runtime and Complexity}.

\begin{table*}[tb]
\caption{Mean calibration, confidence-drop, and inference-time performance with standard deviation of the comparative approaches across a variety of ID and OOD datasets in order of dataset difficulty. The adversarial attack is an $L_2$-PGD attack, and $\Delta$Conf. denotes the mean confidence drop from ID to OOD or adversarial inputs. \colorbox{lightblue}{Highlighted} cells denote the best performance for each metric. * indicates datasets classed as Near-OOD. Isolated best values arising from degenerate behaviour or severely degraded ID accuracy are not highlighted.}
\label{tab:extra-metrics-results}
\centering
\resizebox{0.95\textwidth}{!}{%
\begin{tabular}{r|cccccccc|c}
\toprule
  &
  \multicolumn{1}{c}{\textbf{PostNet}} &
  \multicolumn{1}{c}{\textbf{EDL}} &
  \multicolumn{1}{c}{\textbf{I-EDL}}  &
  \multicolumn{1}{c}{\textbf{H-EDL}} &
  \multicolumn{1}{c}{\textbf{R-EDL}} &
    \textbf{RED}& \textbf{S-EDL}&\multicolumn{1}{c}{\textbf{C-EDL}} &
  \multicolumn{1}{|c}{\textbf{CLEAR}} \\
  \midrule
 Type& \multicolumn{6}{c|}{In-Training}&  \multicolumn{3}{c}{Post-Hoc}\\
    \midrule
    \multicolumn{10}{c}{\textbf{MNIST $\rightarrow$ FashionMNIST}}\\
    \midrule
    NLL $\downarrow$ & \cellcolor{lightblue}\bm{$0.04 \pm 0.00$} & $0.11 \pm 0.00$ & $0.16 \pm 0.00$ & $1.15 \pm 0.12$ & $0.06 \pm 0.00$ & $0.06 \pm 0.01$ & $0.11 \pm 0.00$ & $0.15 \pm 0.00$ & $0.58 \pm 0.01$ \\
    ECE ($\%$) $\downarrow$ & \cellcolor{lightblue}\bm{$0.40 \pm 0.06$} & $8.58 \pm 0.13$ & $12.80 \pm 0.17$ & $66.05 \pm 3.34$ & $3.27 \pm 0.06$ & $0.74 \pm 0.11$ & $8.59 \pm 0.13$ & $11.97 \pm 0.12$ & $30.38 \pm 0.33$ \\
    $\Delta$Conf. ID-OOD ($\%$) $\uparrow$ & $26.31 \pm 4.68$ & $24.72 \pm 4.72$ & $25.87 \pm 5.65$ & $7.14 \pm 2.11$ & $24.94 \pm 4.51$ & $35.12 \pm 10.01$ & $24.72 \pm 4.73$ & $29.43 \pm 5.73$ & \cellcolor{lightblue}\bm{$58.84 \pm 0.33$} \\
    $\Delta$Conf. ID-Adv. ($\%$) $\uparrow$ & $0.30 \pm 1.36$ & $18.13 \pm 4.73$ & $16.76 \pm 2.78$ & $8.64 \pm 1.85$ & $5.61 \pm 2.38$ & $7.20 \pm 4.53$ & $18.85 \pm 4.73$ & $33.27 \pm 3.59$ & \cellcolor{lightblue}\bm{$58.76 \pm 0.36$} \\
    Inf. Time (s) $\downarrow$ & $2.21 \pm 0.06$ & $1.96 \pm 0.02$ & $1.96 \pm 0.02$ & $1.96 \pm 0.04$ & \cellcolor{lightblue}\bm{$1.94 \pm 0.03$} & $2.02 \pm 0.03$ & $57.56 \pm 0.09$ & $13.15 \pm 0.11$ & $2.36 \pm 0.15$ \\
    \midrule
    \multicolumn{10}{c}{\textbf{MNIST $\rightarrow$ KMNIST}}\\
    \midrule
    NLL $\downarrow$ & \cellcolor{lightblue}\bm{$0.04 \pm 0.00$} & $0.11 \pm 0.00$ & $0.16 \pm 0.00$ & $1.15 \pm 0.12$ & $0.06 \pm 0.00$ & $0.06 \pm 0.01$ & $0.11 \pm 0.00$ & $0.15 \pm 0.00$ & $0.58 \pm 0.01$ \\
    ECE ($\%$) $\downarrow$ & \cellcolor{lightblue}\bm{$0.40 \pm 0.06$} & $8.58 \pm 0.13$ & $12.80 \pm 0.17$ & $66.05 \pm 3.34$ & $3.27 \pm 0.06$ & $0.74 \pm 0.11$ & $8.59 \pm 0.13$ & $11.97 \pm 0.12$ & $30.38 \pm 0.33$ \\
    $\Delta$Conf. ID-OOD ($\%$) $\uparrow$ & $35.33 \pm 1.07$ & $33.16 \pm 1.08$ & $37.10 \pm 0.97$ & $8.92 \pm 1.18$ & $33.36 \pm 1.49$ & $52.93 \pm 4.82$ & $33.17 \pm 1.08$ & $40.75 \pm 1.27$ & \cellcolor{lightblue}\bm{$58.84 \pm 0.33$} \\
    $\Delta$Conf. ID-Adv. ($\%$) $\uparrow$ & $0.63 \pm 0.47$ & $17.38 \pm 1.84$ & $16.69 \pm 2.31$ & $9.83 \pm 1.30$ & $3.81 \pm 0.38$ & $4.13 \pm 2.88$ & $18.04 \pm 1.82$ & $34.47 \pm 1.86$ & \cellcolor{lightblue}\bm{$58.59 \pm 0.37$} \\
    Inf. Time (s) $\downarrow$ & $2.14 \pm 0.04$ & $1.94 \pm 0.03$ & \cellcolor{lightblue}\bm{$1.92 \pm 0.03$} & $1.98 \pm 0.06$ & $1.97 \pm 0.03$ & $1.99 \pm 0.02$ & $57.73 \pm 0.62$ & $12.94 \pm 0.17$ & $2.22 \pm 0.02$ \\
    \midrule
    \multicolumn{10}{c}{\textbf{MNIST $\rightarrow$ EMNIST*}}\\
    \midrule
    NLL $\downarrow$ & \cellcolor{lightblue}\bm{$0.04 \pm 0.00$} & $0.11 \pm 0.00$ & $0.16 \pm 0.00$ & $1.15 \pm 0.12$ & $0.06 \pm 0.00$ & $0.06 \pm 0.01$ & $0.11 \pm 0.00$ & $0.15 \pm 0.00$ & $0.58 \pm 0.01$ \\
    ECE ($\%$) $\downarrow$ & \cellcolor{lightblue}\bm{$0.40 \pm 0.06$} & $8.58 \pm 0.13$ & $12.80 \pm 0.17$ & $66.05 \pm 3.34$ & $3.27 \pm 0.06$ & $0.74 \pm 0.11$ & $8.60 \pm 0.12$ & $11.93 \pm 0.13$ & $30.38 \pm 0.32$ \\
    $\Delta$Conf. ID-OOD ($\%$) $\uparrow$ & $26.51 \pm 0.63$ & $26.94 \pm 0.44$ & $29.13 \pm 0.63$ & $7.80 \pm 1.03$ & $24.67 \pm 0.62$ & $37.29 \pm 2.34$ & $26.95 \pm 0.44$ & $33.86 \pm 0.65$ & \cellcolor{lightblue}\bm{$57.36 \pm 0.29$} \\
    $\Delta$Conf. ID-Adv. ($\%$) $\uparrow$ & $2.33 \pm 0.72$ & $18.81 \pm 1.63$ & $18.98 \pm 2.04$ & $9.35 \pm 1.15$ & $6.46 \pm 0.56$ & $4.07 \pm 1.73$ & $19.47 \pm 1.61$ & $34.62 \pm 1.45$ & \cellcolor{lightblue}\bm{$58.73 \pm 0.32$} \\
    Inf. Time (s) $\downarrow$ & $2.95 \pm 0.05$ & $2.73 \pm 0.18$ & $2.51 \pm 0.02$ & \cellcolor{lightblue}\bm{$2.50 \pm 0.03$} & $2.55 \pm 0.04$ & $2.57 \pm 0.04$ & $83.96 \pm 1.80$ & $18.52 \pm 0.45$ & $2.96 \pm 0.13$ \\
    \midrule
    \multicolumn{10}{c}{\textbf{CIFAR10 $\rightarrow$ SVHN}}\\
    \midrule
    NLL $\downarrow$ & \cellcolor{lightblue}\bm{$0.36 \pm 0.01$} & $0.42 \pm 0.00$ & $0.47 \pm 0.01$ & $1.03 \pm 0.11$ & $0.40 \pm 0.01$ & $0.40 \pm 0.01$ & $0.42 \pm 0.00$ & $0.44 \pm 0.00$ & $0.99 \pm 0.01$ \\
    ECE ($\%$) $\downarrow$ & $3.62 \pm 0.22$ & $5.66 \pm 0.17$ & $13.84 \pm 0.38$ & $45.51 \pm 5.04$ & \cellcolor{lightblue}\bm{$3.29 \pm 0.16$} & $4.71 \pm 0.25$ & $5.79 \pm 0.12$ & $11.19 \pm 0.19$ & $33.46 \pm 0.37$ \\
    $\Delta$Conf. ID-OOD ($\%$) $\uparrow$ & $30.53 \pm 2.54$ & $29.80 \pm 0.88$ & $44.15 \pm 0.66$ & $11.73 \pm 1.26$ & $28.35 \pm 1.14$ & $32.16 \pm 1.94$ & $23.53 \pm 0.84$ & $33.88 \pm 0.88$ & \cellcolor{lightblue}\bm{$44.17 \pm 0.43$} \\
    $\Delta$Conf. ID-Adv. ($\%$) $\uparrow$ & $-7.40 \pm 0.27$ & $-9.99 \pm 0.30$ & $0.82 \pm 1.75$ & $9.95 \pm 1.65$ & $2.17 \pm 1.23$ & $15.72 \pm 0.71$ & $-8.69 \pm 0.37$ & $-11.43 \pm 0.40$ & \cellcolor{lightblue}\bm{$42.83 \pm 0.33$} \\
    Inf. Time (s) $\downarrow$ & $6.07 \pm 0.11$ & $5.39 \pm 0.15$ & $6.52 \pm 3.90$ & $5.51 \pm 0.18$ & \cellcolor{lightblue}\bm{$5.05 \pm 0.09$} & $5.25 \pm 0.24$ & $181.17 \pm 2.29$ & $31.59 \pm 0.22$ & $6.96 \pm 0.04$ \\
    \midrule
    \multicolumn{10}{c}{\textbf{CIFAR10 $\rightarrow$ CIFAR100*}}\\
    \midrule
    NLL $\downarrow$ & \cellcolor{lightblue}\bm{$0.36 \pm 0.01$} & $0.42 \pm 0.00$ & $0.47 \pm 0.01$ & $1.03 \pm 0.11$ & $0.40 \pm 0.01$ & $0.40 \pm 0.01$ & $0.42 \pm 0.00$ & $0.43 \pm 0.00$ & $0.99 \pm 0.01$ \\
    ECE ($\%$) $\downarrow$ & $3.62 \pm 0.22$ & $5.66 \pm 0.17$ & $13.84 \pm 0.38$ & $45.51 \pm 5.04$ & \cellcolor{lightblue}\bm{$3.29 \pm 0.16$} & $4.71 \pm 0.25$ & $5.77 \pm 0.18$ & $11.23 \pm 0.29$ & $33.46 \pm 0.37$ \\
    $\Delta$Conf. ID-OOD ($\%$) $\uparrow$ & $20.35 \pm 0.36$ & $20.33 \pm 0.23$ & $30.44 \pm 0.50$ & $7.53 \pm 0.69$ & $19.95 \pm 0.24$ & $19.75 \pm 0.18$ & $20.58 \pm 0.27$ & $25.03 \pm 0.28$ & \cellcolor{lightblue}\bm{$39.63 \pm 0.39$} \\
    $\Delta$Conf. ID-Adv. ($\%$) $\uparrow$ & $-7.02 \pm 0.29$ & $-9.95 \pm 0.14$ & $-0.57 \pm 1.06$ & $10.39 \pm 1.43$ & $-0.27 \pm 0.57$ & $8.07 \pm 0.73$ & $-9.60 \pm 0.18$ & $-11.42 \pm 0.16$ & \cellcolor{lightblue}\bm{$36.78 \pm 0.61$} \\
    Inf. Time (s) $\downarrow$ & $3.69 \pm 0.19$ & $3.29 \pm 0.06$ & $3.24 \pm 0.29$ & $3.16 \pm 0.11$ & $3.24 \pm 0.09$ & \cellcolor{lightblue}\bm{$3.10 \pm 0.09$} & $103.78 \pm 2.37$ & $18.39 \pm 0.23$ & $4.14 \pm 0.10$ \\
    \midrule
    \multicolumn{10}{c}{\textbf{Oxford Flowers (low-shot) $\rightarrow$ Deep Weeds}}\\
    \midrule
    NLL $\downarrow$ & $4.48 \pm 0.04$ & $2.72 \pm 0.00$ & $3.07 \pm 0.01$ & $4.32 \pm 0.00$ & $1.65 \pm 0.02$ & \cellcolor{lightblue}\bm{$0.89 \pm 0.01$} & $2.74 \pm 0.01$ & $3.09 \pm 0.01$ & $3.42 \pm 0.01$ \\
    ECE ($\%$) $\downarrow$ & $4.03 \pm 1.46$ & $71.93 \pm 0.36$ & $75.90 \pm 0.81$ & $72.95 \pm 1.22$ & $56.97 \pm 0.91$ & \cellcolor{lightblue}\bm{$22.84 \pm 0.23$} & $70.92 \pm 0.34$ & $74.75 \pm 0.34$ & $75.14 \pm 0.32$ \\
    $\Delta$Conf. ID-OOD ($\%$) $\uparrow$ & $-0.26 \pm 0.20$ & $4.53 \pm 0.29$ & $2.67 \pm 0.14$ & $0.12 \pm 0.01$ & $20.38 \pm 0.56$ & \cellcolor{lightblue}\bm{$47.12 \pm 0.90$} & $4.38 \pm 0.28$ & $3.36 \pm 0.16$ & $4.40 \pm 0.03$ \\
    $\Delta$Conf. ID-Adv. ($\%$) $\uparrow$ & $0.87 \pm 0.14$ & $3.93 \pm 0.20$ & $2.58 \pm 0.08$ & $0.15 \pm 0.01$ & \cellcolor{lightblue}\bm{$10.88 \pm 0.83$} & $0.88 \pm 1.83$ & $4.34 \pm 0.19$ & $3.56 \pm 0.09$ & $4.40 \pm 0.03$ \\
    Inf. Time (s) $\downarrow$ & $20.62 \pm 2.95$ & $18.48 \pm 0.72$ & \cellcolor{lightblue}\bm{$18.09 \pm 1.22$} & $19.17 \pm 4.19$ & $18.93 \pm 2.45$ & $20.00 \pm 0.49$ & $660.19 \pm 0.23$ & $74.46 \pm 0.31$ & $18.69 \pm 0.45$ \\
    \midrule
    \multicolumn{10}{c}{\textbf{ImageNet $\rightarrow$ CUB}}\\
    \midrule
    NLL $\downarrow$ & $3.35 \pm 0.02$ & $0.99 \pm 0.00$ & $4.91 \pm 0.00$ & $6.47 \pm 0.00$ & $4.36 \pm 0.00$ & \cellcolor{lightblue}\bm{$0.90 \pm 0.00$} & $6.66 \pm 0.00$ & $6.72 \pm 0.00$ & $6.77 \pm 0.00$ \\
    ECE ($\%$) $\downarrow$ & $10.20 \pm 0.25$ & $4.56 \pm 0.05$ & $78.69 \pm 0.01$ & $78.55 \pm 0.00$ & $75.46 \pm 0.03$ & \cellcolor{lightblue}\bm{$4.37 \pm 0.04$} & $75.47 \pm 0.02$ & $75.97 \pm 0.05$ & $75.74 \pm 0.04$ \\
    $\Delta$Conf. ID--OOD ($\%$) $\uparrow$ & $5.24 \pm 0.73$ & \cellcolor{lightblue}\bm{$21.41 \pm 0.33$} & $0.15 \pm 0.00$ & $0.01 \pm 0.00$ & $0.25 \pm 0.00$ & $14.51 \pm 0.13$ & $0.00 \pm 0.00$ & $0.00 \pm 0.00$ & $0.01 \pm 0.00$ \\
    $\Delta$Conf. ID--Adv ($\%$) $\uparrow$ & $-67.34 \pm 0.56$ & $-11.81 \pm 0.11$ & $0.17 \pm 0.00$ & $0.01 \pm 0.00$ & \cellcolor{lightblue}\bm{$0.68 \pm 0.00$} & $-1.70 \pm 0.13$ & $0.00 \pm 0.00$ & $0.00 \pm 0.00$ & $0.02 \pm 0.00$ \\
    Inf. Time (s) $\downarrow$ & $453.67 \pm 115.99$ & \cellcolor{lightblue}\bm{$397.02 \pm 8.76$} & $451.55 \pm 119.19$ & $404.87 \pm 19.59$ & $411.16 \pm 19.94$ & $418.26 \pm 43.74$ & $7356.02 \pm 105.32$ & $1033.67 \pm 55.50$ & $423.66 \pm 44.30$ \\
    \bottomrule
\end{tabular}
}
\end{table*}

\subsection{Ablation Analysis}
\label{sec:appendix-Ablation Analysis}
We assess the sensitivity of CLEAR to five quantities controlling its latent conflict mechanism: the number of latent views $V$, perturbation magnitude $\tau$, evidence down-weighting strength $\delta$, reference conflict quantile $q$, and, for regression, the number of output bins $B$. Unless otherwise stated, we use $V=3$, $\tau=0.05$, $\delta=10$, and $q=0.5$, with all non-ablated parameters held fixed. The first four parameters affect different stages of the same inference process: $V$ determines how many local latent views are sampled, $\tau$ controls their displacement from the original representation, $q$ determines the calibration reference against which raw conflict is compared, and $\delta$ controls how strongly the resulting conflict reduces evidential strength. The regression-specific parameter $B$ instead determines the granularity of the output-conditioned latent geometry.

\begin{figure}
    \centering
    \includegraphics[width=\linewidth]{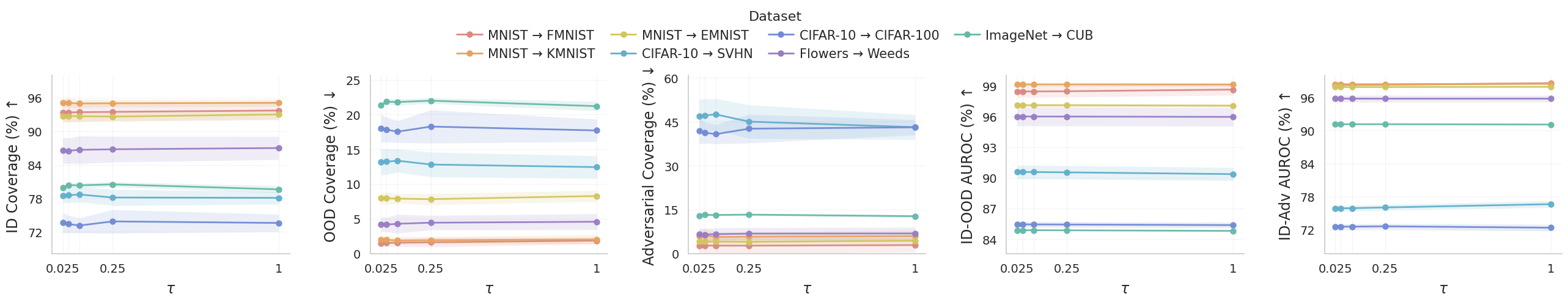}
    \caption{Hyperparameter sensitivity of CLEAR to the perturbation magnitude $\tau$. Each line denotes a dataset. From left to right, panels report ID coverage, OOD coverage, adversarial coverage, ID–OOD AUROC, and ID–adversarial AUROC. All other CLEAR hyperparameters are held fixed at their default values.}
    \label{fig:tauAblation}
\end{figure}

Figure~\ref{fig:tauAblation} examines the perturbation magnitude $\tau$ used to generate latent views, $h_v=h+\tau\rho_{\hat{g}}\eta_v$. Performance is remarkably stable across the evaluated range. ID, OOD, and adversarial coverage and both AUROC measures change only marginally for most datasets, despite the substantial change in the absolute perturbation scale. This robustness is consistent with the calibration-relative construction of CLEAR: $\tau$ changes the neighbourhood explored around the latent representation, but conflict is subsequently interpreted relative to a reference estimated using the same perturbation mechanism. Consequently, CLEAR depends more strongly on whether the local latent behaviour is unusual relative to calibrated ID structure than on the absolute size of the perturbation itself. We use $\tau=0.05$, which provides a local perturbation while lying within the broad region of stable performance.

\begin{figure}
    \centering
    \includegraphics[width=\linewidth]{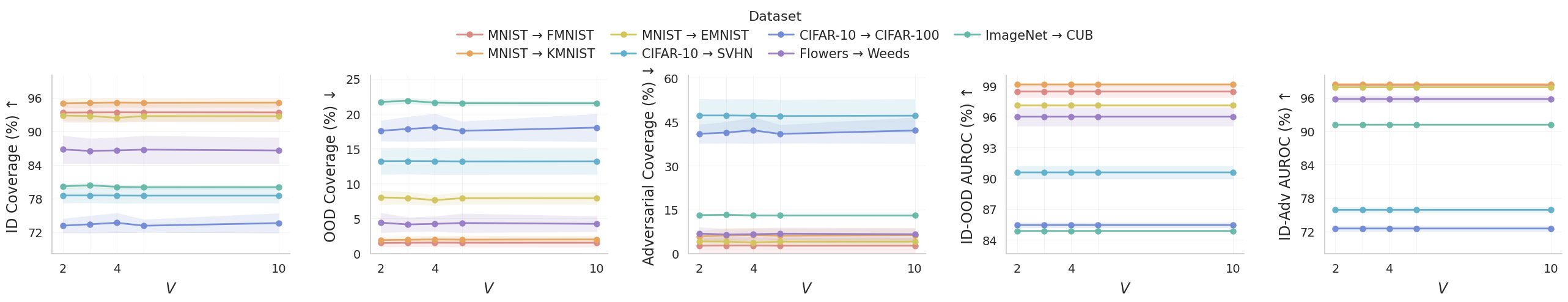}
    \caption{Hyperparameter sensitivity of CLEAR to the number of latent views $V$. Each line denotes a dataset. From left to right, panels report ID coverage, OOD coverage, adversarial coverage, ID–OOD AUROC, and ID–adversarial AUROC. All other CLEAR hyperparameters are held fixed at their default values.}
    \label{fig:viewsAblation}
\end{figure}

Figure~\ref{fig:viewsAblation} varies the number of sampled latent views $V$. With $V$ views, CLEAR evaluates $|\mathcal{P}_V|=\binom{V}{2}=\frac{V(V-1)}{2}$
unordered view pairs before averaging their conflicts. Increasing $V$ therefore increases the number of local comparisons quadratically, but does not directly increase the magnitude of the final score because the pairwise conflicts are averaged. Across the evaluated datasets, the resulting metrics remain almost unchanged as $V$ increases, indicating that a small number of views is already sufficient to characterise the local latent inconsistency used by CLEAR. Additional views consequently provide little empirical benefit while increasing the number of pairwise comparisons. We therefore select $V=3$, the smallest setting beyond the two-view case that provides multiple distinct pairwise comparisons, yielding three view pairs at low computational cost.

\begin{figure}
    \centering
    \includegraphics[width=\linewidth]{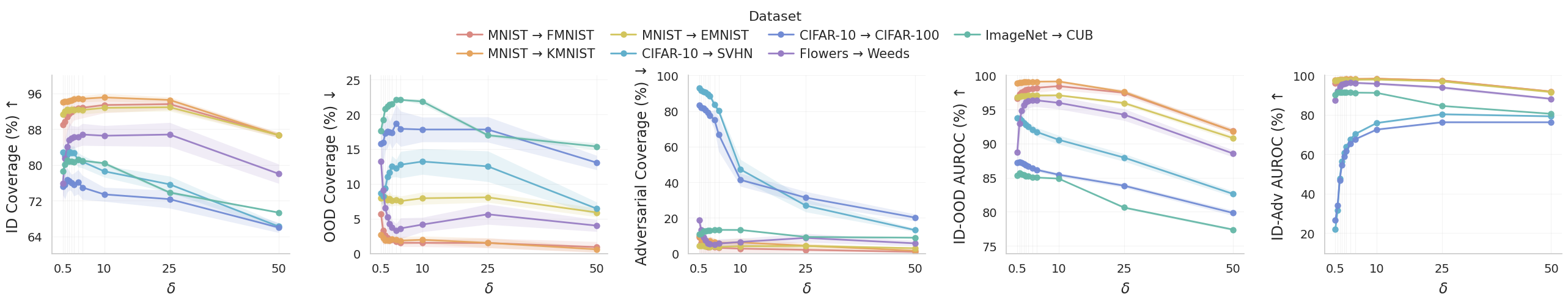}
    \caption{Hyperparameter sensitivity of CLEAR to the evidence down-weighting strength $\delta$. Each line denotes a dataset. From left to right, panels report ID coverage, OOD coverage, adversarial coverage, ID–OOD AUROC, and ID–adversarial AUROC. All other CLEAR hyperparameters are held fixed at their default values.}
    \label{fig:deltaAblation}
\end{figure}

Figure~\ref{fig:deltaAblation} varies the evidence down-weighting strength $\delta$ in $s(x)=\exp\left(-\delta C(x)\right)$. Increasing $\delta$ therefore strengthens the effect of any non-zero conflict while leaving samples with $C(x)=0$ unchanged. The results exhibit the corresponding trade-off. Moving away from very small values substantially improves adversarial rejection and ID-Adv. separation on the more challenging datasets, indicating that weak down-weighting does not sufficiently penalise unsupported evidence. However, increasingly large values eventually reduce ID coverage and ID-OOD AUROC on several datasets, as even moderate latent conflict induces a strong reduction in evidence. In contrast, the easier dataset pairs remain comparatively insensitive across a broad range of $\delta$. We select $\delta=10$, which lies within the stable intermediate region and provides strong OOD and adversarial discrimination without the degradation in retained ID behaviour observed under more aggressive down-weighting.

\begin{figure}
    \centering
    \includegraphics[width=\linewidth]{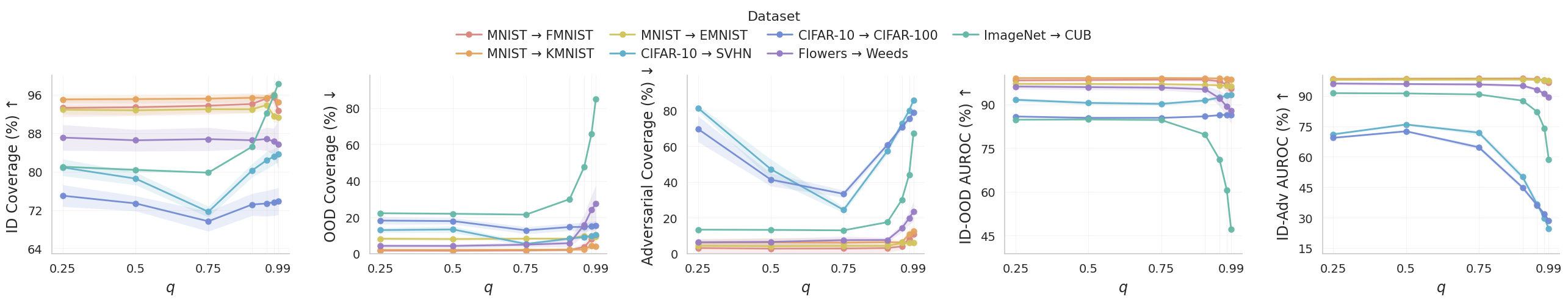}
    \caption{Hyperparameter sensitivity of CLEAR to the reference conflict quantile $q$. Each line denotes a dataset. From left to right, panels report ID coverage, OOD coverage, adversarial coverage, ID–OOD AUROC, and ID–adversarial AUROC. All other CLEAR hyperparameters are held fixed at their default values.}
    \label{fig:referenceAblation}
\end{figure}

CLEAR defines the reference conflict level as the median calibration conflict, $r=\operatorname{median}_{x_i\in\mathcal{D}_{\mathrm{cal}}}\overline{C}_{\mathrm{raw}}(x_i)$, which provides a robust central reference for typical ID latent behaviour and reduces sensitivity to unusually high-conflict calibration examples. Although the method is defined using the median, the underlying reference statistic can be varied by replacing the median with different calibration quantiles. We therefore perform this additional sensitivity analysis in Figure~\ref{fig:referenceAblation} to assess whether the intended median reference is well behaved, rather than treating the reference quantile as a parameter to tune for each dataset.

The results favour an intermediate reference level rather than either extreme. When a low calibration quantile is used, the resulting reference $r$ becomes small and a broad range of positive raw conflict scores are mapped towards large values of $C(x)$. This can reduce discrimination by causing both moderately and strongly conflicting samples to receive similarly aggressive evidence down-weighting. Conversely, very high calibration quantiles produce a large reference level, making CLEAR increasingly permissive because more samples fall below or close to the calibrated reference and therefore receive little adjustment. The median lies within the stable intermediate region across the evaluated datasets, supporting its use as the fixed reference statistic throughout the main experiments.

\begin{figure}
    \centering
    \includegraphics[width=\linewidth]{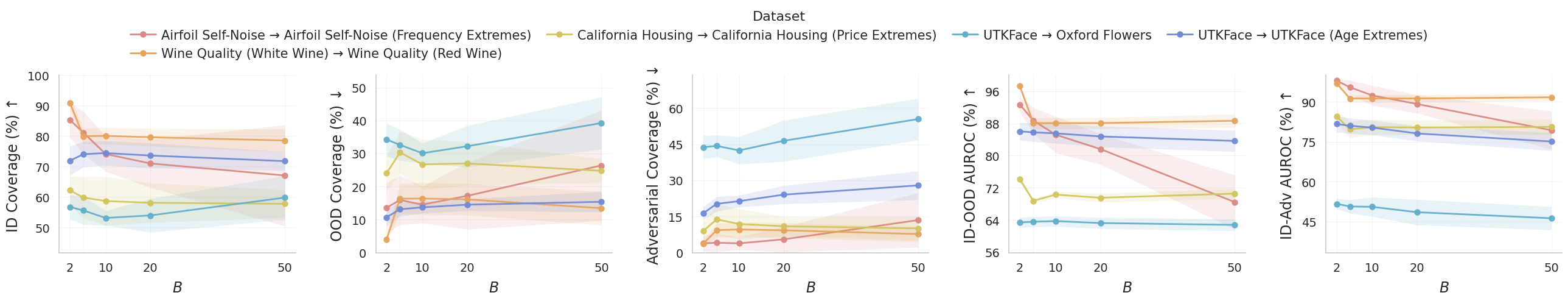}
    \caption{Hyperparameter sensitivity of CLEAR for regression to the number of bins $B$. Each line denotes a dataset. From left to right, panels report ID coverage, OOD coverage, adversarial coverage, ID–OOD AUROC, and ID–adversarial AUROC. All other CLEAR hyperparameters are held fixed at their default values.}
    \label{fig:binsAblation}
\end{figure}

Finally, Figure~\ref{fig:binsAblation} examines the regression-specific number of output bins $B$. The number of regression bins $B$ controls the granularity of the output-conditioned latent geometry. Smaller values produce broader groups with more calibration examples and therefore more stable estimates of $\mu_g$ and $\Sigma_g$, while larger values provide more locally specific geometry at the cost of fewer samples per group. The ablation therefore examines the trade-off between geometric specificity and reliable calibration.

The results indicate that relatively coarse output partitioning is sufficient, with little benefit from increasing $B$ aggressively. ID coverage remains broadly stable across the evaluated range, but larger values of $B$ tend to increase OOD and adversarial coverage for several datasets, indicating weaker rejection as the calibration examples are divided among increasingly small groups. This is reflected more clearly in the separation metrics: ID-OOD and ID-Adv. AUROC generally remain strongest for small-to-moderate $B$, while several datasets deteriorate slightly as $B$ becomes large. In particular, the Airfoil and UTKFace regression settings show increasingly poorer separation at $B=20$ and $B=50$, consistent with the per-bin estimates of $\mu_g$ and $\Sigma_g$ becoming less reliable when supported by fewer calibration examples. Very small $B$, however, produces comparatively coarse output-conditioned regions that may obscure local changes in latent geometry. We therefore use $B=10$, which provides a practical balance between output specificity and sufficiently populated calibration groups, while lying within the region of strong performance across the regression datasets.

\subsection{Threshold Analysis}
\label{sec:appendix-Threshold Analysis}
Table~\ref{tab:threshold-results} examines the effect of the uncertainty measure used for thresholding. CLEAR performs consistently across all four measures, maintaining strong OOD and adversarial separation despite changes in the uncertainty statistic. Differential entropy provides the strongest overall balance, retaining $94.83\%$ of ID samples while reducing OOD and adversarial coverage to $1.84\%$ and $7.21\%$, with ID-OOD AUROC of $98.54\%$ and ID-Adv. AUROC of $97.66\%$.

The alternative measures remain competitive, particularly for adversarial detection. Predictive entropy achieves an AAUROC of $95.36\%$ with adversarial coverage of $4.61\%$, while mutual information and total evidence retain AAUROC above $93\%$. Although these alternatives reduce ID coverage and OOD separation relative to differential entropy, CLEAR continues to distinguish unreliable inputs effectively across all settings. This indicates that the effectiveness of the latent conflict adjustment is not tied to a particular uncertainty statistic, while differential entropy offers the most favourable balance between retaining ID predictions and rejecting OOD and adversarial inputs.

The comparative approaches show greater variation across threshold types, with changes in the uncertainty statistic often producing larger shifts in coverage and AUROC. CLEAR therefore exhibits comparatively stable behaviour across the evaluated thresholding choices, while differential entropy provides a consistent default operating point for the main experiments.

\begin{table*}[tb]
\caption{Mean accuracy, OOD detection, and adversarial attack detection performance with standard deviation of the comparative approaches across a variety of uncertainty thresholding types for the MNIST $\rightarrow$ FashionMNIST setting. For compactness, we denote adversarial AUROC as AAUROC. \colorbox{lightblue}{Highlighted} cells denote the best performance for each metric.}
\label{tab:threshold-results}
\centering
\resizebox{\textwidth}{!}{%
\begin{tabular}{r|cccccccc|c}
\toprule
  &
  \multicolumn{1}{c}{\textbf{PostNet}} &
  \multicolumn{1}{c}{\textbf{EDL}} &
  \multicolumn{1}{c}{\textbf{I-EDL}}  &
  \multicolumn{1}{c}{\textbf{H-EDL}} &
  \multicolumn{1}{c}{\textbf{R-EDL}} &
    \textbf{RED}& \textbf{S-EDL}&\multicolumn{1}{c}{\textbf{C-EDL}} &
  \multicolumn{1}{|c}{\textbf{CLEAR}} \\
  \midrule
 Type& \multicolumn{6}{c|}{In-Training}&  \multicolumn{3}{c}{Post-Hoc}\\
    \midrule
    \multicolumn{10}{c}{\textbf{Differential Entropy (Default)}}\\
\midrule
ID Acc ($\%$) $\uparrow$ & $98.96 \pm 0.11$ & $99.26 \pm 0.07$ & $99.20 \pm 0.07$ & $98.33 \pm 0.38$ & $99.18 \pm 0.08$ & $99.01 \pm 0.12$ & $99.26 \pm 0.07$ & \cellcolor{lightblue}\bm{$99.28 \pm 0.04$}& $99.26 \pm 0.07$ \\
ID Cov ($\%$) $\uparrow$ & $91.93 \pm 1.89$ & $89.16 \pm 2.73$ & $89.49 \pm 3.06$ & $85.63 \pm 5.84$ & $92.33 \pm 2.10$ & $92.15 \pm 1.48$ & $89.19 \pm 2.67$ & $88.29 \pm 3.54$ & \cellcolor{lightblue}\bm{$94.83 \pm 0.85$}\\
OOD Cov ($\%$) $\downarrow$ & $5.14 \pm 2.63$ & $12.57 \pm 8.80$ & $13.20 \pm 10.46$ & $21.84 \pm 16.94$ & $7.83 \pm 5.99$ & $3.64 \pm 1.03$ & $12.68 \pm 8.93$ & $16.60 \pm 9.68$ & \cellcolor{lightblue}\bm{$1.84 \pm 0.91$}\\
Adv Cov ($\%$) $\downarrow$ & $86.16 \pm 10.82$ & $26.73 \pm 13.70$ & $31.36 \pm 16.33$ & $19.44 \pm 14.92$ & $64.58 \pm 12.73$ & $28.79 \pm 9.32$ & $23.05 \pm 13.12$ & $9.20 \pm 7.30$ & \cellcolor{lightblue}\bm{$7.21 \pm 3.90$}\\
AUROC ($\%$) $\uparrow$ & $97.55 \pm 1.12$ & $92.54 \pm 5.70$ & $91.94 \pm 7.22$ & $86.18 \pm 12.94$ & $95.95 \pm 3.22$ & $97.34 \pm 0.85$ & $92.52 \pm 5.73$ & $91.20 \pm 6.14$ & \cellcolor{lightblue}\bm{$98.54 \pm 0.36$}\\
AAUROC ($\%$) $\uparrow$ & $40.19 \pm 12.45$ & $88.78 \pm 5.29$ & $88.75 \pm 4.11$ & $88.59 \pm 10.79$ & $79.67 \pm 3.17$ & $77.72 \pm 8.60$ & $90.23 \pm 4.71$ & $94.91 \pm 2.13$ & \cellcolor{lightblue}\bm{$97.66 \pm 0.53$}\\
\midrule
\multicolumn{10}{c}{\textbf{Predictive Entropy}}\\
\midrule
ID Acc ($\%$) $\uparrow$ & $98.96 \pm 0.11$ & $98.69 \pm 0.16$ & \cellcolor{lightblue}\bm{$99.20 \pm 0.07$}& $98.33 \pm 0.38$ & $99.18 \pm 0.08$ & $99.01 \pm 0.12$ & $98.69 \pm 0.17$ & $98.76 \pm 0.16$ & $98.69 \pm 0.16$ \\
ID Cov ($\%$) $\uparrow$ & $90.99 \pm 2.12$ & $90.20 \pm 2.29$ & $89.96 \pm 2.88$ & $88.31 \pm 4.96$ & \cellcolor{lightblue}\bm{$92.37 \pm 2.12$}& $92.16 \pm 1.49$ & $90.25 \pm 2.55$ & $89.91 \pm 2.57$ & $78.76 \pm 7.87$ \\
OOD Cov ($\%$) $\downarrow$ & $4.23 \pm 2.48$ & $8.72 \pm 3.07$ & $10.84 \pm 8.24$ & $4.55 \pm 2.37$ & $9.03 \pm 6.20$ & \cellcolor{lightblue}\bm{$3.52 \pm 1.05$}& $8.72 \pm 2.90$ & $6.07 \pm 2.06$ & $8.00 \pm 2.71$ \\
Adv Cov ($\%$) $\downarrow$ & $84.30 \pm 11.35$ & $10.58 \pm 5.66$ & $24.85 \pm 12.99$ & $9.80 \pm 6.16$ & $65.87 \pm 12.69$ & $17.61 \pm 6.61$ & $9.31 \pm 4.99$ & \cellcolor{lightblue}\bm{$2.35 \pm 0.94$}& $4.61 \pm 2.39$ \\
AUROC ($\%$) $\uparrow$ & $97.44 \pm 1.06$ & $96.72 \pm 1.32$ & $93.34 \pm 6.00$ & $95.56 \pm 2.49$ & $95.65 \pm 3.40$ & \cellcolor{lightblue}\bm{$97.47 \pm 0.87$}& $96.72 \pm 1.32$ & $97.15 \pm 0.93$ & $93.31 \pm 2.92$ \\
AAUROC ($\%$) $\uparrow$ & $42.20 \pm 12.37$ & $95.72 \pm 1.67$ & $90.56 \pm 4.26$ & $94.73 \pm 2.94$ & $79.35 \pm 3.22$ & $89.22 \pm 3.68$ & $96.11 \pm 1.56$ & \cellcolor{lightblue}\bm{$98.23 \pm 0.56$}& $95.36 \pm 2.77$ \\
\midrule
\multicolumn{10}{c}{\textbf{Mutual Information}}\\
\midrule
ID Acc ($\%$) $\uparrow$ & $98.96 \pm 0.11$ & $98.69 \pm 0.16$ & \cellcolor{lightblue}\bm{$99.20 \pm 0.07$}& $98.33 \pm 0.38$ & $99.18 \pm 0.08$ & $99.01 \pm 0.12$ & $98.69 \pm 0.17$ & $98.76 \pm 0.16$ & $98.69 \pm 0.16$ \\
ID Cov ($\%$) $\uparrow$ & $91.40 \pm 1.47$ & $84.61 \pm 5.71$ & $87.69 \pm 2.97$ & $86.08 \pm 5.87$ & $90.83 \pm 2.54$ & \cellcolor{lightblue}\bm{$92.19 \pm 1.49$}& $84.57 \pm 5.51$ & $86.43 \pm 5.77$ & $75.49 \pm 12.68$ \\
OOD Cov ($\%$) $\downarrow$ & $6.35 \pm 3.65$ & $16.15 \pm 7.00$ & $13.86 \pm 12.13$ & $28.25 \pm 19.40$ & $9.64 \pm 6.51$ & \cellcolor{lightblue}\bm{$3.99 \pm 1.06$}& $16.23 \pm 7.80$ & $11.23 \pm 5.69$ & $10.07 \pm 6.50$ \\
Adv Cov ($\%$) $\downarrow$ & $86.92 \pm 10.22$ & $71.58 \pm 15.78$ & $33.91 \pm 15.05$ & $22.51 \pm 18.23$ & $71.80 \pm 18.28$ & $38.11 \pm 9.72$ & $63.67 \pm 16.56$ & $22.90 \pm 7.48$ & \cellcolor{lightblue}\bm{$8.38 \pm 5.31$}\\
AUROC ($\%$) $\uparrow$ & \cellcolor{lightblue}\bm{$97.19 \pm 1.31$}& $91.68 \pm 4.80$ & $90.51 \pm 8.67$ & $80.93 \pm 14.99$ & $94.95 \pm 3.79$ & $96.98 \pm 0.99$ & $91.68 \pm 4.78$ & $94.15 \pm 3.93$ & $91.47 \pm 3.76$ \\
AAUROC ($\%$) $\uparrow$ & $50.04 \pm 7.93$ & $61.86 \pm 11.32$ & $86.17 \pm 4.22$ & $85.54 \pm 12.83$ & $77.00 \pm 4.49$ & $68.65 \pm 10.55$ & $67.70 \pm 10.15$ & $89.92 \pm 4.35$ & \cellcolor{lightblue}\bm{$93.38 \pm 4.38$}\\
\midrule
\multicolumn{10}{c}{\textbf{Total Evidence}}\\
\midrule
ID Acc ($\%$) $\uparrow$ & $98.96 \pm 0.11$ & $98.69 \pm 0.16$ & \cellcolor{lightblue}\bm{$99.20 \pm 0.07$}& $98.33 \pm 0.38$ & $99.18 \pm 0.08$ & $99.01 \pm 0.12$ & $98.69 \pm 0.17$ & $98.76 \pm 0.16$ & $98.69 \pm 0.16$ \\
ID Cov ($\%$) $\uparrow$ & $91.46 \pm 1.57$ & $84.41 \pm 3.46$ & $87.04 \pm 3.94$ & $86.05 \pm 5.58$ & $90.02 \pm 2.31$ & \cellcolor{lightblue}\bm{$95.73 \pm 0.31$}& $84.13 \pm 5.26$ & $86.37 \pm 4.82$ & $75.63 \pm 12.73$ \\
OOD Cov ($\%$) $\downarrow$ & \cellcolor{lightblue}\bm{$6.45 \pm 3.60$}& $17.91 \pm 10.31$ & $13.80 \pm 11.71$ & $28.22 \pm 18.99$ & $10.37 \pm 7.26$ & $16.83 \pm 11.41$ & $17.60 \pm 10.01$ & $12.02 \pm 6.00$ & $10.63 \pm 6.49$ \\
Adv Cov ($\%$) $\downarrow$ & $87.00 \pm 10.32$ & $72.79 \pm 15.85$ & $34.22 \pm 15.05$ & $22.35 \pm 17.90$ & $71.85 \pm 15.70$ & $77.64 \pm 4.59$ & $65.18 \pm 15.98$ & $25.83 \pm 7.37$ & \cellcolor{lightblue}\bm{$8.64 \pm 5.36$}\\
AUROC ($\%$) $\uparrow$ & \cellcolor{lightblue}\bm{$97.19 \pm 1.31$}& $90.86 \pm 5.39$ & $90.28 \pm 8.88$ & $80.85 \pm 15.00$ & $94.37 \pm 4.20$ & $82.96 \pm 11.28$ & $90.86 \pm 5.37$ & $93.76 \pm 4.34$ & $91.35 \pm 3.82$ \\
AAUROC ($\%$) $\uparrow$ & $38.15 \pm 12.42$ & $59.83 \pm 11.71$ & $85.72 \pm 4.27$ & $85.49 \pm 12.84$ & $75.11 \pm 5.79$ & $29.18 \pm 5.93$ & $65.74 \pm 10.63$ & $88.83 \pm 4.90$ & \cellcolor{lightblue}\bm{$93.27 \pm 4.46$}\\
\bottomrule
\end{tabular}
}
\end{table*}

\subsection{Adversarial Attack Analysis}
\label{sec:appendix-Adversarial Attack Analysis}
Table~\ref{tab:attack-results} evaluates robustness across multiple adversarial attacks and perturbation strengths. CLEAR performs consistently strongly across the evaluated settings, achieving the lowest adversarial coverage in all but one case and the highest AAUROC in all but the weakest $L_2$-PGD setting, where it effectively matches PostNet ($98.93\%$ versus $98.98\%$). Its advantage becomes more pronounced as gradient-based attacks strengthen. Under $L_2$-PGD at strength $1.0$, CLEAR retains adversarial coverage of only $7.21\%$ and AAUROC of $97.66\%$, while several alternatives exhibit substantially greater degradation. A similar pattern is observed for FGSM, where CLEAR achieves near-zero adversarial coverage at strength $1.0$ while maintaining an AAUROC of $97.88\%$.

CLEAR is also particularly strong under AutoAttack, achieving the highest AAUROC at all three perturbation strengths: $98.87\%$, $98.35\%$, and $96.14\%$. At the strongest setting, its adversarial coverage remains low at $23.79\%$, only slightly above H-EDL at $21.56\%$, while CLEAR maintains a substantially higher AAUROC of $96.1\%$ compared with $88.45\%$. Thus, despite the small difference in coverage, CLEAR provides markedly stronger overall separation between clean and adversarial inputs. For Salt and Pepper corruption, CLEAR also achieves the lowest coverage and highest AAUROC across all three strengths, reaching $1.28\%$ coverage and $98.63\%$ AAUROC at strength $1.0$. Overall, CLEAR maintains strong performance across qualitatively different attack mechanisms and degrades substantially less than most comparative approaches as adversarial strength increases.

\begin{table*}[tb]
\caption{Mean adversarial coverage and AUROC performance with standard deviation of the comparative approaches across a variety of adversarial attack types and perturbation strengths for the MNIST $\rightarrow$ FashionMNIST setting. \colorbox{lightblue}{Highlighted} cells denote the best performance for each attack and strength.}
\label{tab:attack-results}
\centering
\resizebox{\textwidth}{!}{%
\begin{tabular}{ll|cccccccc|c}
\toprule
&&
\textbf{PostNet} &
\textbf{EDL} &
\textbf{I-EDL} &
\textbf{H-EDL} &
\textbf{R-EDL} &
\textbf{RED} &
\textbf{S-EDL} &
\textbf{C-EDL} &
\textbf{CLEAR} \\
\midrule
&& \multicolumn{6}{c|}{\textbf{In-Training}}
& \multicolumn{3}{c}{\textbf{Post-Hoc}} \\
\midrule

\multicolumn{11}{c}{\textbf{Adversarial Coverage ($\%$) $\downarrow$}} \\
\midrule
\multirow{3}{*}{$L_2$-PGD} & 0.1 & $0.80 \pm 0.64$ & $8.12 \pm 7.34$ & $8.58 \pm 9.30$ & $20.21 \pm 16.57$ & $3.38 \pm 3.69$ & $1.54 \pm 0.81$ & $8.59 \pm 7.64$ & $13.27 \pm 9.31$ & \cellcolor{lightblue}\bm{$0.65 \pm 0.45$}\\
 & 0.5 & $53.91 \pm 16.17$ & $7.07 \pm 6.38$ & $9.60 \pm 6.35$ & $17.66 \pm 15.24$ & $28.46 \pm 15.85$ & $19.03 \pm 4.72$ & $5.77 \pm 5.75$ & $6.21 \pm 5.89$ & \cellcolor{lightblue}\bm{$1.16 \pm 0.87$}\\
 & 1.0 & $86.16 \pm 10.82$ & $26.73 \pm 13.70$ & $31.36 \pm 16.33$ & $19.44 \pm 14.92$ & $64.58 \pm 12.73$ & $28.79 \pm 9.32$ & $23.05 \pm 13.12$ & $9.20 \pm 7.30$ & \cellcolor{lightblue}\bm{$7.21 \pm 3.90$}\\
\midrule
\multirow{3}{*}{FGSM} & 0.1 & $9.59 \pm 1.92$ & $29.87 \pm 5.88$ & $18.31 \pm 7.79$ & $26.51 \pm 16.57$ & $24.28 \pm 7.80$ & $8.22 \pm 4.64$ & $29.88 \pm 6.31$ & $11.23 \pm 3.10$ & \cellcolor{lightblue}\bm{$4.23 \pm 3.70$} \\
 & 0.5 & $16.97 \pm 7.94$ & $31.42 \pm 13.40$ & $38.04 \pm 14.34$ & $61.58 \pm 25.86$ & $52.62 \pm 17.53$ & $35.16 \pm 18.41$ & $31.78 \pm 13.17$ & $22.40 \pm 12.71$ & \cellcolor{lightblue}\bm{$2.88 \pm 5.05$} \\
 & 1.0 & $15.44 \pm 14.65$ & $91.85 \pm 9.20$ & $28.13 \pm 23.86$ & $59.22 \pm 31.95$ & $49.57 \pm 30.35$ & $68.40 \pm 24.68$ & $92.35 \pm 8.39$ & $80.22 \pm 17.23$ & \cellcolor{lightblue}\bm{$0.01 \pm 0.01$} \\
\midrule
\multirow{3}{*}{S\&P} & 0.1 & $5.00 \pm 2.60$ & $8.22 \pm 1.95$ & $13.19 \pm 10.47$ & $21.79 \pm 16.94$ & $7.75 \pm 6.02$ & $3.51 \pm 0.98$ & $8.39 \pm 2.11$ & $7.89 \pm 2.18$ & \cellcolor{lightblue}\bm{$1.61 \pm 0.40$}\\
 & 0.5 & $4.48 \pm 2.46$ & $8.04 \pm 1.94$ & $12.97 \pm 10.58$ & $21.49 \pm 16.92$ & $7.45 \pm 6.19$ & $3.11 \pm 0.91$ & $8.19 \pm 2.03$ & $7.35 \pm 2.06$ & \cellcolor{lightblue}\bm{$1.47 \pm 0.43$}\\
 & 1.0 & $3.70 \pm 2.27$ & $7.84 \pm 1.85$ & $12.75 \pm 10.73$ & $21.06 \pm 16.91$ & $7.16 \pm 6.41$ & $2.51 \pm 0.86$ & $8.04 \pm 1.94$ & $6.57 \pm 1.78$ & \cellcolor{lightblue}\bm{$1.28 \pm 0.47$}\\
\midrule
\multirow{3}{*}{AA ($L_2$)} & 0.1 & $4.40 \pm 1.98$ & $12.07 \pm 7.66$ & $10.86 \pm 8.34$ & $21.25 \pm 12.44$ & $6.23 \pm 3.70$ & $3.33 \pm 0.72$ & $8.91 \pm 5.88$ & $11.98 \pm 6.23$ & \cellcolor{lightblue}\bm{$1.59 \pm 0.07$} \\
 & 0.5 & $80.64 \pm 12.58$ & $15.51 \pm 5.96$ & $16.14 \pm 6.54$ & $20.73 \pm 12.62$ & $47.19 \pm 13.70$ & $26.86 \pm 9.45$ & $10.09 \pm 4.71$ & $5.96 \pm 4.18$ & \cellcolor{lightblue}\bm{$4.49 \pm 3.25$} \\
 & 1.0 & $99.76 \pm 0.41$ & $81.16 \pm 6.09$ & $81.19 \pm 11.77$ & \cellcolor{lightblue}\bm{$21.56 \pm 13.71$} & $85.53 \pm 4.98$ & $68.42 \pm 8.36$ & $77.43 \pm 6.00$ & $26.55 \pm 5.86$ & $23.79 \pm 8.59$ \\
\midrule
\multicolumn{11}{c}{\textbf{AAUROC ($\%$) $\uparrow$}} \\
\midrule
\multirow{3}{*}{$L_2$-PGD} & 0.1 & $98.98 \pm 0.36$ & $94.93 \pm 4.14$ & $94.24 \pm 5.72$ & $87.02 \pm 12.54$ & $97.86 \pm 1.64$ & $97.65 \pm 0.85$ & $94.70 \pm 4.32$ & $92.88 \pm 5.38$ & \cellcolor{lightblue}\bm{$98.93 \pm 0.30$} \\
 & 0.5 & $78.24 \pm 8.03$ & $95.65 \pm 1.87$ & $95.17 \pm 1.92$ & $88.83 \pm 11.23$ & $93.53 \pm 1.04$ & $86.71 \pm 5.43$ & $96.07 \pm 1.73$ & $96.00 \pm 2.51$ & \cellcolor{lightblue}\bm{$98.59 \pm 0.37$} \\
 & 1.0 & $40.19 \pm 12.45$ & $88.78 \pm 5.29$ & $88.75 \pm 4.11$ & $88.59 \pm 10.79$ & $79.67 \pm 3.17$ & $77.72 \pm 8.60$ & $90.23 \pm 4.71$ & $94.91 \pm 2.13$ & \cellcolor{lightblue}\bm{$97.66 \pm 0.53$} \\
\midrule
\multirow{3}{*}{FGSM} & 0.1 & $96.63 \pm 0.57$ & $88.72 \pm 3.31$ & $92.05 \pm 2.75$ & $85.76 \pm 11.74$ & $91.14 \pm 3.18$ & $93.53 \pm 5.44$ & $88.81 \pm 3.29$ & $95.07 \pm 1.79$ & \cellcolor{lightblue}\bm{$98.16 \pm 0.39$} \\
 & 0.5 & $94.74 \pm 2.53$ & $88.71 \pm 3.54$ & $81.26 \pm 7.36$ & $66.26 \pm 22.34$ & $78.11 \pm 10.34$ & $84.98 \pm 8.05$ & $88.66 \pm 3.55$ & $91.98 \pm 2.70$ & \cellcolor{lightblue}\bm{$97.70 \pm 0.46$} \\
 & 1.0 & $94.92 \pm 3.55$ & $53.19 \pm 14.44$ & $88.36 \pm 10.34$ & $65.80 \pm 32.64$ & $83.19 \pm 14.41$ & $63.24 \pm 14.52$ & $53.02 \pm 14.47$ & $69.35 \pm 12.22$ & \cellcolor{lightblue}\bm{$97.88 \pm 0.64$} \\
\midrule
\multirow{3}{*}{S\&P} & 0.1 & $97.62 \pm 1.10$ & $95.92 \pm 1.24$ & $91.96 \pm 7.23$ & $86.20 \pm 12.94$ & $95.98 \pm 3.23$ & $97.38 \pm 0.85$ & $95.91 \pm 1.25$ & $96.12 \pm 1.53$ & \cellcolor{lightblue}\bm{$98.47 \pm 0.46$} \\
 & 0.5 & $97.90 \pm 1.05$ & $95.99 \pm 1.22$ & $92.02 \pm 7.28$ & $86.30 \pm 12.94$ & $96.13 \pm 3.27$ & $97.60 \pm 0.86$ & $95.98 \pm 1.22$ & $96.32 \pm 1.47$ & \cellcolor{lightblue}\bm{$98.54 \pm 0.46$} \\
 & 1.0 & $98.18 \pm 0.97$ & $96.06 \pm 1.19$ & $92.10 \pm 7.35$ & $86.46 \pm 12.93$ & $96.26 \pm 3.32$ & $97.83 \pm 0.85$ & $96.05 \pm 1.20$ & $96.56 \pm 1.37$ & \cellcolor{lightblue}\bm{$98.63 \pm 0.47$} \\
\midrule
\multirow{3}{*}{AA ($L_2$)} & 0.1 & \cellcolor{lightblue}\bm{$98.07 \pm 0.58$} & $93.56 \pm 4.62$ & $93.71 \pm 5.47$ & $87.48 \pm 5.51$ & $97.06 \pm 1.77$ & $97.71 \pm 0.35$ & $95.48 \pm 2.77$ & $94.41 \pm 3.42$ & $98.87 \pm 0.31$ \\
 & 0.5 & $58.96 \pm 12.05$ & $92.48 \pm 3.66$ & $92.48 \pm 3.43$ & $87.91 \pm 5.68$ & $89.39 \pm 1.53$ & $89.61 \pm 4.40$ & $95.00 \pm 1.56$ & $96.19 \pm 2.01$ & \cellcolor{lightblue}\bm{$98.35 \pm 0.06$} \\
 & 1.0 & $1.82 \pm 1.16$ & $59.46 \pm 1.88$ & $57.38 \pm 6.41$ & $88.45 \pm 5.80$ & $40.13 \pm 4.43$ & $71.40 \pm 21.68$ & $65.47 \pm 2.05$ & $91.02 \pm 1.72$ & \cellcolor{lightblue}\bm{$96.14 \pm 0.14$} \\
\bottomrule
\end{tabular}%
}
\end{table*}

\subsection{Full Object Detection Results}
\label{sec:appendix-Full Object Detection Results}
To evaluate CLEAR beyond classification and regression, we additionally consider object detection on CelebA using YOLO11n~\citep{khanam2024yolov11}. Object detection provides a useful test case because uncertainty should reflect not only whether a detection is correct, but also how accurately its bounding box is localised.

We use YOLO11n as a fixed base detector and train an Evidential Meta Model (EMM)~\citep{shen2023post} on its intermediate features to produce detection-level evidential uncertainty over bounding-box localisation. EMM is well suited to this setting because it adds an evidential uncertainty model on top of a pretrained detector without requiring the detector itself to be modified. This allows YOLO11n to remain unchanged while providing the evidential outputs required by CLEAR. Otherwise, YOLO11n either cannot provide preliminary uncertainty outputs for CLEAR to build upon or would require full retraining if the head is replaced with a Dirichlet head, which would be extremely computationally expensive.

After EMM training, both YOLO11n and EMM are frozen. CLEAR is then calibrated using held-out CelebA detections and applied post-hoc to the EMM latent representation and evidential outputs. The comparison between EMM and EMM+CLEAR therefore isolates the effect of CLEAR: the underlying detector and bounding-box predictions remain unchanged, and only the evidential uncertainty is adjusted according to latent conflict.

We evaluate whether this adjusted uncertainty better reflects localisation quality using localisation error, defined as $1-\mathrm{IoU}$, together with AUROC, AUPRC, FPR95, calibration error, and the correlation between uncertainty and localisation error. We additionally evaluate the same relationship under brightness, blur, noise, compression, and resize/crop covariates.

\begin{table*}[tb]
\centering
\caption{Uncertainty quality on clean CelebA detections. Pearson's $r$ is reported between predictive uncertainty and localisation error ($1-\mathrm{IoU}$), such that higher correlation indicates that poorly localised detections receive greater uncertainty.}
\label{tab:clear_clean_uq}
\resizebox{0.92\textwidth}{!}{%
\begin{tabular}{lccccc}
\toprule
Method & AUROC ($\%$) $\uparrow$ & AUPRC ($\%$) $\uparrow$ & FPR95 ($\%$) $\downarrow$ & ECE ($\%$) $\downarrow$ & Pearson $r \uparrow$ \\
\midrule
EMM & $0.503 \pm 0.030$ & $0.125 \pm 0.028$ & $0.907 \pm 0.030$ & $0.844 \pm 0.023$ & $0.003 \pm 0.084$ \\
CLEAR & $\cellcolor{lightblue}\bm{0.585 \pm 0.042}$ & $\cellcolor{lightblue}\bm{0.211 \pm 0.087}$ & $\cellcolor{lightblue}\bm{0.824 \pm 0.095}$ & $\cellcolor{lightblue}\bm{0.750 \pm 0.160}$ & $\cellcolor{lightblue}\bm{0.168 \pm 0.099}$ \\
\bottomrule
\end{tabular}%
}
\end{table*}

Table~\ref{tab:clear_clean_uq} shows that CLEAR consistently improves uncertainty quality over EMM on clean CelebA detections. CLEAR increases AUROC from $0.503$ to $0.585$ and AUPRC from $0.125$ to $0.211$, while reducing FPR95 from $0.907$ to $0.824$ and ECE from $0.844$ to $0.750$. Importantly, the Pearson correlation between predictive uncertainty and localisation error increases from approximately zero for EMM ($r=0.003$) to $r=0.168$ with CLEAR. This indicates that CLEAR produces uncertainty that is more informative of bounding-box quality, with poorly localised detections more consistently receiving higher uncertainty.

\begin{table}[tb]
\centering
\caption{Linear association between predictive uncertainty and localisation error under covariate corruption on CelebA. Pearson's $r$ is reported, where higher is better.}
\label{tab:clear_corruption_pearson}
\resizebox{0.58\textwidth}{!}{%
\begin{tabular}{lccc}
\toprule
Corruption & EMM $r \uparrow$ & CLEAR $r \uparrow$ & $\Delta r$ \\
\midrule
Brightness & $0.009 \pm 0.017$ & $\cellcolor{lightblue}\bm{0.046 \pm 0.028}$ & $+0.037$ \\
Blur & $0.008 \pm 0.025$ & $\cellcolor{lightblue}\bm{0.052 \pm 0.035}$ & $+0.045$ \\
Noise & $0.022 \pm 0.027$ & $\cellcolor{lightblue}\bm{0.060 \pm 0.022}$ & $+0.039$ \\
Compression & $0.002 \pm 0.022$ & $\cellcolor{lightblue}\bm{0.048 \pm 0.024}$ & $+0.046$ \\
Resize/Crop & $0.009 \pm 0.013$ & $\cellcolor{lightblue}\bm{0.037 \pm 0.024}$ & $+0.027$ \\
\midrule
\textbf{Overall} & $0.010 \pm 0.019$ & $\cellcolor{lightblue}\bm{0.049 \pm 0.023}$ & $+0.039$ \\
\bottomrule
\end{tabular}
}
\end{table}

This behaviour remains consistent under covariate shift. As shown in Table~\ref{tab:clear_corruption_pearson}, CLEAR improves the uncertainty-localisation correlation for every evaluated corruption, increasing the overall Pearson correlation from $0.010$ to $0.049$. The improvement is observed across brightness, blur, noise, compression, and resize/crop corruptions, with the largest gains occurring for compression and blur. 

\begin{figure}
    \centering
    \includegraphics[width=\linewidth]{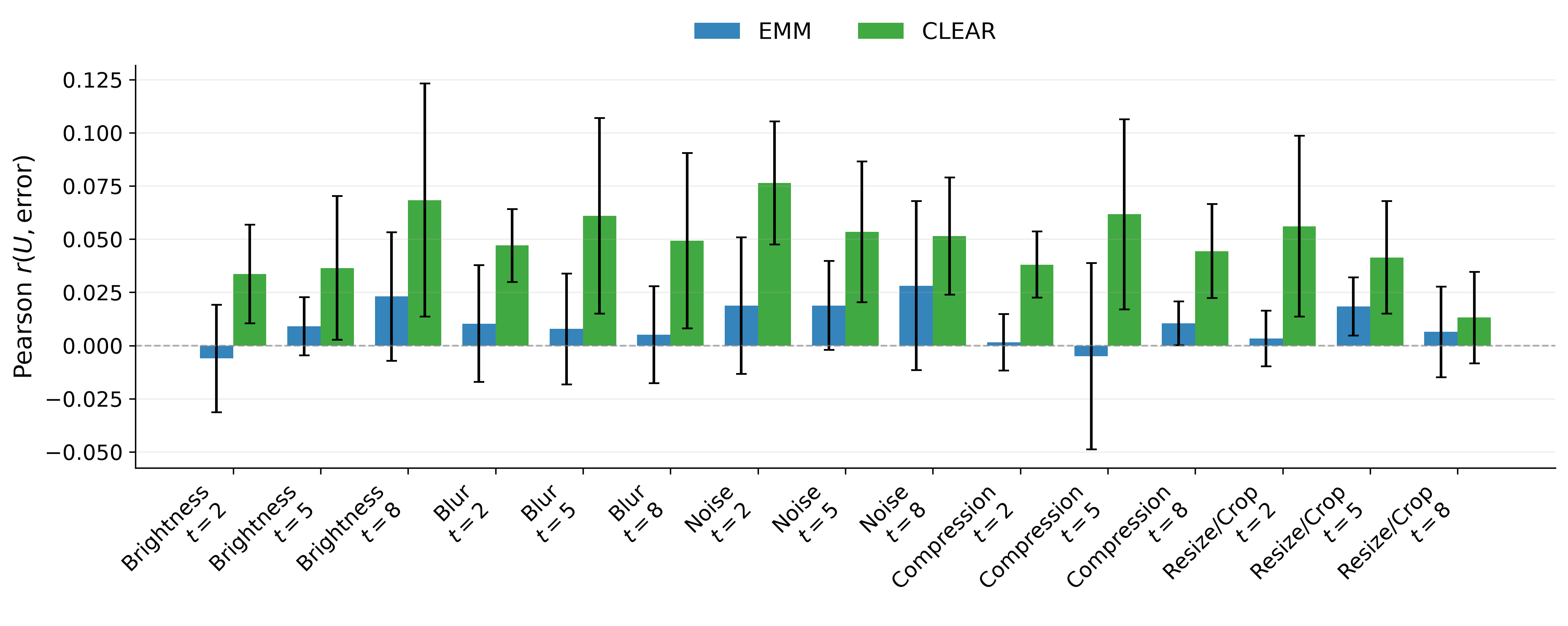}
    \caption{Pearson correlation between predictive uncertainty and localisation error under covariate corruption on CelebA. Higher correlation indicates that detections with larger localisation error receive greater uncertainty.}
    \label{fig:clear_corruption_ranking_pearson}
\end{figure}

Figure~\ref{fig:clear_corruption_ranking_pearson} further shows that this stronger relationship is generally maintained across corruption severities and all types, whereas EMM frequently remains close to zero correlation. Although the absolute correlations remain modest, the consistent improvement across clean data, corruption types, and corruption strengths suggests that CLEAR provides a more reliable localisation-aware uncertainty signal than the underlying EMM alone.

Overall, these results show that CLEAR extends effectively to object detection without modifying the underlying detector or its bounding-box predictions. Across both clean and corrupted CelebA inputs, CLEAR consistently makes the evidential uncertainty more reflective of localisation quality than EMM alone. This supports the broader task-agnostic design of CLEAR, showing that its latent-conflict mechanism can transfer from standard classification and regression settings to structured prediction tasks where uncertainty must capture the reliability of individual detections.

\subsection{Held-Out Analysis of the Latent Whitening Geometry}
\label{sec:appendix-Held-Out Analysis of the Latent Whitening Geometry}

CLEAR uses class/binned-conditional latent statistics to define the local geometry in which latent-view conflict is measured. For each group $g$, the training representations are used to estimate a mean $\mu_g$ and covariance $\Sigma_g = L_g L_g^\top$, and latent representations are subsequently centred and whitened as:
\begin{equation}
    \tilde{z} = L_g^{-1}(z - \mu_g)
\end{equation}
We therefore evaluate whether this training-derived geometry remains meaningful on held-out ID data.

\begin{figure}[t]
    \centering
    \begin{subfigure}[t]{0.49\linewidth}
        \centering
        \includegraphics[width=\linewidth]{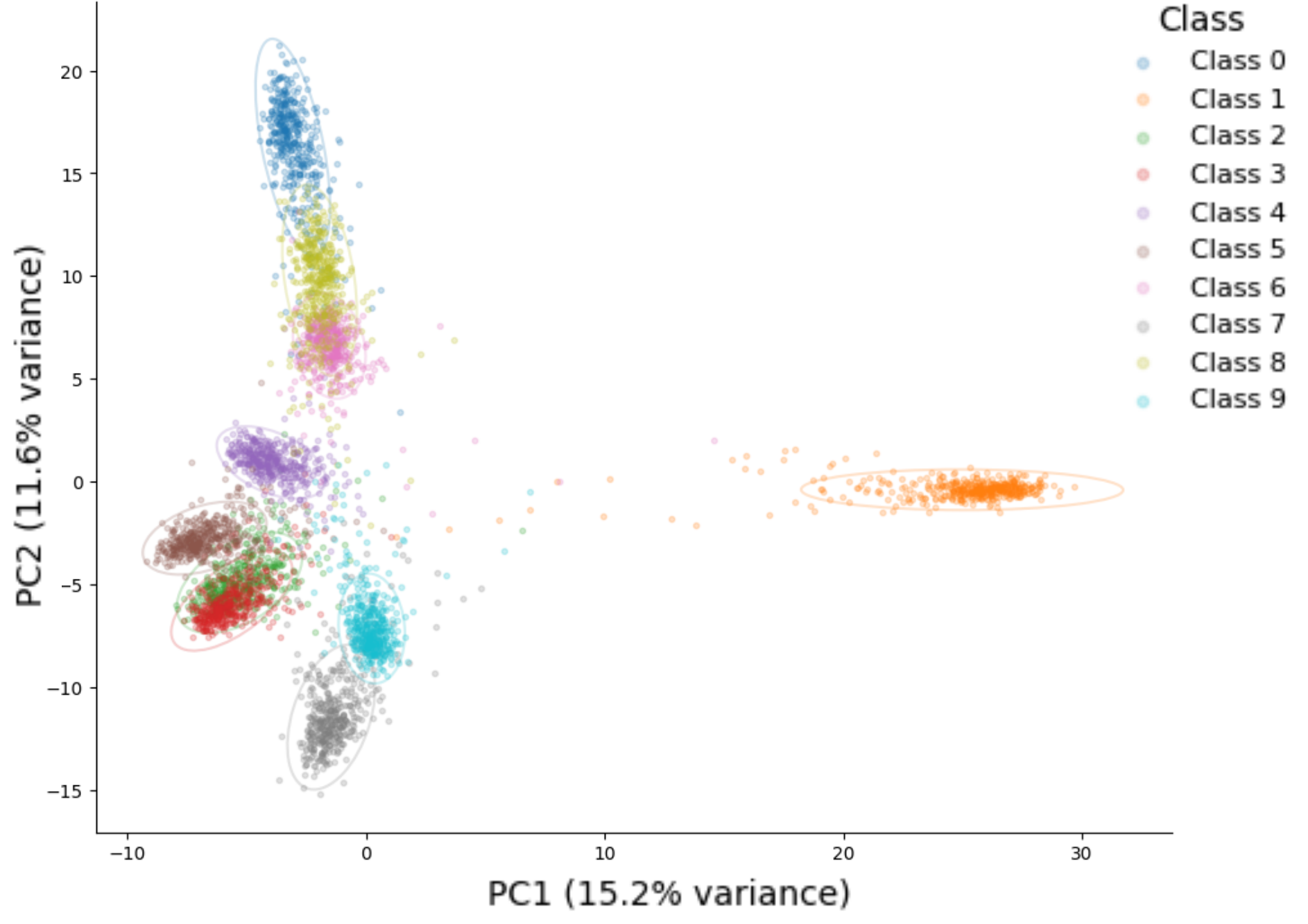}
        \caption{Held-out latent geometry with fitted covariance contours.}
        \label{fig:latent_geometry_pca}
    \end{subfigure}
    \hfill
    \begin{subfigure}[t]{0.49\linewidth}
        \centering
        \includegraphics[width=\linewidth]{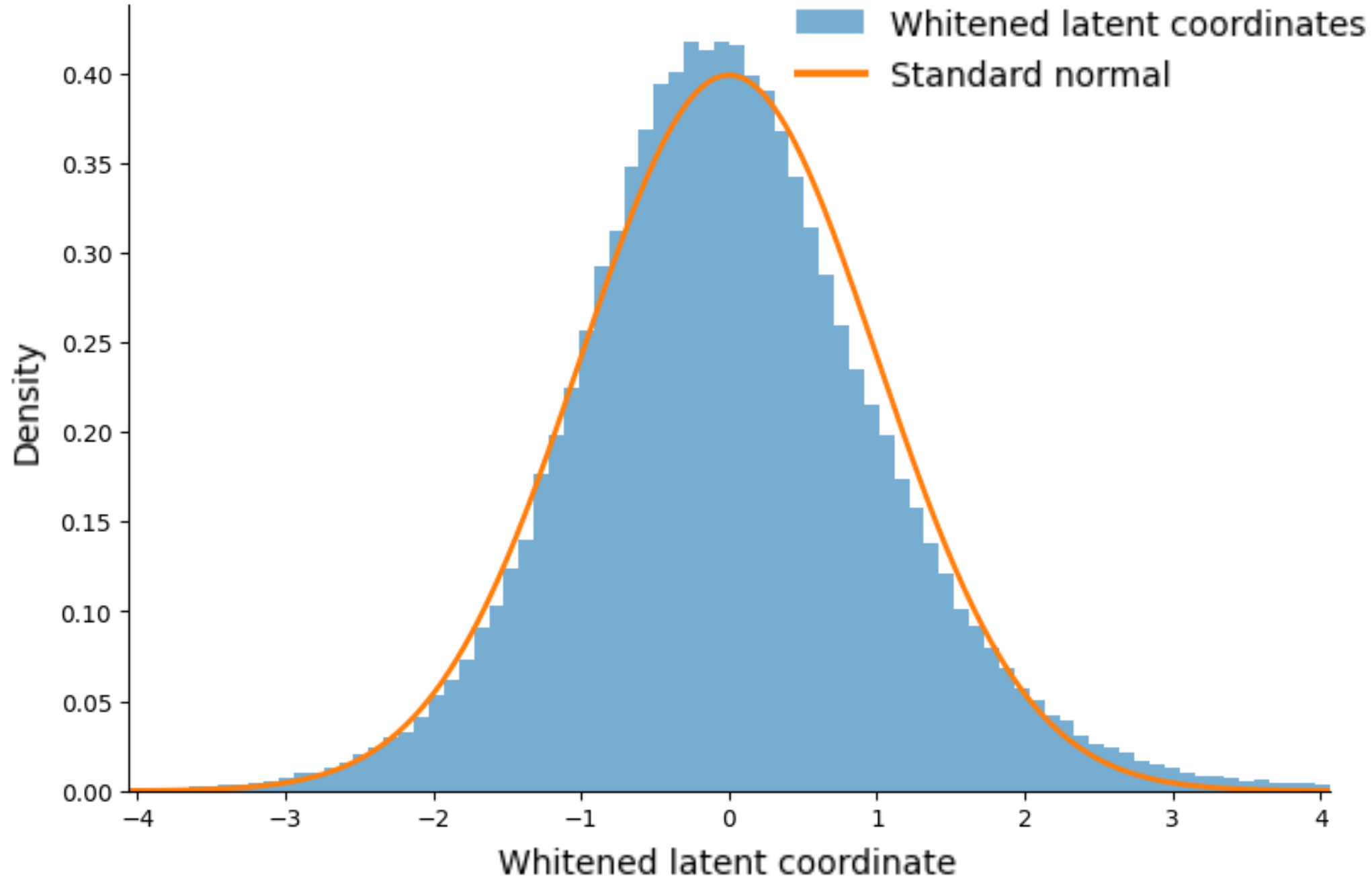}
        \caption{Whitened latent coordinates compared with a standard normal density.}
        \label{fig:latent_geometry_whitened}
    \end{subfigure}
    \caption{Held-out validation of the CLEAR latent geometry on MNIST. \textbf{Left:} PCA projection of held-out latent representations together with class-conditional covariance contours estimated from the training set. \textbf{Right:} Distribution of held-out latent coordinates after class-conditional whitening. The whitened coordinates are centred and approximately bell-shaped, but exhibit heavier tails than the standard normal.}
    \label{fig:latent_geometry}
\end{figure}

Figure~\ref{fig:latent_geometry} provides a qualitative view of the learned geometry on the MNIST dataset. In the PCA projection, the class\nobreakdash- conditional covariance contours broadly align with the orientation and spread of the corresponding held-out latent clusters, indicating that the estimated second-order structure captures the main class-wise geometry. After whitening, the pooled latent coordinates are centred and approximately bell-shaped, but remain broader and heavier-tailed than a standard normal density. Whitening removes class-specific location, scale, and linear covariance structure, but does not impose Gaussianity on the underlying latent representations, which are learned for the prediction task rather than explicitly constrained to follow a Gaussian distribution. Accordingly, CLEAR uses the fitted statistics as a local second-order geometry rather than as an exact generative model of the latent distribution.

We next quantify the effectiveness of the whitening transformation on held-out data. Whitening centres each representation using its class-specific mean and rescales it according to the corresponding Cholesky covariance factor; if the estimated geometry transfers to unseen samples, the transformed representations should have a mean close to zero and covariance close to the identity matrix. Across classes, the normalised whitened mean norm was $0.072$, the mean absolute off-diagonal covariance was $0.071$, and the mean diagonal covariance was $1.286$. These values indicate that the whitening geometry transfers reasonably well to unseen ID samples: the transformed representations remain approximately centred and largely decorrelated, with only moderate deviation from unit variance.

We also test the stronger distributional condition that the whitened latent representations follow an exact multivariate standard normal distribution. This condition is only partially satisfied. The nominal $50\%$ Gaussian region contains $52.9\%$ of held-out samples, but the nominal $95\%$ region contains only $63.4\%$, and the radial Kolmogorov-Smirnov statistic is $0.358$. Thus, the fitted Gaussian geometry captures the central structure of the latent space more faithfully than its tails. The network objective does not explicitly constrain the class-conditional latent representations to be Gaussian, so these deviations primarily concern higher-order distributional structure rather than the mean and covariance geometry used by CLEAR.

Finally, we test whether CLEAR depends critically on the use of Gaussian perturbations for generating latent views. Replacing the Gaussian perturbations with variance-matched Uniform or Rademacher noise leaves the conflict scores essentially unchanged, with Spearman rank correlations of $0.99998$ in both cases relative to the Gaussian baseline. This shows that CLEAR does not rely on exact Gaussianity of either the latent distribution or the perturbation law. Instead, its conflict mechanism primarily depends on the fitted local second-order geometry providing a stable coordinate system in which small perturbations expose latent inconsistency.

\begin{table}[tb]
\centering
\caption{Latent-geometry diagnostics for CLEAR on MNIST. Whitening statistics are computed per class on held-out ID representations using geometry estimated from the training set.}
\label{tab:latent_geometry_diagnostics}
\begin{tabular}{lcc}
\toprule
Diagnostic & Ideal & Result \\
\midrule
Whitened mean norm / $\sqrt{d}$ & $0$ & $0.072$ \\
Mean absolute off-diagonal covariance & $0$ & $0.071$ \\
Mean diagonal covariance & $1$ & $1.286$ \\
Empirical coverage of $50\%$ Gaussian region & $0.50$ & $0.529$ \\
Empirical coverage of $95\%$ Gaussian region & $0.95$ & $0.634$ \\
Radial KS statistic & $0$ & $0.358$ \\
Uniform vs.\ Gaussian conflict Spearman $\rho$ & $1$ & $0.99998$ \\
Rademacher vs.\ Gaussian conflict Spearman $\rho$ & $1$ & $0.99998$ \\
\bottomrule
\end{tabular}
\end{table}

\subsection{Sensitivity to Latent Representation Depth}
\label{sec:appendix-Sensitivity to Latent Representation Depth}
CLEAR uses the penultimate representation by default, as this is the final task-relevant feature representation provided to the evidential head. To assess whether its performance depends on this particular choice, we additionally evaluate CLEAR using representations extracted at progressively deeper stages of the same ResNet-18 backbone. Specifically, we consider the outputs of residual layers 1-3 and the penultimate representation, corresponding to latent dimensionalities of 64, 128, 256, and 512, respectively. For each of 10 runs on MNIST$\rightarrow$FashionMNIST, the underlying evidential model is trained only once and then held fixed across all four settings. 

Table~\ref{tab:latent-layer-ablation} shows that CLEAR remains highly effective across all evaluated representation depths. As expected, ID accuracy is identical across layers because the frozen predictive model is unchanged. The earlier representations perform particularly strongly on this benchmark: layer 1 achieves the highest ID coverage of $98.98\%$ and lowest OOD coverage of $0.41\%$, while layer 2 achieves the highest ID-OOD AUROC of $99.91\%$ and AAUROC of $99.93\%$. The penultimate representation nevertheless retains strong separation, with AUROC and AAUROC of $98.42\%$ and $98.29\%$, respectively.

Importantly, these results show that the penultimate representation used throughout the main experiments was not selected through benchmark-specific layer optimisation; indeed, shallower representations perform better for this particular dataset pair. We retain the penultimate layer as the default because it provides the representation directly consumed by the evidential head and gives a consistent, architecture-independent definition of the latent space across the different models and tasks considered in this work. The results further indicate that alternative intermediate representations could provide additional gains when architecture-specific layer selection is appropriate.

\begin{table*}[tb]
\caption{Sensitivity of CLEAR to the latent representation used for conflict estimation on MNIST$\rightarrow$FashionMNIST over 10 runs. The same trained evidential model is shared across all representations within each run, with only the CLEAR latent geometry recalibrated. AAUROC denotes mean ID-Adv. AUROC across the evaluated $L_2$-PGD attack strengths.}
\label{tab:latent-layer-ablation}
\centering
\resizebox{0.75\textwidth}{!}{%
\begin{tabular}{r|cccc}
\toprule
&
\multicolumn{1}{c}{\textbf{Layer 1}} &
\multicolumn{1}{c}{\textbf{Layer 2}} &
\multicolumn{1}{c}{\textbf{Layer 3}} &
\multicolumn{1}{c}{\textbf{Penultimate}} \\
\midrule
Latent Dim. &
$64$ &
$128$ &
$256$ &
$512$ \\
\midrule
ID Acc ($\%$) $\uparrow$ &
$99.25 \pm 0.06$ &
$99.25 \pm 0.06$ &
$99.25 \pm 0.06$ &
$99.25 \pm 0.06$ \\
ID Cov ($\%$) $\uparrow$ &
$98.98 \pm 0.22$ &
$98.82 \pm 0.27$ &
$97.90 \pm 0.55$ &
$93.81 \pm 1.44$ \\
OOD Cov ($\%$) $\downarrow$ &
$0.41 \pm 0.22$ &
$0.53 \pm 0.16$ &
$0.85 \pm 0.40$ &
$1.82 \pm 0.81$ \\
AUROC ($\%$) $\uparrow$ &
$99.88 \pm 0.04$ &
$99.91 \pm 0.02$ &
$99.80 \pm 0.05$ &
$98.42 \pm 0.45$ \\
AAUROC ($\%$) $\uparrow$ &
$99.88 \pm 0.02$ &
$99.93 \pm 0.02$ &
$99.83 \pm 0.04$ &
$98.29 \pm 0.13$ \\
\bottomrule
\end{tabular}
}
\end{table*}

\subsection{Qualitative Analysis of Latent Perturbation Views}
\label{sec:appendix-Qualitative Analysis of Latent Perturbation Views}
\begin{figure}[tb]
    \centering
    \includegraphics[width=0.9\linewidth]{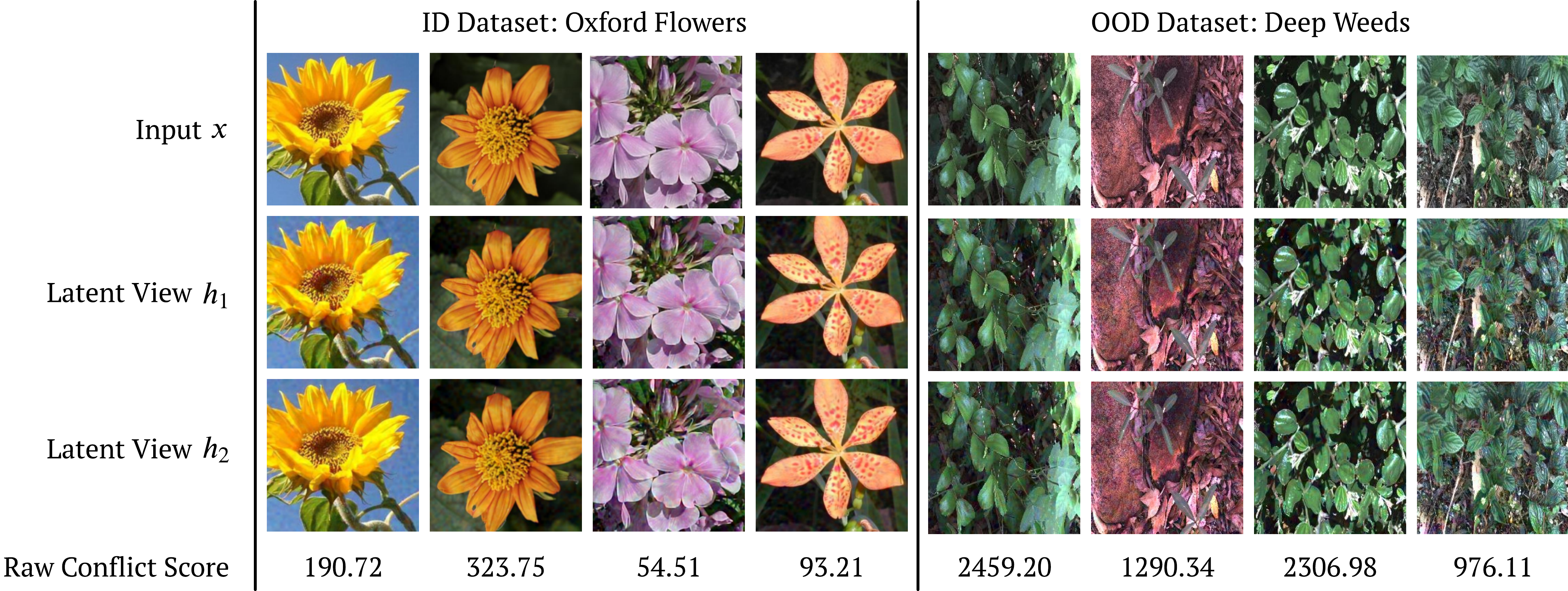}
    \caption{Qualitative visualisation of CLEAR's latent perturbation views for representative Oxford Flowers ID and DeepWeeds OOD inputs. For each input $x$, two covariance-shaped latent views $h_1$ and $h_2$ are visualised through feature inversion. The reconstructed views preserve the dominant semantic content of the input, while the corresponding raw conflict scores are substantially larger for the shown OOD examples than for the ID examples.}
    \label{fig:latentViewExamples}
\end{figure}

Figure~\ref{fig:latentViewExamples} provides an input-space interpretation of the latent perturbations used by CLEAR. For each input, two latent views are sampled using the predicted group's calibrated covariance geometry, following Equation~\ref{eq:latent-views}. To visualise these otherwise high-dimensional representations, we perform feature inversion by optimising a copy of the input image such that its representation under the frozen encoder approaches the corresponding target latent view, while applying pixel-distance and total-variation regularisation to retain an interpretable image. Across the Oxford Flowers examples, the resulting feature-inverted views remain visually close to the corresponding input, preserving the dominant flower structure, colour, and semantic content. The DeepWeeds examples similarly retain their overall appearance, showing that the perturbations remain local rather than moving to arbitrary regions of the representation space. Importantly, however, their latent behaviour differs substantially when evaluated relative to the calibrated Flowers geometry. For the examples shown, raw conflict ranges from $54.51$ to $323.75$ for ID inputs, compared with $976.11$ to $2459.20$ for OOD inputs. Thus, even when the perturbed views remain visually plausible, CLEAR can identify substantially greater incompatibility with the latent structure expected for the predicted group.

This distinction illustrates that CLEAR's conflict score does not simply measure the visible magnitude of a perturbation. Instead, it captures the cross-dimensional behaviour of the perturbed representations after conditioning on the calibrated group geometry. The qualitative examples therefore provide an intuitive view of how locally similar latent perturbations can remain compatible for familiar ID inputs but produce substantially greater conflict for unsupported OOD representations. Feature inversion is used here only as an interpretive visualisation and is not part of CLEAR inference. Moreover, the perturbation magnitude is deliberately enlarged for this analysis to make the corresponding image-space changes perceptible; the absolute conflict values shown should therefore be interpreted qualitatively rather than as reproductions of the operating scores used in the main experiments.


\section{Additional Discussion}
\label{sec:appendix-Additional Discussion}
\paragraph{Trade-offs between OOD and adversarial uncertainty.}
Some comparative approaches achieve stronger \textit{individual} metrics than CLEAR, but Figure~\ref{fig:core-results} shows that these gains are often accompanied by considerably larger losses on complementary uncertainty objectives. Importantly, the trade-off is frequently asymmetric: when CLEAR does not rank first, it generally remains close to the strongest result, whereas methods that outperform CLEAR on one metric can fall substantially behind on another. On MNIST$\rightarrow$KMNIST, for example, CLEAR's AAUROC of $97.36\%$ is only $0.68$ percentage points below H-EDL and $0.15$ below C-EDL, while still achieving the strongest ID coverage, OOD coverage, and ID-OOD AUROC. The contrast is more pronounced on CIFAR10$\rightarrow$CIFAR100, where CLEAR is only $1.76$ points below the best OOD AUROC and $4.41$ points above the best adversarial coverage, yet achieves an AAUROC of $71.93\%$, compared with only $12.22\%$ for C-EDL, the method attaining the highest OOD AUROC. Figure~\ref{fig:core-results} therefore illustrates that CLEAR's occasional losses tend to be comparatively small, while the complementary performance sacrificed by the corresponding best-performing alternatives can be substantially larger. This motivates evaluating uncertainty methods jointly across ID retention, OOD detection, and adversarial detection rather than selecting them according to a single favourable metric.

\paragraph{Behaviour on the more challenging benchmarks.}
The Oxford Flowers and ImageNet experiments expose several operating regimes that are less apparent on the smaller benchmarks. On Oxford Flowers$\rightarrow$DeepWeeds, R-EDL achieves the highest OOD AUROC, but its adversarial separation is considerably weaker than CLEAR. PostNet reports an adversarial coverage close to zero, but this accompanies an ID accuracy of only $5.92\%$ and ID coverage of $14.30\%$, indicating degradation of the underlying predictor rather than useful selective uncertainty. The low-shot nature of Oxford Flowers likely makes class-conditioned density estimation particularly difficult, highlighting the importance of interpreting coverage together with predictive performance.

On ImageNet$\rightarrow$CUB, the much larger 1000-class output space produces more pronounced extremes. I-EDL accepts almost all ID and OOD samples, whereas S-EDL and C-EDL reject almost all samples, including ID inputs. These behaviours likely reflect increased sensitivity of their uncertainty mechanisms at this scale: the substantially larger and more heterogeneous representation and output spaces can alter the magnitude and distribution of evidential, smoothing, and conflict-based uncertainty when method-specific settings are transferred unchanged from smaller benchmarks. These operating points can therefore produce favourable values for individual coverage metrics while providing limited practical selectivity. We deliberately retain common method-specific settings across datasets rather than independently tuning each approach for every benchmark, so these results also reveal differences in how the methods transfer to larger-scale settings. CLEAR avoids either extreme, retaining substantial ID coverage while achieving the highest OOD and adversarial AUROC.

\paragraph{Applying CLEAR to conventional pretrained models.}
A practical limitation of CLEAR is that it requires an evidential model from which both a latent representation and evidential output can be obtained. Consequently, when only a conventional pretrained model is available, directly applying CLEAR may otherwise require retraining the predictor using an evidential objective and evidential head, which can be computationally prohibitive for large models or datasets. A lightweight alternative is to use an Evidential Meta Model (EMM)~\citep{shen2023post}, which learns an evidential output from intermediate representations of a frozen pretrained model. This allows the original predictor to remain unchanged while providing the evidential representation required by CLEAR. Depending on the task, such a meta-model can parameterise an appropriate evidential distribution, such as a Dirichlet distribution for classification or a Normal-Inverse-Gamma distribution for regression. Our object-detection experiments, seen in Appendix~\ref{sec:appendix-Full Object Detection Results}, demonstrate this strategy in practice: EMM is trained over frozen YOLO11n features to model bounding-box localisation uncertainty, after which CLEAR is applied post-hoc to the resulting evidential model. The resulting improvements over EMM alone show that CLEAR can therefore be used with existing pretrained predictors without requiring the underlying model itself to be retrained as an evidential network.

\paragraph{Regression coverage trade-offs.}
The lower ID coverage observed for CLEAR on some regression benchmarks arises because its uncertainty adjustment is driven by support in the calibrated latent geometry. Samples that are only weakly supported by this geometry receive increased uncertainty, which can include difficult or atypical ID examples as well as genuinely shifted inputs. In addition, the operating threshold is selected to maximise ID-OOD separation rather than to enforce a fixed ID retention rate, so improved rejection of OOD and adversarial inputs can be accompanied by reduced ID coverage. Importantly, CLEAR preserves the underlying regression prediction, meaning that this effect reflects a change in selective uncertainty rather than degraded predictive accuracy.


\section{Experimental Setup}
\label{sec:appendix-Experimental Setup}

This section describes our experimental setup in detail to support reproducibility and clarify the evaluation protocol. We first introduce the datasets used in our experiments, followed by the model architectures considered. We then describe the comparative methods and training procedure before presenting the task-specific configurations and implementation details used across each experimental setting.

\subsection{Datasets}
\label{sec:appendix-datasets}
\begin{figure*}[htb]
    \centering
    \begin{subfigure}{0.475\linewidth}
        \centering
        \includegraphics[width=\linewidth]{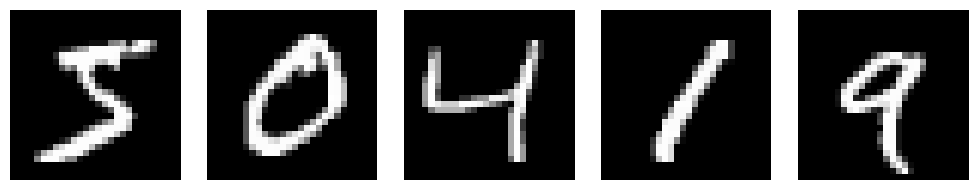}
        \caption{MNIST}
    \end{subfigure}
    \hfill
    \begin{subfigure}{0.475\linewidth}
        \centering
        \includegraphics[width=\linewidth]{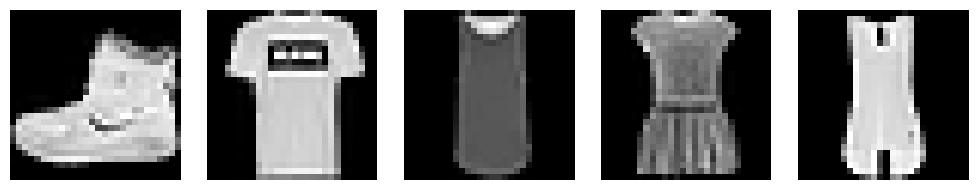}
        \caption{FMNIST}
    \end{subfigure}
    \begin{subfigure}{0.475\linewidth}
        \centering
        \includegraphics[width=\linewidth]{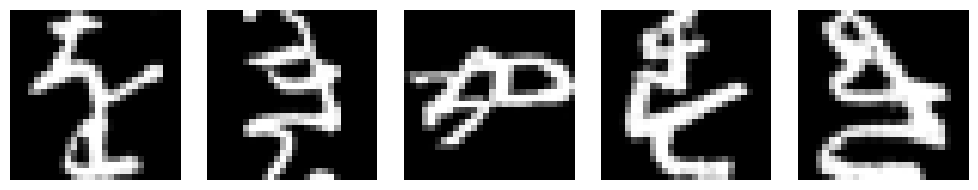}
        \caption{KMNIST}
    \end{subfigure}
    \hfill
    \begin{subfigure}{0.475\linewidth}
        \centering
        \includegraphics[width=\linewidth]{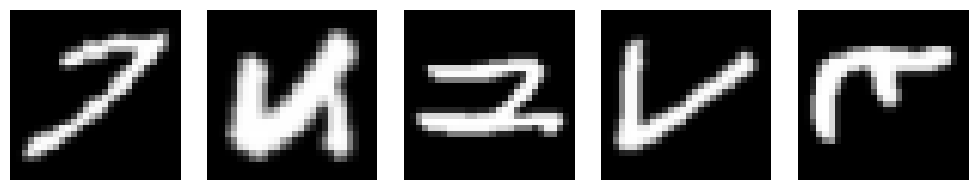}
        \caption{EMNIST}
    \end{subfigure}
    \begin{subfigure}{0.475\linewidth}
        \centering
        \includegraphics[width=\linewidth]{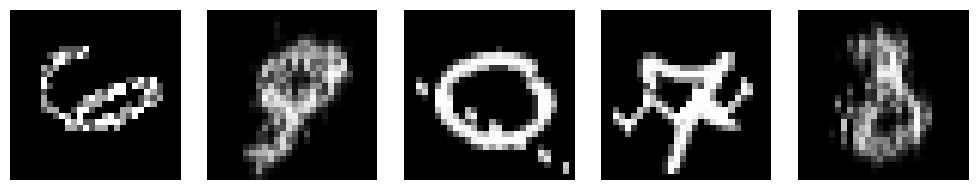}
        \caption{MNIST-C}
    \end{subfigure}
    \hfill
    \begin{subfigure}{0.475\linewidth}
        \centering
        \includegraphics[width=\linewidth]{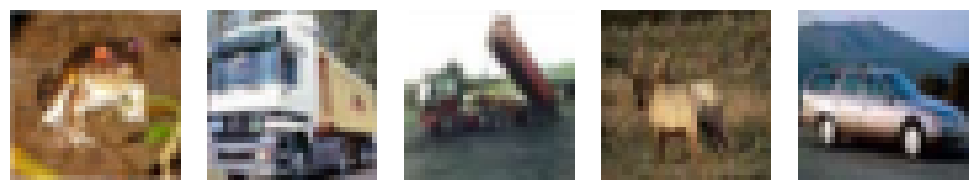}
        \caption{CIFAR10}
    \end{subfigure}
    \begin{subfigure}{0.475\linewidth}
        \centering
        \includegraphics[width=\linewidth]{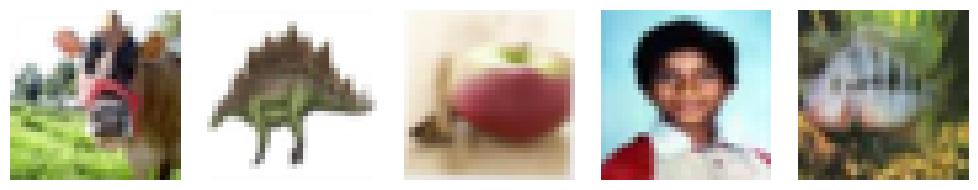}
        \caption{CIFAR100}
    \end{subfigure}
    \hfill
    \begin{subfigure}{0.475\linewidth}
        \centering
        \includegraphics[width=\linewidth]{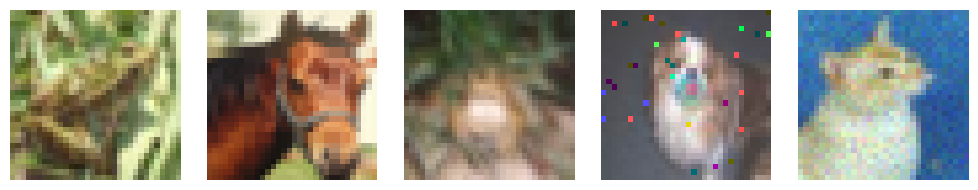}
        \caption{CIFAR-10-C (severity 1)}
    \end{subfigure}
    \begin{subfigure}{0.475\linewidth}
        \centering
        \includegraphics[width=\linewidth]{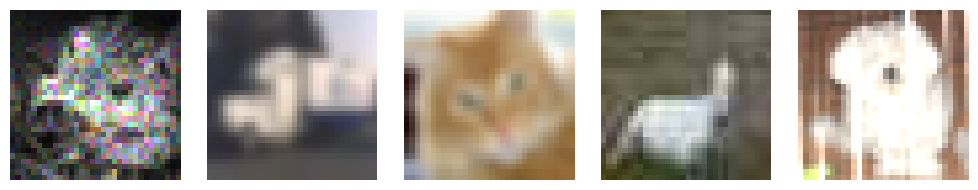}
        \caption{CIFAR-10-C (severity 5)}
    \end{subfigure}
    \hfill
    \begin{subfigure}{0.475\linewidth}
        \centering
        \includegraphics[width=\linewidth]{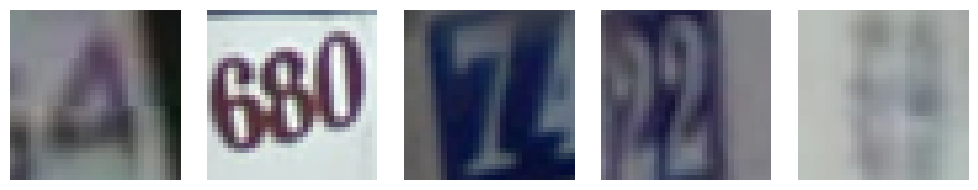}
        \caption{SVHN}
    \end{subfigure}
    \begin{subfigure}{0.475\linewidth}
        \centering
        \includegraphics[width=\linewidth]{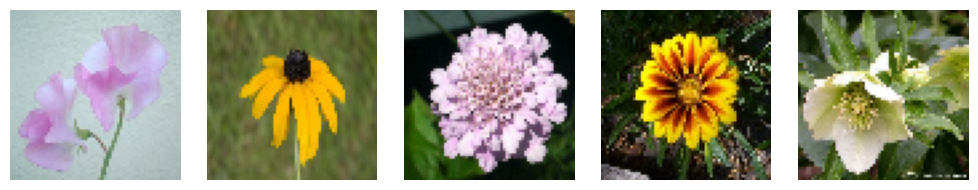}
        \caption{Oxford Flowers}
    \end{subfigure}
    \hfill
    \begin{subfigure}{0.475\linewidth}
        \centering
        \includegraphics[width=\linewidth]{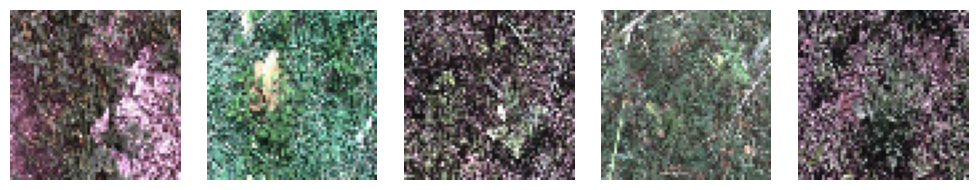}
        \caption{Deep Weeds}
    \end{subfigure}
    \begin{subfigure}{0.475\linewidth}
        \centering
        \includegraphics[width=\linewidth]{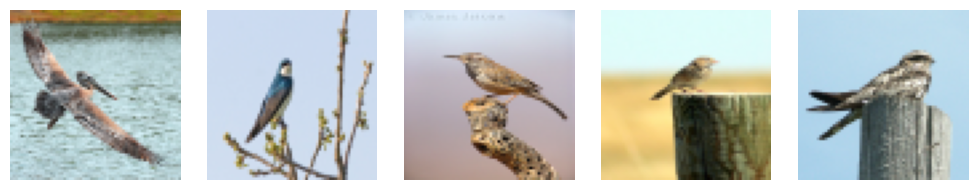}
        \caption{Caltech-UCSD Birds-200-2011 (CUB)}
    \end{subfigure}
    \hfill
    \begin{subfigure}{0.475\linewidth}
        \centering
        \includegraphics[width=\linewidth]{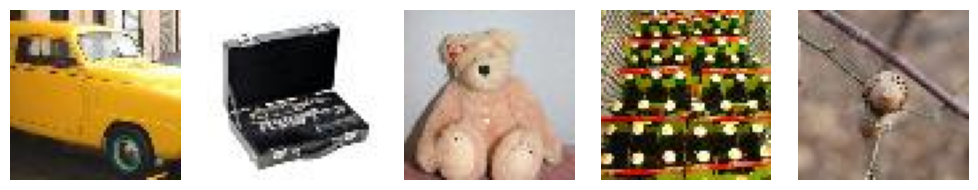}
        \caption{ImageNet}
    \end{subfigure}
    \begin{subfigure}{0.475\linewidth}
        \centering
        \includegraphics[width=\linewidth]{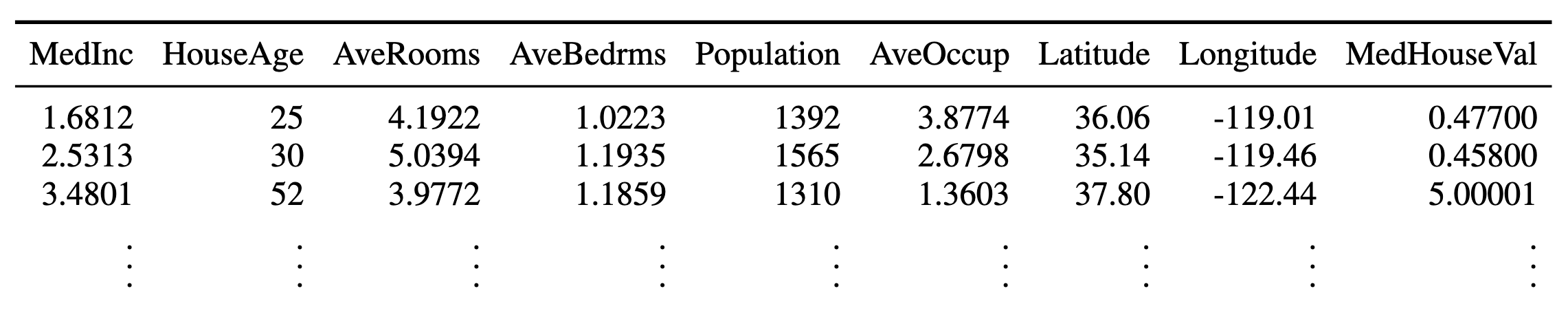}
        \caption{California Housing}
    \end{subfigure}
    \hfill
    \begin{subfigure}{0.475\linewidth}
        \centering
        \includegraphics[width=\linewidth]{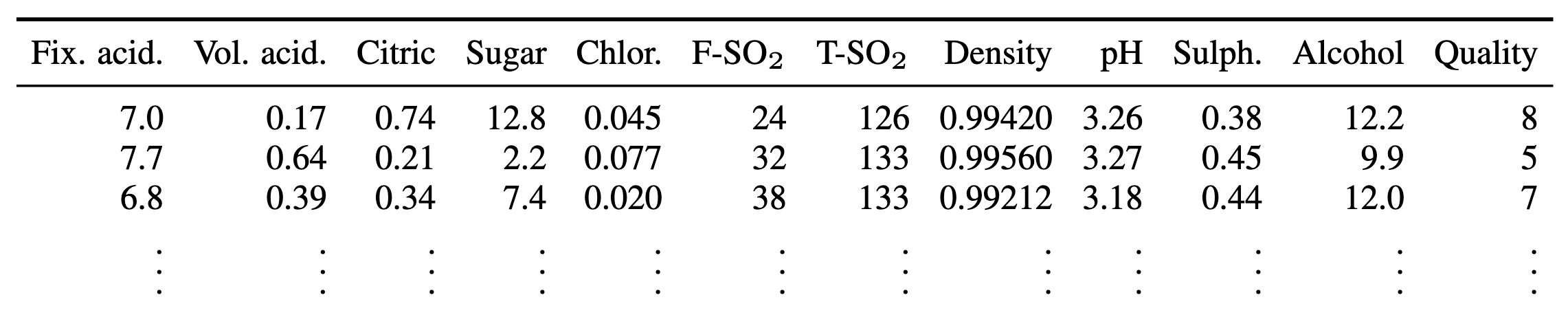}
        \caption{Wine Quality}
    \end{subfigure}
    \begin{subfigure}{0.475\linewidth}
        \centering
        \includegraphics[width=\linewidth]{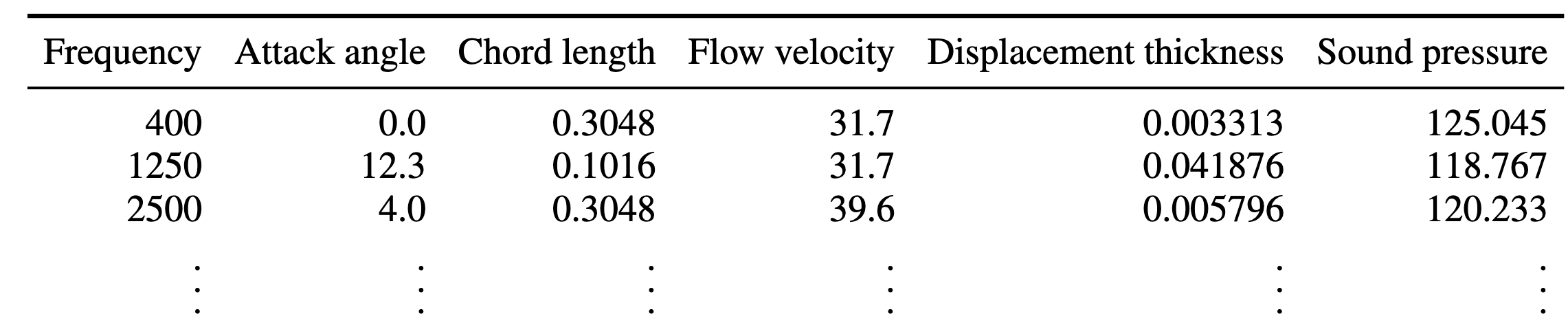}
        \caption{Airfoil Self-Noise}
    \end{subfigure}
    \hfill
    \begin{subfigure}{0.475\linewidth}
        \centering
        \includegraphics[width=\linewidth]{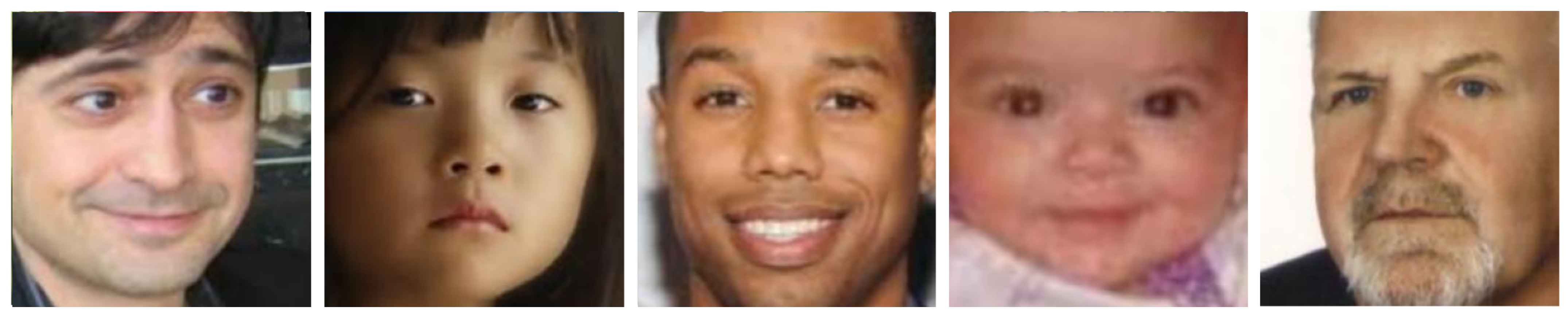}
        \caption{UTKFace}
    \end{subfigure}
    \begin{subfigure}{0.475\linewidth}
        \centering
        \includegraphics[width=\linewidth]{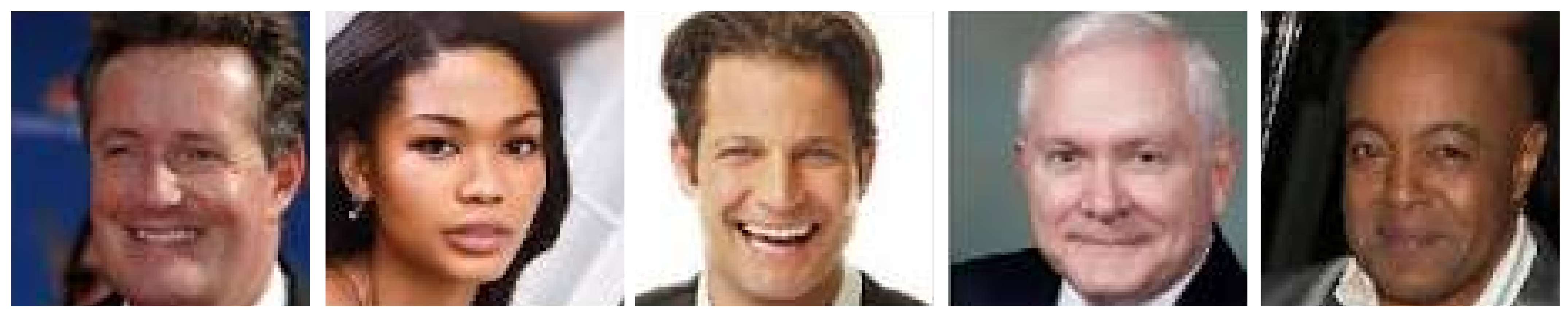}
        \caption{CelebA}
    \end{subfigure}
\caption{Example images and tables from the datasets used in our experimental evaluation.}
\label{fig:appendix-dataset-vis}
\end{figure*}

We evaluate CLEAR across a diverse set of established benchmarks spanning different domains and levels of task complexity. This enables a comprehensive assessment under varied experimental conditions. Dataset-specific details, including sample counts, input resolution, class composition, and data splits, are provided below, with representative examples shown in Figure~\ref{fig:appendix-dataset-vis}.

\begin{itemize}[noitemsep, nolistsep]
    \item \textbf{MNIST~\citep{lecun1998gradient}:} contains 28x28 greyscale images of handwritten digits from 0-9, spanning 10 classes. We use a split of 60,000 training samples, 8000 test samples, and 2000 validation samples. Pixel values are normalised to $[0,1]$. In our experiments, MNIST serves as one of the ID datasets. It is paired with FashionMNIST and KMNIST as far-OOD datasets because they contain no overlapping classes while retaining similar visual characteristics, including greyscale appearance, low resolution, and a hand-drawn style. MNIST is also paired with EMNIST as a near-OOD dataset because the two datasets share similar visual characteristics and overlapping classes, with EMNIST containing handwritten digits in addition to handwritten letters.
    \item \textbf{FashionMNIST~\citep{xiao2017fashion}:} contains 28x28 greyscale images of clothing items such as coats, bags, and t-shirts, spanning 10 classes. We use the widely adopted standard split of 60,000 training samples, 8000 test samples, and 2000 validation samples. Pixel values are normalised to $[0,1]$. In our experiments, FashionMNIST is used as a far-OOD pairing with the ID MNIST dataset.
    \item \textbf{KMNIST~\citep{clanuwat2018deep}:} contains 28x28 greyscale images of handwritten Japanese characters, specifically Kuzushiji characters from classical Japanese literature, spanning 10 classes. We use the widely adopted standard split of 60,000 training samples, 8000 test samples, and 2000 validation samples. Pixel values are normalised to $[0,1]$. In our experiments, KMNIST is used as a far-OOD pairing with the ID MNIST dataset.
    \item \textbf{EMNIST~\citep{cohen2017emnist}:} contains 28x28 greyscale images of handwritten Latin letters and digits from 0-9, spanning 47 classes. We use a split of 112,800 training samples, 15,040 test samples, and 3760 validation samples. Pixel values are normalised to $[0,1]$. In our experiments, EMNIST is used as a near-OOD pairing with the ID MNIST dataset because it shares similar visual characteristics and overlapping handwritten digit classes.
    \item \textbf{MNIST-C~\citep{mu2019mnistc}:} contains the same data points as the original MNIST dataset, with images corrupted using 15 different distributional shifts such as shot noise, glass blur, and zigzag artefacts. In our experiments, MNIST-C is used as a near-OOD pairing with the ID MNIST dataset because it contains the same underlying data under distributional shift.
    \item \textbf{CIFAR10~\citep{krizhevsky2009learning}:} contains 64x64 RGB images of real-world objects such as birds, trucks, airplanes, and frogs, spanning 10 classes. We use the widely adopted standard split of 50,000 training samples, 8000 test samples, and 2000 validation samples. Pixel values are normalised to $[0,1]$ across all three RGB channels. In our experiments, CIFAR10 serves as one of the ID datasets. CIFAR10 is paired with FashionMNIST and SVHN as far-OOD datasets because they contain no overlapping classes while exhibiting similar visual characteristics, including coloured appearance and real-world imagery. CIFAR-10 is also paired with CIFAR-100 as a near-OOD dataset because the two datasets share similar visual characteristics and overlapping classes. Specifically, CIFAR-10 contains broad object categories such as cat, dog, truck, and ship, which correspond to finer-grained classes such as house cat, beagle, pickup truck, and cruise ship in CIFAR-100.
    \item \textbf{CIFAR100~\citep{krizhevsky2009learning}:} contains 64x64 RGB images of real-world objects spanning 100 classes, including fine-grained counterparts of classes found in CIFAR10. For example, CIFAR10 contains the class \textit{truck}, whereas CIFAR100 contains classes such as \textit{pickup truck} and \textit{train}. We use the widely adopted standard split of 50,000 training samples, 8000 test samples, and 2000 validation samples. Pixel values are normalised to $[0,1]$ across all three RGB channels. In our experiments, CIFAR100 is used as a near-OOD pairing with the ID CIFAR10 dataset because the datasets share similar visual characteristics and overlapping classes.
    \item \textbf{CIFAR-10-C~\citep{hendrycks2019robustness}:} consists of the same data points as the traditional CIFAR-10 dataset, but where the images have been corrupted with 15 different distributional shifts (e.g., shot noise, JPEG-artefacts, zigzag artefacts, etc.). In our experimental evaluation, we use this dataset as a near-OOD pairing with the ID CIFAR-10 dataset due to it sharing the same dataset but with some distributional shift.
    \item \textbf{SVHN~\citep{netzer2011reading}:} contains 64x64 RGB images of house numbers collected from Google Street View, spanning 10 digit classes from 0-9. We use a split of 73,257 training samples, 10,832 test samples, and 5200 validation samples. Pixel values are normalised to $[0,1]$ across all three RGB channels. In our experiments, SVHN is used as a far-OOD pairing with the ID CIFAR10 dataset due to the substantial domain shift between street-level digit photographs and object-centric natural images.
    \item \textbf{Oxford Flowers~\citep{Nilsback08}:} contains 224x224 RGB images of flowers commonly found in the United Kingdom, spanning 102 classes such as Water Lily and Wild Pansy. We use a split of 4913 training samples, 819 test samples, and 819 validation samples. Pixel values are normalised to $[0,1]$ across all three RGB channels. In our experiments, Oxford Flowers is used as an ID dataset for classification, and as an OOD pairing for the UTKFace regression task. It is paired with Deep Weeds because the two datasets contain no overlapping classes while sharing similar visual characteristics, including natural features such as leaves and flowers.
    \item \textbf{Deep Weeds~\citep{olsen2019deepweeds}:} contains 224x224 RGB images of weed species found in Australia, spanning 8 classes such as Snake weed and Rubber vine. We use the standard split of 10,505 training samples, 3502 test samples, and 3502 validation samples. Pixel values are normalised to $[0,1]$ across all three RGB channels. In our experiments, Deep Weeds serves as a far-OOD pairing with the ID Oxford Flowers dataset due to the substantial domain shift between photographs of flowers and photographs of weeds.
    \item \textbf{California Housing~\citep{pace1997sparse}:} contains 20,640 California district records with eight numerical input features and median house value as the regression target. We use it as an ID dataset and evaluate a target-shift setting in which districts with extreme house prices form the OOD subset.
    \item \textbf{Wine Quality~\citep{cortez2009modeling}:} contains physicochemical measurements of red and white wines, with eleven numerical input features and a sensory quality score as the regression target. We train on white-wine samples and use red-wine samples as OOD inputs, testing whether uncertainty reflects a change in the input distribution while the prediction target remains wine quality.
    \item \textbf{Airfoil Self-Noise~\citep{brooks1989airfoil}:} contains 1,503 observations with five numerical input features describing airfoil operating conditions and scaled sound pressure level as the regression target. We use it to assess covariate shift, defining OOD samples by extreme values of the input frequency rather than by the regression target.
    \item \textbf{UTKFace~\citep{zhifei2017cvpr}:} contains over 20,000 face images with age annotations spanning 0-116 years. We use age prediction as an image regression task and evaluate two OOD settings: Oxford Flowers images, which introduce a substantial image-domain shift, and faces at age extremes, which remain within the face domain but differ in target distribution.
    \item \textbf{CelebA~\cite{liu2015faceattributes}:} consists of $178\times218$ aligned RGB images of celebrity faces, spanning a single object-detection class (face). We utilise the standard split of 162,770 training images, 19,867 calibration images, and 19,962 test images. Images were resized to $640\times640$ pixels, with pixel normalisation bounded $[0,1]$ across all three RGB channels.
    \item \textbf{ImageNet~\citep{deng2009imagenet}:} consists of 224x224 RGB images across 1000 object classes, each corresponding to a subset of ImageNet. The dataset contains 1,281,167 training samples, 50,000 validation samples, and 100,000 test samples without publicly available labels. Pixel values were normalised to the range $[0,1]$ across all RGB channels. In our experimental evaluation, ImageNet is used as an ID dataset. ImageNet was paired with Caltech-UCSD Birds-200-2011 because they share very few overlapping classes but exhibit similar visual characteristics.
    \item \textbf{Caltech-UCSD Birds-200-2011 (CUB)~\citep{welinder2010caltech}:} consists of 224x224 RGB images of 200 bird species, collected from various natural environments. The dataset contains a total of 11,788 images, with a standard split of 5,994 training samples and 5,794 test samples. Pixel values were normalised to the range $[0,1]$ across all RGB channels. In our experimental evaluation, CUB is used as an OOD dataset paired with ImageNet.
\end{itemize}

For datasets lacking predefined partitions, we created manual splits using stratification to maintain consistent class distributions across subsets. In all experiments, the validation portion of each dataset was employed to calibrate the ID–OOD threshold for uncertainty-based rejection. This calibration enabled evaluation of both ID and OOD coverage on the corresponding test sets.

\subsection{Models}
\label{sec:appendix-Models}

To assess the performance and generalisability of CLEAR, we assess its performance against multiple models and architectures across our experimental evaluation. Below, we provide a concise description of each model together with the implementation details and hyperparameter settings adopted in our experiments:

\begin{itemize}[noitemsep, nolistsep]
    \item \textbf{ResNet-18~\citep{he2016deep}:} is used for the MNIST classification and image-based regression experiments. We train it from scratch, setting the initial convolution to stride one and removing the initial max-pooling layer for small images. The task-specific head is trained for 10 epochs, followed by 40 epochs of backbone training. The resulting model is frozen before CLEAR is applied.
    \item \textbf{EfficientNet-B0~\citep{tan2019efficientnet}:} is used for the CIFAR experiments. We initialise the TorchVision model from ImageNet-1K pretrained weights and set its first convolution to stride one for small images. The task-specific head is trained for 10 epochs, followed by 40 epochs of backbone fine-tuning. The resulting model is frozen before CLEAR is applied.
    \item \textbf{DenseNet-121~\citep{huang2017densely}:} is used for the Oxford Flowers and DeepWeeds experiments. We initialise the TorchVision model from ImageNet-1K pretrained weights and train it on the in-distribution Oxford Flowers task, with 10 epochs of head training followed by 40 epochs of backbone fine-tuning. The resulting model is frozen before CLEAR is applied.
    \item \textbf{ViT-B/16~\citep{dosovitskiy2021an,vaswani2017attention}:} is used for the ImageNet and CUB experiments. We initialise the TorchVision model from ImageNet-1K pretrained weights and train its task-specific head for 10 epochs on ImageNet, keeping the backbone frozen. The resulting model is frozen before CLEAR is applied.
    \item \textbf{Tabular MLP:} is used for the tabular regression experiments. It comprises two fully connected layers of width 64, each followed by a ReLU activation. Its learned 64-dimensional representation is used by the regression head and CLEAR.
    \item \textbf{YOLO11n~\citep{khanam2024yolov11}:} is used as the base object detector for the object detection experiments. We initialise the lightweight nano variant from COCO-pretrained weights and finetuned for 20 epochs for CelebA. Following fine-tuning, all detector parameters are frozen during CLEAR.
\end{itemize}

\subsection{ID-OOD Thresholds}
\label{sec:appendix-ID-OOD Thresholds}

To support abstention from uncertain predictions, we determine the optimal ID–OOD threshold using the validation set corresponding to each experimental configuration. Our evaluation considers several ID–OOD scoring metrics, enabling a rigorous assessment of CLEAR alongside the baseline methods.

\begin{itemize}[noitemsep, nolistsep]
\item \textbf{Differential Entropy:} quantifies the dispersion, or uncertainty, represented by the Dirichlet distribution. It is defined as:

\vspace{0.25em}
$\qquad\qquad\quad \sum_{k=1}^{K} \ln \Gamma(\alpha_k) - \ln \Gamma(S) - \sum_{k=1}^{K} (\alpha_k - 1)(\Psi(\alpha_k) - \Psi(S))$
\vspace{0.4em}

where $B(\boldsymbol{\alpha})$ denotes the multivariate Beta function, $S$ is the total concentration parameter, and $\Psi(\cdot)$ denotes the digamma function. Higher entropy values correspond to greater uncertainty and are therefore typically associated with OOD inputs. This score is applicable only to Dirichlet-based uncertainty models, such as posterior networks and evidential networks.

\item \textbf{Predictive Entropy:} quantifies uncertainty in the model's predictive class distribution. Higher values indicate greater uncertainty, with the maximum attained when all classes are equally probable. This score is applicable to any model that provides predictive class probabilities.

\item \textbf{Mutual Information:} measures epistemic uncertainty by quantifying how much information the model parameters contribute to the predictive distribution. It is defined as the difference between the predictive entropy and the expected conditional entropy under the posterior over model parameters:

\vspace{0.25em}
$\qquad\qquad\qquad\quad\; -H\!\left(\mathbb{E}_{q(\omega)}[\,p(y\mid x,\omega)\,]\right)\;-\;\mathbb{E}_{q(\omega)}\!\left[\,H\!\left(p(y\mid x,\omega)\right)\right]$
\vspace{0.4em}

where $H(p)=-\sum_{k=1}^{K} p_k \log p_k$ and $q(\omega)$ denotes the explicit or implicit posterior over model parameters.

\item \textbf{Maximum Probability:} provides a simple confidence score based on the model's most probable prediction. It is defined as the maximum softmax probability across all classes:

\vspace{0.25em}
$\qquad\qquad\qquad\qquad\qquad\qquad\quad \max_{k \in \{1,\dots,K\}} \; p(y=k \mid x)$
\vspace{0.4em}

Higher values indicate greater confidence in the predicted class, whereas lower values indicate greater uncertainty.

\end{itemize}

Following prior work~\citep{shen2023post, barker2026robust}, the optimal ID-OOD threshold is determined using the validation datasets. For a selected scoring metric from those described above, we first compute scores for both ID and OOD validation samples. These scores are then used to construct a receiver operating characteristic (ROC) curve, treating ID samples as positive and OOD samples as negative. The ROC curve characterises how effectively the selected metric separates ID and OOD data. The decision threshold is chosen to maximise $\text{TPR} - \text{FPR}$, corresponding to the point of optimal separation. This criterion provides a principled balance between retaining ID inputs and rejecting OOD inputs, avoids ad hoc threshold tuning, and enables clear reporting of ID and OOD coverage under deployment-like conditions.

Following calibration, the selected threshold is fixed and applied to the corresponding test sets. During evaluation, any test input whose score under the chosen metric crosses this threshold is rejected. This provides a consistent basis for comparing ID retention and OOD rejection across all models and datasets.

\subsection{Comparative Approaches}
\label{sec:appendix-Comparative Methods}

To evaluate the effectiveness of CLEAR, we compare it against a range of recent EDL and uncertainty quantification approaches that are representative of the current state-of-the-art. Each approach is briefly described below, alongside any implementation-specific details or hyperparameter choices used in our experiments:

\begin{itemize}[noitemsep, nolistsep]
    \item \textbf{Posterior Networks~\citep{charpentier2020posterior}:} provide a foundational approach to uncertainty estimation in classification that predates and informs later evidential deep learning methods. They avoid requiring OOD data during training by directly modelling a closed-form posterior over categorical distributions. Specifically, the model predicts Dirichlet concentration parameters $\alpha=\beta_{\text{prior}}+\beta$ for each input $x$, where $\beta$ denotes pseudo-counts obtained from class-conditional density estimates in a learned latent space and $\beta_{\text{prior}}$ is a fixed symmetric prior. The resulting Dirichlet distribution represents both aleatoric and epistemic uncertainty in closed form.
    \item \textbf{Evidential Deep Learning (EDL)~\citep{sensoy2018evidential}:} provides a deterministic alternative to Bayesian neural networks for uncertainty estimation by modelling class probabilities through a Dirichlet distribution whose parameters are derived from the model's non-negative outputs. The model is trained using an objective that combines squared prediction error, predictive variance, and a KL divergence to a uniform prior. This evidential formulation enables both aleatoric and epistemic uncertainty to be quantified in closed form without requiring sampling or OOD examples during training. EDL serves as the base approach for the subsequent comparative methods. As the base method, its hyperparameters are shared across the remaining comparative approaches to ensure a fair comparison and are described in Appendix~\ref{sec:appendix-Training Details}.
    \item \textbf{Fisher Information-Based Evidential Deep Learning ($\mathbf{\mathcal{I}}$-EDL)~\citep{deng2023uncertainty}:} extends EDL by incorporating the Fisher Information Matrix (FIM) to adaptively weight the loss according to the informativeness of the predicted evidence. The central idea is that classes associated with greater evidence contain less Fisher information and should therefore be regularised less strongly. The method introduces a regularisation term based on the log-determinant of the FIM to discourage overconfident predictions:

    \vspace{0.25em}
    $\qquad\qquad\qquad \mathcal{L}^{|\mathcal{I}|}_i = \sum_{j=1}^K \log \psi^{(1)}(\alpha_{ij}) + \log \left(1 - \frac{\psi^{(1)}(\alpha_{i0})}{\sum_{j=1}^K \psi^{(1)}(\alpha_{ij})} \right)$
    \vspace{0.4em}

    In our experiments, we set $\lambda_{|\mathcal{I}|}=0.001$ and anneal it linearly over the first 10 epochs, following the guidance in Appendix C.2 of the original paper together with our manual tuning.
    \item \textbf{Hyper-Opinion Evidential Deep Learning (H-EDL)~\citep{qu2024hyper}:} extends classical EDL through hyper-opinions, which generalise multinomial opinions from Subjective Logic by representing uncertainty over both singleton classes and composite subsets. This enables H-EDL to capture both sharp evidence, which supports a single class, and vague evidence, which supports multiple plausible classes, thereby improving robustness to ambiguous inputs. The method introduces an opinion projection mechanism that maps hyper-opinions to standard Dirichlet-based predictions, allowing training within the conventional EDL framework. This projection also addresses the vanishing gradient problem that can limit traditional EDL on fine-grained tasks. In our experiments, we adopt the two-stage training procedure described in the original paper without introducing additional hyperparameters.
    \item \textbf{Relaxed Evidential Deep Learning (R-EDL)~\citep{chen2024r}:} addresses overconfidence in standard EDL by relaxing two nonessential assumptions. First, it replaces the fixed prior weight, typically equal to the number of classes, with a tunable scalar hyperparameter $\lambda$ that controls the contribution of the base rate $\alpha$ in the Dirichlet construction:

    \vspace{0.25em}
    $\qquad\qquad\qquad\qquad\qquad\qquad\qquad\qquad \alpha(x)=e(x)+\lambda$
    \vspace{0.4em}

    Second, R-EDL removes the variance-penalising regularisation term from the EDL objective and instead directly optimises the projected class probabilities $P(x)=\frac{\alpha(x)}{S}$ against the one-hot labels. This simplification provides a better balance between evidence magnitude and evidence proportion, particularly in OOD settings. In our experiments, we follow the guidance from Appendix C.2 and Figure 1(b) of the original paper and use $\lambda=0.1$ as a fixed prior weight throughout training.
    \item \textbf{Regularized Evidential Model (RED)~\citep{pandey2023learn}:} addresses the inability of conventional evidential models to learn effectively from samples mapped to zero-evidence regions. RED introduces a vacuity-guided correct-evidence regularisation term that encourages greater evidence for the ground-truth class, with its influence strongest when evidence is low. This enables the model to learn from samples that would otherwise produce negligible gradient updates while retaining the standard evidential learning objective.
    \item \textbf{Smoothed Evidential Deep Learning (S-EDL)~\citep{kopetzki2021evaluating}:} improves the robustness of standard evidential models by applying median smoothing to their uncertainty estimates. Given an input $x$, a collection of noisy samples $\{x_s\}\sim\mathcal{N}(x,\sigma)$ is generated. The final uncertainty estimate is obtained by taking the median across these samples, improving robustness to adversarial perturbations by reducing the influence of outliers. In our experiments, we follow the original setup and use a Gaussian noise scale of $\sigma=0.01$ with $50$ samples per input, based on guidance from the original paper and our manual tuning. S-EDL is a purely post-hoc method.
    \item \textbf{Conflict-aware Evidential Deep Learning (C-EDL)~\citep{barker2026robust}:} is a lightweight post-hoc extension of EDL that improves OOD and adversarial robustness without retraining the base model. C-EDL generates multiple label-preserving transformed views of an input and measures disagreement between their evidential outputs using intra-class variability and inter-class conflict. The resulting conflict score is used to reduce evidential strength when disagreement is high, increasing uncertainty while largely preserving predictions for consistent ID inputs.
    \item \textbf{Deep Evidential Regression (DER)~\citep{amini2020deep}:} extends evidential learning to continuous prediction by placing a Normal-Inverse-Gamma (NIG) distribution over the mean and variance of a Gaussian likelihood. A single network predicts the NIG parameters $(\gamma,\nu,\alpha,\beta)$, where $\gamma$ represents the predictive mean and the remaining parameters determine the evidential strength and associated aleatoric and epistemic uncertainty. Training combines the negative log-likelihood of the resulting Student-$t$ predictive distribution with an error-dependent evidence regulariser, which penalises high evidence when the prediction is inaccurate. This enables both forms of uncertainty to be obtained analytically from a single forward pass without requiring sampling or OOD examples during training. The setup is described in Section~\ref{sec:Preliminaries}. In our experiments, we use an evidential regularisation weight of $\lambda_{\mathrm{reg}}=10^{-3}$.
    \item \textbf{Uncertainty Regularised Evidential Regression Network (UR-ERN)~\citep{ye2024uncertainty}:} extends DER by addressing the high uncertainty area (HUA), in which the constrained activations used to ensure valid evidential parameters can produce vanishing gradients and prevent highly uncertain samples from contributing effectively to learning. UR-ERN introduces an additional uncertainty regularisation term that maintains a learning signal within this region, with its influence increasing with the discrepancy between the predicted value and ground truth. This allows the model to move samples away from poorly learned high-uncertainty regions while retaining the original evidential regression objective and NIG uncertainty formulation. In our experiments, we use an uncertainty regularisation weight of $10^{-3}$.
    \item \textbf{Evidential Meta Model (EMM)~\citep{shen2023post}:} is a post-hoc uncertainty approach that combines representations from multiple frozen intermediate layers using a lightweight evidential meta-model. We adapt EMM to object detection by applying ROIAlign to predefined detector features for each predicted bounding box and parameterising a Normal-Inverse-Gamma distribution over its four localisation coordinates. The meta-model is trained for 20 epochs using Adam with an initial learning rate of $10^{-2}$, cosine annealing to $10^{-6}$, an IoU matching threshold of $0.5$, evidential regularisation weight $\lambda_{\mathrm{reg}}=0.1$, and gradient clipping at $1.0$.
\end{itemize}

\subsection{Adversarial Attacks}
\label{sec:appendix-Adversarial Attacks}

This section presents the adversarial attack strategies used to assess the robustness of uncertainty-aware models. Adversarial attacks introduce carefully constructed, often imperceptible perturbations that can cause a model to misclassify while remaining highly confident. These attacks are particularly challenging for uncertainty-aware methods because they can disrupt both aleatoric and epistemic uncertainty signals, causing adversarial inputs to appear in-distribution or evade uncertainty-based detection.

To provide a comprehensive robustness evaluation, we consider both gradient-based and gradient-free attacks. Gradient-based attacks represent white-box settings, where access to model gradients is used to construct adversarial perturbations. Conversely, gradient-free attacks apply random or structured perturbations without relying on gradient information and more closely reflect black-box settings. Where supported, attacks are implemented using Foolbox~\citep{rauber2017foolbox} under the following configurations:

\begin{itemize}[noitemsep, nolistsep]
    \item \textbf{$L_2$ Projected Gradient Descent (L2PGD)~\citep{madry2018towards}:} is an iterative, white-box adversarial attack that perturbs inputs within a bounded $L_2$-norm ball to maximise the model’s loss. At each iteration, the input is updated in the direction of the gradient of the loss with respect to the input, followed by projection back onto the $L_2$-ball of radius $\epsilon$. This results in smooth, high-precision perturbations that remain less perceptible to humans:

    \vspace{0.25em}
    $\qquad\qquad\qquad\qquad\qquad x'_{t+1} = \text{Proj}^{L_2}_\epsilon(x'_t+\alpha \cdot \nabla_xJ(x'_t, y))$
    \vspace{0.4em}

    where $\alpha$ is the step size and $J(x, y)$ is the loss function.
    \item \textbf{Fast Gradient Sign Method (FGSM)~\citep{goodfellow2015explaining}:} is a single-step white-box adversarial attack that perturbs the input in the direction of the sign of the gradient of the loss:

    \vspace{0.25em}
    $\qquad\qquad\qquad\qquad\qquad x' = x + \epsilon \cdot \text{sign}(\nabla_x J(x, y))$
    \vspace{0.4em}
    
    where $\epsilon$ controls the perturbation magnitude. FGSM generates perceptible but targeted perturbations with minimal computational overhead.
    
    \item \textbf{Salt \& Pepper Noise:} is a non-gradient-based black-box perturbation that randomly sets a proportion of input pixels to their minimum or maximum value. This form of structured noise simulates impulsive corruption and tests the model’s resilience to sparse, high-intensity artefacts. It does not rely on model gradients and is agnostic to internal model parameters.
    \item \textbf{AutoAttack $L_2$~\citep{croce2020reliable}:} is an ensemble of complementary adversarial attacks comprising Auto-PGD with cross-entropy and difference-of-logits-ratio losses, the Fast Adaptive Boundary attack, and the black-box Square Attack. Adversarial perturbations are constrained by $\|x'-x\|_2 \leq \epsilon$. The attacks are applied sequentially to inputs that remain correctly classified, combining gradient-based and query-based strategies to evaluate robustness without model-specific hyperparameter tuning.
\end{itemize}

\subsection{Training Details}
\label{sec:appendix-Training Details}
To ensure that differences between uncertainty methods are not driven by method-specific optimisation, we use a common training configuration wherever possible. All models are trained with a batch size of 64 using the Adam optimiser with weight decay $10^{-4}$. For MNIST, CIFAR-10, and Oxford Flowers, training follows the two-stage procedure described above: the task-specific head is first trained with the feature extractor frozen, followed by joint fine-tuning of the backbone and head. The head is initially trained with a learning rate of $10^{-3}$; during fine-tuning, the backbone and head use learning rates of $10^{-5}$ and $10^{-4}$, respectively. For ImageNet, the ImageNet-1K pretrained ViT-B/16 backbone remains frozen and only the task-specific head is trained, using a learning rate of $10^{-4}$. The tabular MLP uses a learning rate of $10^{-3}$ throughout. Learning rates are reduced by a factor of $0.5$ when validation loss does not improve for five epochs, with a minimum learning rate of $10^{-8}$. A dropout rate of $0.25$ is used in the shared predictive architecture. These optimisation settings are kept fixed across comparative approaches for a given dataset, with method-specific parameters varying only where required by the corresponding uncertainty formulation.

\subsection{Runtime and Complexity}
\label{sec:Runtime and Complexity}

All experiments were implemented in PyTorch and executed on a large performance GPU cluster using a maximum of three Nvidia A40 GPUs, 32 CPU cores, 167GB of memory (per GPU). All models were trained from scratch or used pre-trained weights. All runs used random seeds.

CLEAR introduces a one-off post-hoc calibration stage followed by a lightweight adjustment at inference. During calibration, estimating the group-conditioned covariance structure over the $d$-dimensional latent representations requires $\mathcal{O}(|\mathcal{D}_{\mathrm{cal}}|d^2)$ operations. Computing the Cholesky factor and latent scale for each group contributes $\mathcal{O}(Kd^3)$ for classification, or $\mathcal{O}(Bd^3)$ for regression. Estimating the reference conflict additionally evaluates $T$ perturbation draws for each calibration example, giving $\mathcal{O}(|\mathcal{D}_{\mathrm{cal}}|T(Vd^2+V^2d))$. The resulting calibration complexity is therefore:
\begin{equation}
\mathcal{O}\!\left(
|\mathcal{D}_{\mathrm{cal}}|d^2
+ Kd^3
+ |\mathcal{D}_{\mathrm{cal}}|T(Vd^2+V^2d)
\right)
\end{equation}
for classification, with $K$ replaced by $B$ for regression. This cost is incurred only once after training.

At inference, CLEAR requires only the original forward pass through the frozen evidential model, since the $V$ perturbation views are generated directly in latent space. Whitening these views using the calibrated Cholesky factor requires $\mathcal{O}(Vd^2)$ operations. The $V(V-1)/2$ unordered view pairs are then evaluated in $\mathcal{O}(V^2d)$ time, since the cross-dimensional conflict in Equation 14 can be computed using sums over the $d$ latent dimensions rather than explicitly evaluating all $d(d-1)$ interactions. The additional CLEAR inference complexity is therefore:
\begin{equation}
\mathcal{O}\!\left(Vd^2+V^2d\right)
\end{equation}
With the default $V=3$, this reduces to $\mathcal{O}(d^2)$ additional computation per prediction.

The empirical inference times in Table~\ref{tab:extra-metrics-results} are consistent with this analysis. Across the evaluated benchmarks, CLEAR remains close to the corresponding single-pass EDL inference time. On MNIST$\rightarrow$FashionMNIST, for example, inference increases from $1.96$\,s for EDL to $2.36$\,s for CLEAR, while C-EDL and S-EDL require $13.15$\,s and $57.56$\,s, respectively. The difference becomes particularly pronounced for larger inputs: on Oxford Flowers, CLEAR requires $18.69$\,s compared with $18.48$\,s for EDL, $74.46$\,s for C-EDL, and $660.19$\,s for S-EDL. Similarly, on ImageNet, CLEAR requires $423.66$\,s compared with $397.02$\,s for EDL and $7356.02$\,s for S-EDL. These results show that operating directly on the latent representation introduces a relatively small overhead compared with the frozen base model while avoiding the repeated full-network evaluations required by more computationally expensive post-hoc approaches.


\end{document}